\documentclass[journal, 10pt]{IEEEtran} 
\usepackage{times}
\usepackage{amsmath,amssymb,float,arydshln,color}
\usepackage{psfrag,setspace,wrapfig,subfigure,float}
\usepackage[utf8]{inputenc}
\usepackage{pifont}
\usepackage{dsfont}
\usepackage{epsfig}
\usepackage{epstopdf}
\usepackage{graphicx}
\usepackage{scrextend}
\usepackage{hyperref}
\usepackage[ruled]{algorithm2e}
\usepackage{amsfonts}
\usepackage{cite}
\usepackage{hyperref}
\usepackage{url}
\usepackage{booktabs}       
\usepackage{hhline}

\usepackage[table]{xcolor}
\usepackage{multirow}
\usepackage{tabu}
\allowdisplaybreaks

\newtheorem{lemma}{{Lemma}}
\newtheorem{assumption}{{ Assumption}}
\newtheorem{theorem}{{Theorem}}

\newcommand{\HRule}{\rule{\linewidth}{0.5mm}}

\def\tran{^{\mathsf{T}}}

\DeclareMathOperator*{\argmin}{arg\,min}
\newcommand{\bp}{ \begin{proof}}
	\newcommand{\ep}{\end{proof} }

\newcommand{\bm}[1]{\mbox{\boldmath $#1$}}

\newcommand{\be}{\begin{equation}}
\newcommand{\ee}{\end{equation}}
\newcommand{\bqq}{\begin{eqnarray}}
\newcommand{\eqq}{\end{eqnarray}}
\newcommand{\bal}{\begin{align}}
\newcommand{\eal}{\end{align}}
\newcommand{\bqn}{\begin{eqnarray*}}
	\newcommand{\eqn}{\end{eqnarray*}}
\newcommand{\nn}{\nonumber}
\newcommand{\ba}{\left[ \begin{array}}
	\newcommand{\ea}{\\ \end{array} \right]}
\newcommand{\qd}{\hfill{$\blacksquare$}}
\newcommand{\define}{\;\stackrel{\Delta}{=}\;}

\def\btheta  {{\boldsymbol \theta}}

\def\bpsi       {{\boldsymbol \psi}}
\def\bsigma  {{\boldsymbol \sigma}}
\def\bphi  {{\boldsymbol \phi}}

\def\bgamma {{\boldsymbol \gamma}}

\def\H{{\boldsymbol{H}}}

\def\g{{\boldsymbol{g}}}
\def\h{{\boldsymbol{h}}}

\def\n{{\boldsymbol{n}}}

\def\s{{\boldsymbol{s}}}

\def\w{{\boldsymbol{w}}}

\def\z{{\boldsymbol{z}}}

\newcommand{\vA}{{\mathbf{A}}}
\newcommand{\vB}{{\mathbf{B}}}
\newcommand{\vC}{{\mathbf{C}}}

\newcommand{\cA}{{\mathcal{A}}}
\newcommand{\cB}{{\mathcal{B}}}

\newcommand{\cD}{{\mathcal{D}}}

\newcommand{\cI}{{\mathcal{I}}}
\newcommand{\cJ}{{\mathcal{J}}}

\newcommand{\cL}{{\mathcal{L}}}

\newcommand{\cN}{{\mathcal{N}}}

\newcommand{\cR}{{\mathcal{R}}}

\newcommand{\cV}{{\mathcal{V}}}

\newcommand{\cX}{{\mathcal{X}}}

\newcommand{\sw}{{\scriptstyle{\mathcal{W}}}}

\newcommand{\swb}{{\boldsymbol{\scriptstyle{\mathcal{W}}}}}
\newcommand{\syb}{{\boldsymbol{\scriptstyle{\mathcal{Y}}}}}

\newcommand{\sy}{{\scriptstyle{\mathcal{Y}}}}

\newcommand{\sz}{{\scriptstyle{\mathcal{Z}}}}

\newcommand{\sxb}{\boldsymbol{\scriptstyle{\mathcal{X}}}}

\newcommand{\tws}{\widetilde{\scriptstyle{\boldsymbol{\mathcal{W}}}}}
\newcommand{\tys}{\widetilde{\scriptstyle{\boldsymbol{\mathcal{Y}}}}}

\newcommand{\tcA}{\overline{{\mathcal{A}}}}
\newcommand{\barA}{\overline{A}}

\newcommand{\col}{{\mathrm{col}}}
\newcommand{\diag}{{\mathrm{diag}}}
\newcommand{\cHb}{\boldsymbol{\mathcal{H}}}
\newcommand{\cTb}{\boldsymbol{\mathcal{T}}}

\newcommand{\prox}{\mbox{prox}}
\newcommand{\RR}{\mathbb{R}}
\newcommand{\grad}{{\nabla}}

\def\bE{\mathbb{E}}

\newcommand{\eq}[1]{\begin{align}#1\end{align}}
\newcommand{\beqn}{\begin{eqnarray}}
\newcommand{\eeqn}{\end{eqnarray}}

\newcommand{\nnb}{\nonumber \\}

\DeclareFontFamily{U}{mathx}{\hyphenchar\font45}
\DeclareFontShape{U}{mathx}{m}{n}{
	<5> <6> <7> <8> <9> <10>
	<10.95> <12> <14.4> <17.28> <20.74> <24.88>
	mathx10
}{}
\DeclareSymbolFont{mathx}{U}{mathx}{m}{n}
\DeclareFontSubstitution{U}{mathx}{m}{n}
\DeclareMathAccent{\widebar}{0}{mathx}{"73}

\def\real{{\mathbb{R}}}

\def\Zint{{\mathchoice{\setbox1=\hbox{\sf Z}\copy1\kern-.75\wd1\box1}
		{\setbox1=\hbox{\sf Z}\copy1\kern-.75\wd1\box1}
		{\setbox1=\hbox{\scriptsize\sf Z}\copy1\kern-.75\wd1\box1}
		{\setbox1=\hbox{\scriptsize\sf Z}\copy1\kern-.75\wd1\box1}}}

\makeatletter
\def\hlinewd#1{%
	\noalign{\ifnum0=`}\fi\hrule \@height #1 \futurelet
	\reserved@a\@xhline}
\makeatother

\begin{document}
	\def\helvetica{phvr7t.tfm}
	\def\helveticaoblique{phvro7t.tfm}
	\def\helveticabold{phvb7t.tfm}
	\def\helveticaboldoblique{phvbo7t.tfm}
	
	\font\sfb=\helveticabold
	=\helveticaboldoblique
	\title{
		Variance-Reduced Stochastic
		Learning by Networked Agents under Random Reshuffling
	}
	\author{Kun Yuan, Bicheng Ying, Jiageng Liu and Ali H. Sayed\\
		
		\thanks{\scriptsize{K. Yuan and B. Ying are with the Department of Electrical Engineering, University of California, Los Angeles, CA 90095 USA. J. Liu is with the Department of Mathematics, University of California, Los Angeles, CA 90095 USA. Email:\{kunyuan, ybc, bioliu\}@ucla.edu. A. H. Sayed is with the School of Engineering, Ecole Polytechnique Federale de Lausanne (EPFL), Switzerland. Email: ali.sayed@epfl.ch. This  work was supported in part by NSF grants CCF-1524250 and ECCS-1407712.}}
		
		}

	\maketitle
	\begin{abstract}\vspace{-0mm}
		A new amortized variance-reduced gradient (AVRG) algorithm was developed in \cite{ying2017convergence}, which has constant storage requirement in comparison to SAGA and balanced gradient computations in comparison to SVRG. One key advantage of the AVRG strategy is its amenability to decentralized implementations. In this work, we show how AVRG can be extended to the network case where multiple learning agents are assumed to be connected by a graph topology. In this scenario, each agent observes data that is spatially distributed and all agents are only allowed to communicate with direct neighbors. Moreover, the amount of data observed by the individual agents may differ drastically. For such situations, the balanced gradient computation property of AVRG becomes a real advantage in reducing idle time caused by unbalanced local data storage requirements, which is characteristic of other reduced-variance gradient algorithms. The resulting diffusion-AVRG algorithm is shown to have linear convergence to the exact solution, and is much more memory efficient than other alternative algorithms. In addition, we propose a mini-batch strategy to balance the communication and computation efficiency for diffusion-AVRG. When a proper batch size is employed, it is observed in simulations that diffusion-AVRG is more computationally efficient than exact diffusion or EXTRA while maintaining almost the same communication efficiency.  
		
	\end{abstract}\vspace{-1.5mm}
	\begin{IEEEkeywords}
		diffusion strategy, variance-reduction, stochastic gradient descent, memory efficiency, SVRG, SAGA, AVRG
	\end{IEEEkeywords}\vspace{-0.5mm}
	
	\setlength{\abovedisplayskip}{1.2mm}
	\setlength{\belowdisplayskip}{1.2mm}
	\section{Introduction and Motivation}
	This work considers empirical risk minimization under the decentralized network setting. For most traditional machine learning tasks, the training data are usually stored at a single computing unit \cite{bottou2010large,schmidt2013minimizing,johnson2013accelerating,defazio2014saga}. This unit can access the entire data set and can carry out training procedures in a centralized fashion. However, to enhance performance and accelerate convergence speed, there have also been extensive studies on replacing this centralized mode of operation by distributed mechanisms \cite{li2014communication,shamir2014communication,jaggi2014communication,lee2015distributed,konevcny2016federated}. In these schemes, the data may either be artificially distributed onto a collection of computing nodes (also known as {\em workers}), or it may already be physically collected by dispersed nodes or devices. These nodes can be smart phones or tablets, wireless sensors, wearables, drones, robots or self-driving automobiles. 
	Each node is usually assigned a local computation task and the objective is to enable the nodes to 
	converge towards the global minimizer of a central learning model. Nevertheless, in most of these distributed implementations, there continues to exist a central node, referred to as the {\em master}, whose purpose is to regularly collect intermediate iterates from the local workers, conduct global update operations, and distribute the  updated information back to all workers.
	
	Clearly, this mode of operation is not fully decentralized because it involves coordination with a central node. Such architectures are not ideal for on-device intelligence settings \cite{konevcny2016federated, hardy2017distributed} for various reasons. First, the transmission of local information to the central node, and back from the central node to the dispersed devices, can be expensive especially when communication is conducted via multi-hop relays or when the devices are moving and the network topology is changing. Second, there are privacy and secrecy considerations where individual nodes may be reluctant to share information with remote centers.
	Third, there is a critical point of failure in centralized architectures: when the central node fails, the operation comes to a halt. Moreover, the master/worker structure requires each node to complete its local computation before aggregating them at the master node, and the efficiency of the algorithms will therefore be dependent on the slowest worker.
	
	Motivated by these considerations, in this work we develop a fully decentralized solution for multi-agent network situations where nodes process the data locally and are allowed to communicate only with their immediate {\em neighbors}.  We shall assume that the dispersed nodes are connected through a network topology and that information exchanges are only allowed among neighboring devices. By ``neighbors'' we mean nodes that can communicate directly to each other as allowed by the graph topology. For example, in wireless sensor networks, neighboring nodes can be devices that are within the range of radio broadcasting. Likewise, in smart phone networks, the neighbors can be devices that are within the same local area network. In the proposed algorithm, there will be no need for a central or master unit and the objective is to enable each dispersed node to learn {\em exactly} the global model despite their limited localized interactions.
	
	
	\vspace{-1mm}
	\subsection{Problem Formulation}
	
	In a connected and undirected network with $K$ nodes, if node $k$ stores local data samples $\{x_{k,n}\}_{n=1}^{N_k}$, where $N_k$ is the size of the local samples, then the data stored by the entire network is:
	\eq{
		\{x_{n}\}_{n=1}^N \hspace{-1.5mm}\define \hspace{-1.5mm} \Big\{ \{x_{1,n}\}_{n=1}^{N_1}, \{x_{2,n}\}_{n=1}^{N_2}, \cdots, \{x_{K,n}\}_{n=1}^{N_K} \Big\}, 
	}
	where $N = \sum_{k=1}^{K}N_k$. We consider minimizing an empirical risk function, $J(w)$, which is defined as the sample average of loss values over {\em all} observed data samples in the network:\vspace{-1mm}
	\eq{\label{prob-emp-into}
		w^\star \define \argmin_{w\in \real^M}\;\;  J(w) &\define \frac{1}{N}\sum_{n=1}^{N}Q(w;x_n) \nnb
		&= \frac{1}{N}\sum_{k=1}^{K}\sum_{n=1}^{N_k}Q(w;x_{k,n}).
	}
	Here, the notation $Q(w;x_n)$ denotes the loss value evaluated at $w$ and the $n$-th sample, $x_n$.  We also introduce the local empirical risk function, $J_k(w)$, which is defined as the sample average of loss values over the  {\em local} data samples stored at node $k$, i.e., over $\{x_{k,n}\}_{n=1}^{N_k}$:\vspace{-1mm}
	\eq{\label{local-cost-func}
		J_k(w) \define \frac{1}{N_k}\sum_{n=1}^{N_k}Q(w;x_{k,n}).
	}
	Using the local empirical risk functions, $\{J_k(w)\}$, it can be verified that the original global optimization problem \eqref{prob-emp-into} can be reformulated as the equivalent problem of minimizing the weighted aggregation of
	$K$ local empirical risk functions:
	\setlength{\belowdisplayskip}{-1mm}
	\eq{\label{prob-emp-dist}
		w^\star \define \argmin_{w\in \real^M}\;\;  J(w) \define \sum_{k=1}^{K}q_k J_k(w).\vspace{-1mm}
	}\setlength{\belowdisplayskip}{0.7mm}
	\hspace{-1.2mm}where $q_k \stackrel{\Delta}{=}N_k/N$.
	The following assumptions are standard in the distributed optimization literature, and they are automatically satisfied by many loss functions of interest in the machine learning literature (such as quadratic losses, logistic losses --- see, e.g., \cite{sayed2014adaptation,sayed2014adaptive}). For simplicity in this article, we assume the loss functions are smooth, although the arguments can  be extended to deal with non-smooth losses, as we have done in \cite{ying2017performance,ying2017performance2}.
	
	\begin{assumption}\label{ass}
		The loss function, $Q(w;x_n)$, is convex,  twice-differentiable, and has a $\delta$-Lipschitz continuous gradient, i.e., for any $w_1,w_2\in \RR^M$ and $1\leq n\leq N$:
		\eq{\label{lsc}
			\|\grad_w Q(w_1; x_n)  - \grad_w Q(w_2; x_n) \| \le \delta \|w_1 - w_2\|
		}
		where $\delta > 0$. Moreover, there exists at least one loss function $Q(w;x_{n_o})$ that is strongly convex, i.e.,
		\eq{\label{strongly-convex}
			\grad^2_w Q(w;x_{n_o}) \ge \nu I_M >0,\;\;\mbox{\em for some $n_o$}.
		}
		\qd\vspace{-2mm}
	\end{assumption}\vspace{-3mm}

	\subsection{Related Work}\label{sec: related work}
	There exists an extensive body of research on solving optimization problems of the form \eqref{prob-emp-dist} in a fully decentralized manner. Some recent works include techniques such as ADMM\cite{mota2013d,shi2014linear}, DLM\cite{ling2015dlm}, EXTRA\cite{shi2015extra}, ESUM\cite{mokhtari2016decentralized},  DIGing\cite{nedich2016achieving}, Aug-DGM\cite{xu2015augmented} and exact diffusion\cite{yuan2017exact1,yuan2017exact2}. These methods provide linear convergence rates and are proven to converge to the {\em exact} minimizer, $w^{\star}$. The exact diffusion method, in particular, has been shown to have a wider stability range than EXTRA implementations (i.e., it is stable for a wider range of step-sizes, $\mu$), and is also more efficient in terms of communications than DIGing. However, all these methods require the evaluation of the true gradient vector of each $J_k(w)$ at each iteration. It is seen from the definition
	(\ref{local-cost-func}), and depending on the size $N_k$, that this computation can be prohibitive for large-data scenarios.
	
	One can resort to replacing the true gradient by a stochastic gradient approximation, as is commonplace in traditional diffusion or consensus algorithms \cite{sayed2014adaptive,sayed2014adaptation,nedic2009distributed,braca2008running,dimakis2010gossip,kar2013consensus+,kar2012distributed,yuan2016convergence}. In these implementations, each node $k$ approximates the true gradient vector $\grad J_k(w)$ by using one random sample gradient, $\grad Q(w;x_{k, \n})$, where $\n \in \{1,2,\cdots,N_k\}$ is a uniformly-distributed random index number. While this mode of operation is efficient, it has been proven to converge linearly only to a small $O(\mu)-$neighborhood around the exact solution $w^\star$ \cite{chen2015learning} where $\mu$ is the constant step-size. If convergence to the exact solution is desired, then one can employ decaying step-sizes instead of constant step-sizes; in this case, however, the convergence rate will be slowed down appreciably. An alternative is to employ variance-reduced techniques to enable convergence to the exact minimizer while employing a stochastic gradient approximation. One proposal along these lines is the DSA method \cite{mokhtari2016dsa}, which is based on the variance-reduced SAGA method \cite{defazio2014saga,schmidt2013minimizing}. However, similar to SAGA, the DSA method suffers from the same huge memory requirement since each node $k$ will need to store an estimate for each possible gradient $\{\grad Q(w;x_{k,n})\}_{n=1}^{N_k}$. This requirement is a burden when $N_k$ is large, as happens in applications involving large data sets. \vspace{-2mm}
	
	\subsection{Contribution}\label{llkad.contro}
	{This paper has three main contributions. First, we derive a fully-decentralized variance-reduced stochastic-gradient algorithm with significantly reduced memory requirements. We refer to the technique as the diffusion-AVRG method (where AVRG stands for the ``amortized variance-reduced gradient'' method proposed in the related work \cite{ying2017convergence} for single-agent learning). Unlike DSA\cite{mokhtari2016dsa}, the proposed method does not require extra memory to store gradient estimates.
		In addition,  diffusion-AVRG involves balanced gradient calculations and is amenable to scenarios in which the size of the data is unevenly distributed across the nodes. In contrast, diffusion-SVRG (an algorithm that builds upon exact diffusion and SVRG\cite{johnson2013accelerating}) introduces {\em imbalances} in the gradient calculations and hence suffers from significant idle time and delays in decentralized implementations --- see the discussions in Section \ref{sec:compare-svrg}. We also extend diffusion-AVRG to handle non-smooth but proximable cost functions.
		
		Second, we establish a linear convergence guarantee for diffusion-AVRG. The convergence proof is challenging for various reasons. One source of complication is the decentralized nature of the algorithm with nodes only allowed to interact locally. Second, due to the bias in the gradient estimate introduced by random reshuffling over data (i.e. sampling data without replacement), current analyses used for SVRG \cite{johnson2013accelerating}, SAGA \cite{defazio2014saga}, or DSA\cite{mokhtari2016dsa} are not suitable; these analyses can only deal with uniform sampling and unbiased gradient constructions. Third, the proposed diffusion-AVRG falls into a primal-dual structure where random reshuffling has not been studied throughly before.
		
		Third, this paper proposes mini-batch techniques to balance computations and communications in diffusion-AVRG. One potential drawback of diffusion-AVRG is that by approximating the true gradient with {\em one single} data sample, the algorithm requires more iterations and hence more communications to reach satisfactory accuracy. This limits the application of diffusion-AVRG in scenarios where communication is expensive. This issue can be solved by the mini-batch technique. Instead of sampling one single data per iteration, we suggest sampling a batch of data to make better approximations of the true gradient and hence speed up convergence rate and reduce communications. The size of mini-batch will determine the trade-off between computational and communication efficiencies. Interestingly, it is observed in simulations that when an appropriate batch-size is chosen, diffusion-AVRG with mini-batch can be more computation efficient while maintaining almost the same communication efficiency as exact diffusion. 
		
	}
	
	\textbf{Notation} Throughout this paper we use $\diag\{x_1,\cdots,x_N\}$ to denote a diagonal matrix consisting of diagonal entries ${x_1,\cdots,x_N}$,  and use $\col\{x_1,\cdots,x_N\}$ to denote a column vector formed by  stacking ${x_1,\cdots,x_N}$. For symmetric matrices $X$ and $Y$, the notation $X \le Y$ or $Y\ge X$ denotes $Y - X$ is positive semi-definite. For a vector $x$, the notation $x \succeq 0$ denotes that each element of $x$ is non-negative. For a matrix $X$, we let $\|X\|$ denote its $2$-induced norm (maximum singular value), and $\lambda(X)$ denote its eigenvalues. The notation $\mathds{1}_K = \col\{1,\cdots,1\} \in \RR^{K}$, and $0_K = \col\{0,\cdots,0\} \in \RR^{K}$. For a nonnegative diagonal matrix $\Lambda = \diag\{\lambda_1,\cdots,\lambda_K\}$, we let $\Lambda^{1/2} = \diag\{\lambda_1^{1/2},\cdots,\lambda_K^{1/2}\}$.
	
	\section{{\color{black}Two Key Components}}
	
	{\color{black}In this section we review two useful techniques that will be blended together to yield the diffusion-AVRG scheme. The first technique is the exact diffusion algorithm from \cite{yuan2017exact1,yuan2017exact2}, which is able to converge to the {\em exact} minimizers of the decentralized optimization problem \eqref{prob-emp-dist}. The second technique is the amortized variance-reduced (AVRG) algorithm proposed in our earlier work\cite{ying2017convergence}, which has balanced computations per iteration and was shown there to converge linearly under random reshuffling. Neither of the methods alone is sufficient to solve the multi-agent optimization problem \eqref{prob-emp-dist} in a decentralized and efficient manner. This is because exact diffusion is decentralized but not efficient for the current problem, while AVRG is efficient but not decentralized.}
	
	\subsection{Exact Diffusion Algorithm}
	
	Thus, consider again the aggregate optimization problem (\ref{prob-emp-dist}) over a strongly-connected network with $K$ nodes, where the $\{q_k\}$ are positive scalars. Each local risk $J_k(w)$ is a differentiable and convex cost function, and the global risk $J(w)$ is strongly convex. To implement the exact diffusion algorithm, we need to associate a combination matrix $A=[a_{\ell k}]_{\ell,k=1}^K$ with the network graph, where a positive weight $a_{\ell k}$ is used to scale data that flows from node $\ell$ to $k$ if both nodes happen to be neighbors; if nodes $\ell$ and $k$ are not neighbors, then we set $a_{\ell k} = 0$.  In this paper we assume $A$ is symmetric and doubly stochastic, i.e.,\vspace{-1mm}
	\eq{
		a_{\ell k}=a_{k\ell},\;\;A = A\tran\;\mbox{and}\;\; A \mathds{1}_K = \mathds{1}_K
	}
	where $\mathds{1}$ is a vector with all unit entries. Such combination matrices can be easily generated in a decentralized manner through the Laplacian  rule, maximum-degree rule, Metropolis rule or other rules (see, e.g., Table 14.1 in \cite{sayed2014adaptation}). We further introduce $\mu$ as the step-size parameter for all nodes, and let $\cN_k$ denote the set of neighbors of node $k$ (including node $k$ itself).
	
	\begin{table}[t]
		\noindent \HRule\\
		\noindent \textbf{\small Algorithm 1 (Exact diffusion strategy for each node $k$)} \vspace{-2mm}\\
		\HRule\\
		\noindent \small Let $\barA=(I_{N}+A)/2$ and $\overline{a}_{\ell k} = [\,\barA\,]_{\ell k}$. Initialize $w_{k,0}$ arbitrarily, and let $\psi_{k,0} = w_{k,0}$.\\
		\textbf{{ Repeat}} iteration $i=1,2,3\cdots$
		{
			\eq{
				\hspace{10mm} \psi_{k,i+1} &= w_{k,i} - \mu\, q_k \grad J_k(w_{k,i}), \hspace{6mm} \mbox{ (adaptation)} \label{ed-adapt}\\
				\phi_{k,i+1} &= \psi_{k,i+1} + w_{k,i}  - \psi_{k,i}, \hspace{8mm} \mbox{ (correction)} \label{ed-correct} \\
				w_{k,i+1} &= \sum_{\ell\in \cN_k} \overline{a}_{\ell k} \phi_{\ell,i+1}. \hspace{1.67cm} \mbox{ (combination)} \label{ed-comb}\\[-5.5mm]\nn
			}\vspace{-0mm}}
		\textbf{{End}}\\[-2mm]
		\HRule
		\vspace{-1mm}
	\end{table}
	
	The exact diffusion algorithm\cite{yuan2017exact1} is listed in (\ref{ed-adapt})--(\ref{ed-comb}).
	The subscript $k$ refers to the node while the subscript $i$ refers to the iteration.
	%
	It is observed  that there is no central node that performs global updates. Each node performs a local update (see equation \eqref{ed-adapt}) and then combines its iterate with information collected from the neighbors (see equation \eqref{ed-comb}). The correction step \eqref{ed-correct} is necessary to guarantee exact convergence. Indeed, it is proved in \cite{yuan2017exact2} that the local variables $w_{k,i}$ converge to the exact minimizer of problem \eqref{prob-emp-dist}, $w^\star$,  at a linear convergence rate under relatively mild conditions. However, note from \eqref{local-cost-func} that it is  expensive to calculate the gradient $\grad J_k(w)$ in step \eqref{ed-adapt}, especially when $N_k$ is large. In the proposed algorithm derived later, we will replace the true gradient $\grad J_k(w)$ in \eqref{ed-adapt} by an amortized variance-reduced gradient, denoted by $\widehat{\grad\hspace{-0.3mm} J_k}(\hspace{-0.3mm}\w_{k,i\hspace{-0.3mm}-\hspace{-0.3mm}1}\hspace{-0.3mm})$.
	
	{\color{black}
		\subsection{Amortized Variance-Reduced Gradient (AVRG) Algorithm}
		The AVRG construction \cite{ying2017convergence} is a centralized solution to optimization problem \eqref{prob-emp-into}. It belongs to the class of variance-reduced methods. There are mainly two families of variance-reduced stochastic algorithms to solve problems like \eqref{prob-emp-into}:  SVRG \cite{johnson2013accelerating,xiao2014proximal} and SAGA \cite{defazio2014saga,schmidt2013minimizing}. The SVRG solution employs two loops --- the true gradient is calculated in the outer loop and the variance-reduced stochastic gradient descent is performed within the inner loop. For this method, one disadvantage is that the inner loop can start only after the calculation of the true gradient is completed in the outer loop. This leads to an {\em unbalanced} gradient calculation. For large data sets,  the calculation of the true gradient can be time-consuming leading to significant idle time, which is not well-suited for decentralized solutions. More details are provided later in Sec.~\ref{sec-davrg-unbalanced}. In comparison, the SAGA solution has a single loop. However, it requires significant storage to estimate the true gradient, which is again prohibitive for effective decentralization on nodes or devices with limited memory.
		
		\begin{table}[t]
			\noindent \HRule\\
			\noindent \textbf{\small Algorithm 2 (AVRG strategy)} \vspace{-2mm}\\
			\HRule\\
			\noindent \small Initialize $\w_0^{0}$ arbitrarily; let $\g^0 = 0$, $\grad Q(\w_{0}^0;x_n) \leftarrow 0$ for $n\in\{1,2,\cdots,N\}$. \\
			\noindent \ \textbf{Repeat} epoch $t=0,1,2,\cdots$: \\
			{\color{white}{h}}\hspace{4mm} generate a random permutation function $\bsigma^t$ and set $\g^{t+1}=0$;\vspace{1mm}\\
			{\color{white}{h}}\hspace{4mm} \textbf{Repeat} iteration\ $i=0, 1,\cdots, N-1$:\vspace{-3mm} \\
			\begin{align}
				\hspace{-1cm}	\n^t_i &= \bsigma^t(i+1) \label{sample-2}\\
				\w^t_{i+1} &= \w^t_{i} - \mu \Big( \grad Q(\w^t_{i};x_{\n^t_i} ) - \grad Q(\w^t_{0};x_{\n^t_i} ) + \g^t\Big)\label{sgd-2}\\
				\g^{t+1} &\leftarrow \g^{t+1} + \frac{1}{N}\grad Q(\w^t_{i};x^t_{\n_i} ) \label{avrg-full-grad}
			\end{align}
			{\color{white}{h}}\hspace{4mm} \textbf{End}\\
			{\color{white}{h}}\hspace{4mm} set $\w_0^{t+1} = \w_N^t$; \\
			\vspace{3mm}\noindent \textbf{End} \vspace{-4mm}
			\\[-1mm] \HRule \\[-3mm]
		\end{table}
		
		These observations are the key drivers behind the introduction of the amortized variance-reduced gradient (AVRG) algorithm in \cite{ying2017convergence}: it avoids the disadvantages of both SVRG and SAGA for decentralization, and has been shown to converge at a linear rate to the true minimizer. AVRG is based on the idea of removing the outer loop from SVRG and amortizing the calculation of the true gradient within the inner loop evenly. To guarantee convergence, random reshuffling is  employed in each epoch. Under random reshuffling, the algorithm is run multiple times over the data where each run is indexed by $t$ and is referred to as an epoch. For each epoch $t$, a uniform random permutation function $\bsigma^t$ is generated and data are sampled according to it. AVRG is listed in Algorithm 2, which has balanced computation costs per iteration with the calculation of two gradients $\grad Q(\w^t_{i};x_{\n_i} )$ and $\grad Q(\w^t_{0};x_{\n_i} )$. Different from SVRG and SAGA, the stochastic gradient estimate $\widehat{\grad J}(\w_i^t) = \grad Q(\w^t_{i};x_{\n_i} ) - \grad Q(\w^t_{0};x_{\n_i} ) + \g^t$ is biased.
		However, it is explained in \cite{ying2017convergence} that $\bE\|\widehat{\grad J}(\w_i^t) - \grad J(\w_i^t)\|^2$ will approach  $0$ as epoch $t$ tends to infinity, which implies that 
		AVRG 
		is an asymptotic unbiased variance-reduced method. 
	}

	\section{Diffusion--AVRG Algorithm for Balanced Data Distributions}
	We now design a fully-decentralized algorithm to solve (\ref{prob-emp-dist}) by combining the exact diffusion strategy \eqref{ed-adapt}--\eqref{ed-comb} and the AVRG mechanism \eqref{sample-2}--\eqref{avrg-full-grad}.  We consider first the case in which all nodes store the same amount of local data, i.e., $N_1 = \cdots = N_K = \widebar{N}= N/K$.
	For this case, the cost function weights $\{q_k\}$ in problem \eqref{prob-emp-dist} are equal, $q_1 = \cdots = q_K = 1/K$,
	%
	and it makes no difference whether we keep these scaling weights or remove them from the aggregate cost. The proposed diffusion-AVRG algorithm to solve \eqref{prob-emp-dist} is listed in Algorithm 3 under Eqs. \eqref{sampling}--\eqref{ed-comb-2}.  
	Since each node has the same amount of local data samples, Algorithm 3 can be described in a convenient format involving epochs $t$ and an inner iterations index $i$ within each epoch. For each epoch or run $t$ over the data,  the original data is
	randomly reshuffled so that the sample of index $ i+1$ at agent $k$ becomes
	the sample of index $\n_{k,i}^t=\bm{\sigma}_k^t(i+1)$ in that run. Subsequently, at each inner iteration $i$, each node $k$ will first generate an amortized variance-reduced gradient $\widehat{\grad J}_k(\w_{k,i}^t)$ via \eqref{sampling}--\eqref{approximate-gradeint}, and then apply it into exact diffusion \eqref{ed-adapt-2}--\eqref{ed-comb-2} to update $\w_{k,i+1}^t$. Here, the notation $\w_{k,i}^t$ represents the estimate that agent $k$ has for $w^{\star}$ at iteration $i$ within epoch $t$. With each node combining information from neighbors, there is no central node in this algorithm. Moreover, unlike DSA\cite{mokhtari2016dsa}, this algorithm does not require extra memory to store gradient estimates. The linear convergence of diffusion-AVRG is established in the following theorem.
	
	\begin{table}[t]
		\noindent \HRule \vspace{0.1mm}\\
		\noindent \textbf{\small Algorithm 3 (diffusion-AVRG at node $k$ for balanced data)}\vspace{-2mm}\\
		\HRule\\
		\noindent \small \textbf{ Initialize} $\w^0_{k,0}$ arbitrarily; let $\bpsi^0_{k,0} = \w^0_{k,0}$, $\g_k^0=0$, and $\grad Q(\w_{0}^0;x_{k,n}) \leftarrow 0$, $\ 1\leq n \le \widebar{N}$, where $\widebar{N}=N/K$.\\
		\textbf{Repeat }epoch $t=0,1,2,\cdots$ \\
		{\color{white}{h}}\hspace{4mm} generate a random permutation function $\bsigma^t_k$ and set $\g_k^{t+1}=0$.
		
		{\color{white}{h}}\hspace{4mm} \textbf{Repeat } iteration $i=0,1,\cdots,\widebar{N}-1$:
		\eq{
			\n_{k,i}^t &= \bsigma^t_k(i+1),  \label{sampling}\\
			\widehat{\grad J}_k(\w^t_{k,i}) &= \grad Q(\w^t_{k,i};x_{k,\n_{k,i}^t} ) \hspace{-1mm}-\hspace{-1mm} \grad Q(\w^t_{k,0};x_{k,\n_{k,i}^t} ) + \g_k^t,   \label{sgd-3}\\
			\g_k^{t+1} &\leftarrow \g_k^{t+1} + \frac{1}{\widebar{N}}\grad Q(\w^t_{k,i};x_{k,\n_{k,i}^t} ),  \label{approximate-gradeint} \\
			&\hspace{-11mm}\mbox{update $\w^t_{k,i+1}$ with exact diffusion:}\nnb
			\bpsi^t_{k,i+1} &= \w^t_{k,i} - \mu \widehat{\grad J}_k(\w^t_{k,i}), \label{ed-adapt-2}\\
			\bphi^t_{k,i+1} &= \bpsi^t_{k,i+1} + \w^t_{k,i}  - \bpsi^t_{k,i},\\
			\w^t_{k,i+1} &= \sum_{\ell\in \cN_k} \overline{a}_{\ell k} \bphi^t_{\ell,i+1}. \label{ed-comb-2}
			\nn\\[-7mm]
		}
		{\color{white}{hh}}\hspace{4mm} \textbf{End} \\
		{\color{white}{hh}}\hspace{4mm} set $\w^{t+1}_{k,0} = \w^{t}_{k,\widebar{N}}$ and $\bpsi^{t+1}_{k,0} = \bpsi^{t}_{k,\widebar{N}}$ \\
		\noindent \ \textbf{End}
		\\[-2mm] \HRule\vspace{-4mm}
	\end{table}
	
	\begin{theorem}[\sc Linear Convergence]\label{tho-balanced-a23} Under Assumption \ref{ass}, if the step-size $\mu$ satisfies
		\eq{\label{23hsdbs8-1}
			{\color{black}\mu \le C\left(\frac{\nu (1-\lambda)}{\delta^2 \widebar{N}}\right),}
		}
		then, for any $k\in \{1,2,\cdots, K\}$, it holds that\vspace{-1mm}
		\eq{\label{zljshdjdhdudsdsacsd-0}
			\bE\|\w_{k,0}^{t+1} - w^\star\|^2  \le D \rho^t,
		}
		where
		\eq{
			\rho = \frac{1- \frac{\overline{N}}{8}a \mu \nu}{1- 8 b \mu^3\delta^4 \widebar{N}^3 / \nu} < 1.
		}
		The constants $C, D, a, b$ are positive constants independent of $\widebar{N}$, $\nu$ and $\delta$; they are defined in the appendices. {\color{black}The constant $\lambda = \lambda_2(A) < 1$ is the second largest eigenvalue of the combination matrix $A$.} 
		\qd
	\end{theorem}
	
	\noindent The detailed proof is given in Appendix \ref{app-proof-main-theorem}, along with supporting appendices {in the supplemental material. We summarize the main proof idea as follows.} 
	
	\noindent {\textbf{Sketch of the Proof.} We start by transforming the exact diffusion recursions \eqref{ed-adapt-2}--\eqref{ed-comb-2} into an equivalent linear error dynamics driven by perturbations due to  gradient noise (see Lemma \ref{lm-mse-recursion}): 
		\eq{\label{sketch-proof-main-recursion}
			\hspace{-3mm}
			\ba{c}
			\bE\|\bar{\sxb}^t_{i+1}\|^2\hspace{-4mm} \\
			\bE\|\check{\sxb}^t_{i+1}\|^2 \hspace{-4mm}
			\ea
			& \hspace{-0.8mm}\preceq \hspace{-0.8mm}
			A 
			\ba{c}
			\hspace{-2mm}\bE\|\bar{\sxb}^t_{i}\|^2\hspace{-2mm} \vspace{0.5mm}\\
			\hspace{-2mm} \ \ \bE\|\check{\sxb}^t_{i}\|^2 \hspace{-2mm}
			\ea \hspace{-0.8mm}+\hspace{-0.8mm}
			\ba{c}
			\frac{2\mu}{\nu} \bE\|\s(\swb^t_{i})\|^2 \vspace{0.5mm}\\
			c \mu^2 \bE\|\s(\swb^t_{i})\|^2
			\ea,
		}
		where $\bar{\sxb}^t_{i}$ and $\check{\sxb}^t_{i}$ are auxiliary variables with the property:
		\eq{
			\bE\|\w_{k,i}^t - w^\star\|^2 \le C(\bE\|\bar{\sxb}^t_{i}\|^2 + \bE\|\check{\sxb}^t_{i}\|^2)
		}
		and $C$ is some positive constant. As a result, the proof of linear convergence of $\bE\|\w_{k,i}^t - w^\star\|^2$ reduces to the linear convergence of $\bE\|\bar{\sxb}^t_{i}\|^2$ and $\bE\|\check{\sxb}^t_{i}\|^2$, which can be studied via the linear recursion \eqref{sketch-proof-main-recursion}. The matrix $A$ appearing in \eqref{sketch-proof-main-recursion} also has useful  properties. It can be proved that when the step-size $\mu$ is sufficiently small, it holds that $\rho(A)<1$ where $\rho(\cdot)$ represents the spectrum radius. The term $\s(\swb^t_{i})$ in \eqref{sketch-proof-main-recursion} is the stochastic gradient noise introduced by the gradient constructions \eqref{sampling}--\eqref{approximate-gradeint} and $c$ is a constant.
		
		A second crucial step is to bound gradient noise $\bE\|\s(\swb^t_{i})\|^2$. It is proved in Lemma \ref{lm-gradient-noise} that
		\eq{\label{zx23b9-sketch}
			& \bE\|\s(\swb_i^t)\|^2\nnb
			&\le 6 b \delta^2 \bE\|\bar{\sxb}^t_{i} - \bar{\sxb}^t_{0}\|^2 + 12 b \delta^2 \bE\|\check{\sxb}^t_{i}\|^2 + 18 b \delta^2 \bE\|\check{\sxb}^t_{0}\|^2 \nnb
			&\quad +\hspace{-0.5mm} \frac{3b\delta^2}{\widebar{N}}\sum_{j=0}^{\widebar{N}-1}\bE\|\bar{\sxb}^{t-1}_{j} \hspace{-0.8mm}-\hspace{-0.8mm} \bar{\sxb}^{t-1}_{\widebar{N}}\|^2 \hspace{-1mm}+\hspace{-1mm} \frac{6b\delta^2}{\widebar{N}}\sum_{j=0}^{\widebar{N}-1}\bE\|\check{\sxb}^{t-1}_{j}\|^2
		}
		where $b$ is a constant. It is observed in \eqref{zx23b9-sketch} that multiple non-trivial quantities such as inner difference in current epoch $\bE\|\bar{\sxb}^t_{i} - \bar{\sxb}^t_{0}\|^2$, and inner difference in previous epoch $\bE\|\bar{\sxb}^{t-1}_{j} \hspace{-0.8mm}-\hspace{-0.8mm}\bar{\sxb}^{t-1}_{\widebar{N}}\|^2$ arise. By establishing some supporting inequalities to bound these quantities (see Lemmas \ref{lm-check-x}--\ref{lm-inner-recursion}) and combing with \eqref{sketch-proof-main-recursion}, we finally introduce an energy function involving $\bE\|\bar{\sxb}^t_{i}\|^2$ and $\bE\|\check{\sxb}^t_{i}\|^2$ and show that it decays exponentially fast (Lemma \ref{tho-balanced}), which concludes the proof. \qd} 
	

\section{Diffusion--AVRG Algorithm for Unbalanced Data Distributions}\label{sec-davrg-unbalanced}

When the size of the data collected at the nodes may vary drastically, some challenges arise.
For example, assume we select $\widehat{N}=\max_k\{N_k\}$ as the epoch size for all nodes. When node $k$ with a smaller $N_k$ finishes its epoch,  it will have to stop and wait for the other nodes to finish their epochs. Such an implementation is inefficient because nodes will be idle while they could be assisting in improving the convergence performance.

\begin{figure*}[!h]
	\centering
	\includegraphics[scale=0.42]{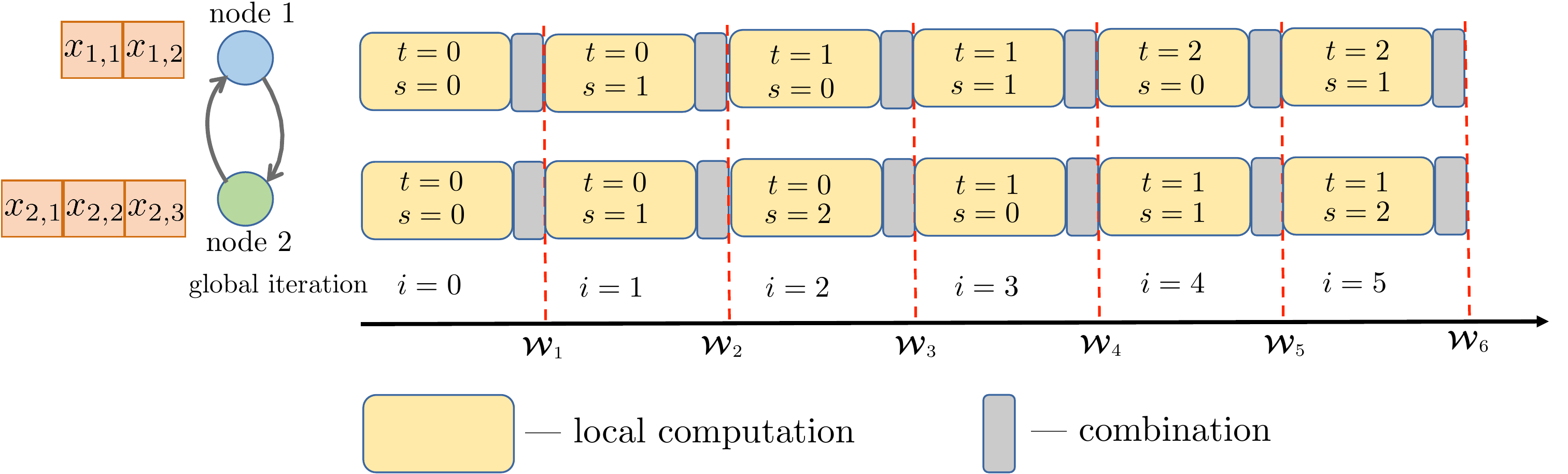}\vspace{-1mm}
	\centering
	\caption{\small{Illustration of the operation of diffusion-AVRG for a two-node network.}}\vspace{-3mm}\label{fig-d-avrg}
\end{figure*}

We instead assume that nodes will continue updating without any idle time. If a particular node $k$ finishes running over all its data samples during an epoch, it will then continue its next epoch right away. In this way, there is no need to introduce a uniform epoch. We list the method in Algorithm 4; this listing includes the case of balanced data as a special case. In other words, we have a single diffusion-AVRG algorithm. We are describing it in two formats (Algorithms 3 and 4) for ease of exposition so that readers can appreciate the simplifications that occur in the balanced data case.

\begin{table}[b]
	\noindent \HRule \vspace{1mm}\\
	\noindent \textbf{\small Algorithm 4 (diffusion-AVRG at node $k$ for unbalanced data)}\vspace{-1mm}\\
	\HRule\\
	\noindent \small \textbf{ Initialize} $\w_{k,0}$ arbitrarily; let $q_k = N_k/N$, $\bpsi_{k,0} = \w_{k,0}$, $\g_k^0=0$, and $\grad Q(\btheta^0_{k,0};x_{k,n} )\leftarrow 0,\; 1\le n\le N_k$\\
	\textbf{Repeat } $i=0,1,2,\cdots$ \\
	{\color{white}{h}}calculate $t$ and $s$ such that $i \hspace{-0.8mm}=\hspace{-0.8mm} t {N}_k \hspace{-0.8mm}+\hspace{-0.8mm} s$, where $t \in \mathbb{Z}_{+}$ and $s \hspace{-0.8mm}=\hspace{-0.8mm} {\color{white}j}\mbox{mod}(i, {N}_k)$;
	
	
	{\color{white}{h}}\textbf{If} $s = 0$: \\
	{\color{white}{hh}}generate a random permutation$\bsigma^t_k$; let $\g_k^{t+1}\hspace{-0.5mm}=\hspace{-0.5mm}0, \ \btheta^t_{k,0}=\w_{k,i}$;\\
	{\color{white}{h}}\textbf{End}\\
	{\color{white}{h}}generate the local stochastic gradient:
	\eq{
		\n_{s}^t &= \bsigma^t_k(s+1),  \label{avrg-s-1}\\
		\widehat{\grad J}_k(\w_{k,i}) &= \grad Q(\w_{k,i};x_{k,\n_s^t} ) - \grad Q(\btheta^t_{k,0};x_{k,\n_s^t} ) + \g_k^t,  \label{avrg-sgd-1}\\
		\g_k^{t+1} &\leftarrow \g_k^{t+1} + \frac{1}{N_k}\grad Q(\w_{k,i};x_{k,\n_s^t} ), \label{avrg-full-1}\\
		&\hspace{-1.55cm}\mbox{update $\w_{k,i+1}$ with exact diffusion:} \nnb
		\bpsi_{k,i+1} &= \w_{k,i} - \mu q_k \widehat{\grad J}_k(\w_{k,i}), \hspace{19mm} \label{ed-adapt-3-1}\\
		\bphi_{k,i+1} &= \bpsi_{k,i+1} + \w_{k,i}  - \bpsi_{k,i}, \label{ed-corr-3-1}\\
		\w_{k,i+1} &= \sum_{\ell\in \cN_k} \overline{a}_{\ell k} \bphi_{\ell,i+1}. \label{ed-comb-3-1}
	}
	\noindent \vspace{-2mm} \ \textbf{End}\\
	\HRule \vspace{-5mm}
\end{table}

In Algorithm 4, at each iteration $i$, each node $k$ will update its $\w_{k,i}$ to $\w_{k,i+1}$ by exact diffusion  \eqref{ed-adapt-3-1}--\eqref{ed-comb-3-1} with stochastic gradient. Notice that $q_k$ has to be used to scale the  step-size in \eqref{ed-adapt-3-1} because of the spatially unbalanced data distribution. To generate the local stochastic gradient $\widehat{\grad J}_k(\w_{k,i})$, node $k$ will transform the {\em global} iteration index $i$ to its own {\em local} epoch index $t$ and {\em local} inner iteration $s$. With $t$ and $s$ determined, node $k$ is able to generate $\widehat{\grad J}_k(\w_{k,i})$ with the AVRG recursions \eqref{avrg-s-1}--\eqref{avrg-full-1}. Note that $t,s,\bsigma_k^{t},\btheta_{k,0}^t, \n_s^t$ are all local variables hidden in node $k$ to help generate the local stochastic gradient $\widehat{\grad J}_k(\w_{k,i})$ and do not appear in exact diffusion \eqref{ed-adapt-3-1}--\eqref{ed-comb-3-1}. Steps \eqref{avrg-s-1}--\eqref{ed-corr-3-1} are all local update operations within each node while step \eqref{ed-comb-3-1} needs communication with neighbors. It is worth noting that the local update \eqref{avrg-s-1}--\eqref{ed-corr-3-1} for each node $k$ at each iteration requires the same amount of computations no matter how different the sample sizes $\{N_k\}$ are. This balanced computation feature guarantees the efficiency of diffusion-AVRG and reduces waiting time. Figure \ref{fig-d-avrg} illustrates the operation of Algorithm 4 for a two-node network with $N_1 = 2$ and $N_2 =3$. That is, the first node collects two samples while the second node collects three samples. For each iteration index $i$, the nodes will determine the local values for their indices $t$ and $s$. These indices are used to generate the local variance-reduced gradients $\widehat{\grad J}_k(\w_{k,i})$. Once node $k$ finishes its own local epoch $t$, it will start its next epoch $t+1$ right away. Observe that the local computations has similar widths because each node has a balanced computation cost per iteration. Note that $\swb_i = [\w_{1,i}; \w_{2,i}]$ in Figure  \ref{fig-d-avrg}.\vspace{-2mm}

\subsection{Comparison with Decentralized SVRG}\label{sec:compare-svrg}

\begin{figure*}
	\centering
	\includegraphics[scale=0.42]{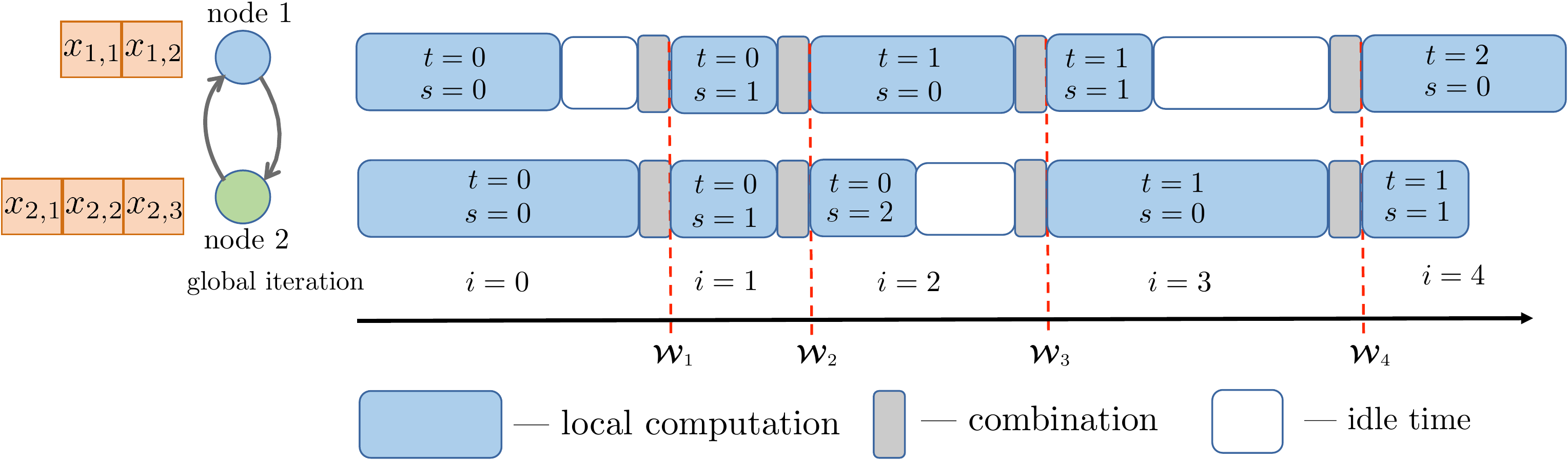}\vspace{-2mm}
	\caption{\small{Illustration of what would go wrong if one attempts a diffusion-SVRG implementation for a two-node network, and why diffusion-AVRG is the recommended implementation.}\vspace{-3mm}}\label{fig-d-svrg}
	\vspace{5mm}
\end{figure*}

AVRG is not the only variance-reduced algorithm that can be combined with exact diffusion. In fact, SVRG is another alternative to save memory compared to SAGA. SVRG has two loops of calculation: it needs to complete the calculation of the true gradient before starting the inner loop. Such two-loop structures are not suitable for decentralized setting, especially when data can be distributed unevenly. To illustrate this fact assume, for the sake of argument, that we combine exact diffusion with SVRG to obtain a diffusion-SVRG variant, which we list in Algorithm 5. Similar to diffusion-AVRG, each node $k$ will transform the global iteration index $i$ into a local epoch index $t$ and a local inner iteration $s$, which are then used to generate $\widehat{\grad J}(\w_{k,i})$ through SVRG. At the very beginning of each local epoch $t$, a true local gradient has to be calculated in advance; this step causes a pause before the update of $\bphi_{k,i+1}$. Now since the neighbors of node $k$ will be waiting for $\bphi_{k,i+1}$ in order to update their own $\w_{\ell, i+1}$, the pause by node $k$ will cause all its neighbors to wait. These waits reduce the efficiency of this decentralized implementation, which explains why the earlier diffusion-AVRG algorithm is preferred. Fig. \ref{fig-d-svrg} illustrates the diffusion-SVRG strategy with $N_1=2$ and $N_2=3$.  Comparing Figs. \ref{fig-d-avrg} and \ref{fig-d-svrg}, the balanced calculation resulting from AVRG effectively reduces idle times and enhances the efficiency of the decentralized implementation.
\begin{table}[h!]
	\noindent \HRule \vspace{1mm}\\
	\noindent \textbf{\small Algorithm 5 (diffusion-SVRG at node $k$ for unbalanced data)} 
	\small
	\vspace{-1mm}\\
	\HRule\\
	\noindent \textbf{ Initialize} $\w_{k,0}$ arbitrarily; let $q_k = N_k/N$, $\bpsi_{k,0} = \w_{k,0}$\\
	\textbf{Repeat } $i=0,1,2,\cdots$ \\
	{\color{white}{h}} calculate $t$ and $s$ such that $i \hspace{-0.8mm}=\hspace{-0.8mm} t {N}_k \hspace{-0.8mm}+\hspace{-0.8mm} s$, where $t \in \mathbb{Z}_{+}$ and $s \hspace{-0.8mm}=\hspace{-0.8mm}$ ${\color{white}jj}\mbox{mod}(i, {N}_k)$;  \vspace{1mm}\\
	{\color{white}{h}}\hspace{0.5mm} \textbf{If} $s = 0$: \\
	{\color{white}{h}}\hspace{2mm} generate a random permutation function $\bsigma^t_k$, set $\btheta^t_{k,0}=\w_{k,i}$\\ {\color{white}{h}}\hspace{2mm} and compute the full gradient: \vspace{-4mm}\\
	\eq{
		\g_k^{t}=\frac{1}{N_k}\sum_{n=1}^{N_k} \nabla Q(\btheta^t_{k,0};x_{k,n}), 
	}
	{\color{white}{h}}\hspace{0.5mm} \textbf{End}\\
	\mbox{\hspace{3.2mm}generate the local stochastic gradient}:
	\eq{
		\n_s^t &= \bsigma_k^t(s+1),  \\
		\widehat{\grad J}_k(\w_{k,i}) &= \grad Q(\w_{k,i};x_{k,\n_s^t} ) - \grad Q(\btheta^t_{k,0};x_{k,\n_s^t} ) + \g_k^t,  
	}
	{\color{white}{h}}\hspace{0mm} \ update $\w_{k,i+1}$ with exact diffusion: \vspace{-4mm}\\
	\eq{
		\bpsi_{k,i+1} &= \w_{k,i} - \mu q_k \widehat{\grad J}_k(\w_{k,i}), \\ 
		\bphi_{k,i+1} &= \bpsi_{k,i+1} + \w_{k,i}  - \bpsi_{k,i},  \\
		\w_{k,i+1} &= \sum_{\ell\in \cN_k} \overline{a}_{\ell k} \bphi_{\ell,i+1}. 
	}
	\noindent \ \textbf{End}
	\\ \HRule\vspace{-3mm}
\end{table}

\section{diffusion-AVRG with Mini-batch Strategy}\label{sec. minibatch}
Compared to exact diffusion \cite{yuan2017exact1,yuan2017exact2}, diffusion-AVRG allows each agent to sample one gradient at each iteration instead of calculating the true gradient with $N_k$ data. This property enables diffusion-AVRG to be more computation efficient than exact diffusion. It is observed in Figs. \ref{mnist-batch} and \ref{rcv-batch} from Section \ref{sec. simulation} that in order to reach the same accuracy, diffusion-AVRG needs less gradient calculation than exact diffusion. 

However, such computational advantage comes with extra communication costs. In the exact diffusion method listed in Algorithm 1, it is seen that agent $k$ will communicate after calculating its true gradient $\grad J(w) = \frac{1}{N_k}\sum_{n=1}^{N_k}Q(w;x_{k,n})$. But in the diffusion-AVRG listed in Algorithms 2 and 3, each agent will communicate after calculating only one stochastic gradient. Intuitively, in order to reach the same accuracy, diffusion-AVRG needs more iterations than exact diffusion, which results in more communications. The communication comparison for diffusion-AVRG and exact diffusion are also shown in Figs. \ref{mnist-batch} and \ref{rcv-batch} in Section \ref{sec. simulation}.

In this section we introduce the mini-batch strategy to balance the computation and communication of diffusion-AVRG. For simplicity, we consider the situation where all local data size $N_k$ are equal to $\widebar{N}$, but the strategy can be extended to handle the spatially unbalanced data distribution case.
Let the batch size be $B$, and the number of batches $L \define \widebar{N}/B$. The local data in agent $k$ can be partitioned as
\eq{
	\{x_{k,n}\}_{n=1}^{\widebar{N}} \hspace{-1mm}=\hspace{-1mm} \left\{ \hspace{-0.5mm} \{x_{k,n}^{(1)}\}_{n=1}^B, \{x_{k,n}^{(2)}\}_{n=1}^B,  \cdots, \{x_{k,n}^{(L)}\}_{n=1}^B \hspace{-0.5mm} \right\},
}
where the superscript $(\ell)$ indicates the $\ell$-th mini-batch. In addition, the local cost function $J_k(w)$ can be rewritten as
\eq{\label{J_k-minibatch}
	J_k(w) &= \frac{1}{\widebar{N}}\sum_{n=1}^{\widebar{N}}Q(w;x_{k,n}) = \frac{B}{\widebar{N}}\sum_{\ell=1}^{L}\frac{1}{B}\sum_{n=1}^{B}Q(w;x_{k,n}^{(\ell)}) \nnb
	&= \frac{1}{L}\sum_{\ell=1}^{L}Q_k^{(\ell)}(w),
}
where the last equality holds because $L = \widebar{N}/B$ and 
\eq{\label{cost-function-minibatch}
	Q_k^{(\ell)}(w) \define \frac{1}{B}\sum_{n=1}^{B}Q(w;x_{k,n}^{(\ell)})
}
is defined as the cost function over the $\ell$-th batch in agent $k$. Note that the mini-batch formulations \eqref{J_k-minibatch} and \eqref{cost-function-minibatch} are the generalization of cost function \eqref{local-cost-func}. When $B = 1$, formulations \eqref{J_k-minibatch} and \eqref{cost-function-minibatch} will reduce to \eqref{local-cost-func}. Moreover, it is easy to prove that $\{Q^{\ell}_k(w)\}_{k=1,\ell=1}^{K,L}$ satisfy Assumption \ref{ass}.

Since the mini-batch formulations \eqref{J_k-minibatch} and \eqref{cost-function-minibatch} fall into the form of problem \eqref{local-cost-func} and \eqref{prob-emp-dist}, we can directly extend Algorithm 3 to the mini-batch version with the convergence guarantee. The only difference is for each iteration, a batch, rather than a sample will be picked up, and then length of batches is $L$ rather than $\widebar{N}$. We also list the mini-batch algorithm in Algorithm 6.

\begin{table}[t]
	\noindent \HRule \vspace{0.1mm}\\
	\noindent \textbf{\small Algorithm 6 (diffusion-AVRG with mini-batch at node $k$)}\vspace{-2mm}\\
	\HRule\\
	\noindent \small \textbf{ Initialize} $\w^0_{k,0}$ arbitrarily; let $\bpsi^0_{k,0} = \w^0_{k,0}$, $\g_k^0=0$; equally partition the data into $L$ batches, and each batch has size $B$. Set $\grad Q_{k}^{(\ell)}(\w_{0}^0) \leftarrow 0$, $\ 1\leq \ell \le \widebar{L}$\\
	\textbf{Repeat }epoch $t=0,1,2,\cdots$ \\
	{\color{white}{h}}\hspace{4mm} generate a random permutation function $\bsigma^t_k$ and set $\g_k^{t+1}=0$.
	
	{\color{white}{h}}\hspace{4mm} \textbf{Repeat } iteration $i=0,1,\cdots,L-1$:
	\eq{
		{\boldsymbol \ell}_{k,i}^t &= \bsigma^t_k(i+1),  \label{sampling-b}\\
		\widehat{\grad J}_k(\w^t_{k,i}) &= \grad Q_k^{({\boldsymbol \ell}_{k,i}^t)}(\w^t_{k,i}) \hspace{-1mm}-\hspace{-1mm} \grad Q_k^{({\boldsymbol \ell}_{k,i}^t)}(\w^t_{k,0}) + \g_k^t,   \label{sgd-3-b}\\
		\g_k^{t+1} &\leftarrow \g_k^{t+1} + \frac{1}{L} \grad Q_k^{({\boldsymbol \ell}_{k,i}^t)}(\w^t_{k,i}),  \label{approximate-gradeint-b} \\
		&\hspace{-11mm}\mbox{update $\w^t_{k,i+1}$ with exact diffusion:}\nnb
		\bpsi^t_{k,i+1} &= \w^t_{k,i} - \mu \widehat{\grad J}_k(\w^t_{k,i}), \label{ed-adapt-2-b}\\
		\bphi^t_{k,i+1} &= \bpsi^t_{k,i+1} + \w^t_{k,i}  - \bpsi^t_{k,i},\\
		\w^t_{k,i+1} &= \sum_{\ell\in \cN_k} \overline{a}_{\ell k} \bphi^t_{\ell,i+1}. \label{ed-comb-2-b}
		\nn\\[-7mm]
	}
	{\color{white}{hh}}\hspace{4mm} \textbf{End} \\
	{\color{white}{hh}}\hspace{4mm} set $\w^{t+1}_{k,0} = \w^{t}_{k,L}$ and $\bpsi^{t+1}_{k,0} = \bpsi^{t}_{k,L}$ \\
	\noindent \ \textbf{End}
	\\[-2mm] \HRule\vspace{-4mm}
\end{table}

%

Diffusion-AVRG with mini-batch stands in the middle point between standard diffusion-AVRG and exact diffusion. For each iteration, Algorithm 6 samples $B$ gradients, rather than $1$ gradient or $\widebar{N}$ gradients, and then communicates. The size of $B$ will determine the computation and communication efficiency, and there is a trade-off between computation and communication. When given the actual cost in real-world applications, we can determine the Pareto optimal for the batch-size. In our simulation shown in Section \ref{sec. simulation}, when best batch-size is chosen, diffusion-AVRG with mini-batch can be much more computation efficient while maintaining almost the same communication efficiency with exact diffusion.

{
	\section{Proximal diffusion-AVRG}\label{sec.prox}
	In this section we extend the diffusion-AVRG algorithm to handle non-smooth cost functions. Thus, consider now problems of the form:
	\eq{\label{prob-prox}
		\argmin_{w\in \RR^M}\ J(w) + R(w), \ \mbox{where}\ J(w)=\sum_{k=1}^{K}q_k J_k(w) 
	}
	where $J_k(w)$ is defined in \eqref{local-cost-func}, and $R(w)$ is a convex but possibly non-differentiable regularization term. The assumptions over $J(w)$ remain the same, while we assume that $R(w)$ is proximable, i.e., the proximal problem
	\eq{
		\hspace{-1mm}w^+ \hspace{-0.5mm}=\hspace{-0.5mm} \prox_{\mu R}(w^-) \hspace{-0.5mm}=\hspace{-0.5mm} \argmin_{w}\Big\{\hspace{-0.5mm}R(w) \hspace{-0.5mm}+\hspace{-0.5mm} \frac{1}{2\mu}\|w - w^-\|^2 \hspace{-0.5mm}\Big\}
	}
	has a closed-form solution. Without loss of generality, we consider the situation where all local data sizes $N_k$ are equal to $\widebar{N}$. For this situation it holds that $q_k = 1/K$ for $k=1,\cdots, K$. In the following, we first design a deterministic distributed algorithm to solve problem \eqref{prob-prox}, and then extend it to the stochastic setting with the help of AVRG.
	
	We let $w_k \in \RR^M$ be a local estimate of variable $w$ in agent $k$. In the following we introduce some notations.
	\eq{
		\sw &\define \mathrm{col}\{w_1,\cdots, w_K\}\in \RR^{MK} \\	
		\cA &\define A \otimes I_M \in \RR^{MK\times MK} \\
		\widebar{\cA} &\define \widebar{A} \otimes I_M \in \RR^{MK\times MK} \\
		\cV &\define V \otimes I_M \in \RR^{MK\times MK} \\
		\cJ(\sw) &\define \sum_{k=1}^{K} J_k(w_k),\quad  \cR(\sw) \define \sum_{k=1}^{K} R(w_k)  \label{defi-R}
	}
	where $\bar{A} = (A+I_K)/2$ and ``$\otimes$'' indicates the Kronecker product. Since $A$ is symmetric and doubly stochastic, the matrix $I-A$ is positive semidefinite and it can be decomposed as $(I-A)/2 = U\Sigma U\tran$. The matrix $V$ is defined as $V = U\Sigma^{1/2}U\tran$ and it holds that $V^2 = (I-A)/2$ and $\mbox{null}(V) = \mbox{span}(\mathds{1}_K)$ \cite{yuan2017exact1}. To solve problem \eqref{prob-prox}, we propose the following primal-dual algorithm
	\eq{
		\begin{cases}
			\sz_i &\hspace{-3mm}= \widebar{\cA}(\sw_{i-1} - \mu \grad \cJ(\sw_{i-1})) - \cV \sy_{i-1}, \\
			\sy_i &\hspace{-3mm}= \sy_{i-1} + \cV \sz_i, \\
			\sw_i &\hspace{-3mm}= \prox_{\mu \cR}(\sz_i). \label{prox-update}
		\end{cases}
	}
	where $\sy\in \RR^{MK}$ is the dual variable. We claim the fixed point of the above recursions are solutions to problem \eqref{prob-prox}. To see that, we assume $(\sw^\star, \sy^\star, \sz^\star)$ are fixed points of recursion \eqref{prox-update}, and therefore it holds that
	\eq{
		\begin{cases}
			\sz^\star &\hspace{-3mm}= \widebar{\cA}(\sw^\star - \mu \grad \cJ(\sw^\star)) - \cV \sy^\star, \\
			\sy^\star &\hspace{-3mm}= \sy^\star + \cV \sz^\star, \\
			\sw^\star &\hspace{-3mm}= \prox_{\mu \cR}(\sz^\star). \label{prox-update-fixed-point}
		\end{cases}
	}
	From the second recursion in \eqref{prox-update-fixed-point}, we have
	\eq{
		\cV \sz^\star = 0 \Longleftrightarrow z_1^\star = \cdots = z_K^\star = z^\star	 \label{z-star}
	}
	where $z_k^\star \in \RR^M$ is the $k$-th block of vector $\sz^\star$. The ``$\Longleftrightarrow$'' sign holds because of the fact that $\mbox{null}(V) = \mbox{span}(\mathds{1}_K)$. Next, from the third equation of \eqref{prox-update-fixed-point} and the  definition of $\cR(\sw)$ in \eqref{defi-R}, we have
	\eq{
		w_k^\star = \prox_{\mu R}(z_k^\star) \overset{\eqref{z-star}}{=} \prox_{\mu R}(z^\star),
	}
	which implies that $w_1^\star = \cdots = w_K^\star = w^\star$ and the optimality condition
	\eq{\label{opt-cond-prox}
		0 \in \mu\, \partial R(w^\star) + (w^\star - z^\star).	
	}
	We further multiply $\frac{1}{K}(\mathds{1}\tran \otimes I_M)$ to both sides of the first equation in \eqref{prox-update-fixed-point} from the left to get
	\eq{\label{xcn238sd8}
		z^\star = w^\star - \frac{\mu}{K} \sum_{k=1}^{K} \grad J_k(w^\star)
	}
	where we also used the fact that $\mathds{1}\tran A = \mathds{1}\tran$, and $\mathds{1}\tran V = 0$. By substituting \eqref{xcn238sd8} into \eqref{opt-cond-prox}, we get
	\eq{
		0 \in \partial R(w^\star) + \frac{1}{K}\sum_{k=1}^{K}\grad J_k(w^\star),
	}
	which indicates that $w^\star$ is the optimal solution to problem \eqref{prob-prox}. Therefore, if the proposed recursion \eqref{prox-update} is convergent, its limiting point is the optimal solution to problem \eqref{prob-prox}. 
	
	Recursion \eqref{prox-update} can be rewritten in a more elegant manner. By eliminating the dual variable $\sy$ from the recursion, we get  
	\eq{
		\begin{cases}
			\sz_i &\hspace{-3mm}= \widebar{\cA}\Big(\sz_{i-1} \hspace{-0.8mm}+\hspace{-0.8mm} \sw_{i-1} \hspace{-0.8mm}-\hspace{-0.8mm} \sw_{i-2} \hspace{-0.8mm}-\hspace{-0.8mm} \mu \grad \cJ(\sw_{i-1}) \hspace{-0.8mm}+\hspace{-0.8mm} \mu \grad \cJ(\sw_{i-2})\Big), \\
			\sw_i &\hspace{-3mm}= \prox_{\mu \cR}(\sz_i),
		\end{cases}
	}
	which can be further written in a distributed manner:
	\eq{\label{prox-ed}
		\begin{cases}
			\psi_{k,i} &\hspace{-3mm}= w_{k,i-1} - \mu \grad J_k(w_{k,i-1}), \hspace{5mm} \\
			\phi_{k,i} &\hspace{-3mm}= \psi_{k,i} + z_{k,i-1}  - \psi_{k,i-1}, \hspace{8mm}  \\
			z_{k,i} &\hspace{-3mm}= \sum_{\ell\in \cN_k} \overline{a}_{\ell k} \phi_{\ell,i}, \hspace{2.32cm}  \\
			w_{k,i} &\hspace{-3mm}= \argmin_{w}\{R(w) + \frac{1}{2\mu}\|w - z_{k,i}\|^2\}. 
		\end{cases}
	}
	Recursion \eqref{prox-ed} is almost the same as the exact diffusion in \cite{yuan2017exact1} except for the additional proximal step. It is observed when $R(w)=0$, the recursion \eqref{prox-ed} reduces to the exact diffusion in \cite{yuan2017exact1}. 
	
	Using the proximal exact diffusion recursion \eqref{prox-ed}, we can easily extend it to a variance-reduced stochastic algorithm by replacing the true gradient with a stochastic one generated by the AVRG method. We list the prox-diffusion-AVRG method in Algorithm 7. Due to space limitations, we leave a formal verification of the convergence of Algorithm 7 for future work. Instead, we illustrate its convergence behavior with simulations over real datasets in Sec. \ref{sec. simulation}.
}

\begin{table}[t]
	\noindent \HRule \vspace{0.1mm}\\
	\noindent \textbf{\small Algorithm 7 (Prox-diffusion-AVRG at node $k$ for balanced data)}\vspace{-2mm}\\
	\HRule\\
	\noindent \small \textbf{ Initialize} $\w^0_{k,0}$ arbitrarily; let $\bpsi^0_{k,0} = \z^0_{k,0}$, $\g_k^0=0$, and $\grad Q(\w_{0}^0;x_{k,n}) \leftarrow 0$, $\ 1\leq n \le \widebar{N}$, where $\widebar{N}=N/K$.\\
	\textbf{Repeat }epoch $t=0,1,2,\cdots$ \\
	{\color{white}{h}}\hspace{4mm} generate a random permutation function $\bsigma^t_k$ and set $\g_k^{t+1}=0$.
	
	{\color{white}{h}}\hspace{4mm} \textbf{Repeat } iteration $i=0,1,\cdots,\widebar{N}-1$:
	\eq{
		\n_{k,i}^t &= \bsigma^t_k(i+1),  \label{sampling-prox}\\
		\widehat{\grad J}_k(\w^t_{k,i}) &= \grad Q(\w^t_{k,i};x_{k,\n_{k,i}^t} ) \hspace{-1mm}-\hspace{-1mm} \grad Q(\w^t_{k,0};x_{k,\n_{k,i}^t} ) + \g_k^t,   \label{sgd-3-prox}\\
		\g_k^{t+1} &\leftarrow \g_k^{t+1} + \frac{1}{\widebar{N}}\grad Q(\w^t_{k,i};x_{k,\n_{k,i}^t} ),  \label{approximate-gradeint-prox} \\
		&\hspace{-11mm}\mbox{update $\w^t_{k,i+1}$ with exact diffusion:}\nnb
		\bpsi^t_{k,i+1} &= \w^t_{k,i} - \mu \widehat{\grad J}_k(\w^t_{k,i}), \label{ed-adapt-2-prox}\\
		\bphi^t_{k,i+1} &= \bpsi^t_{k,i+1} + \z^t_{k,i}  - \bpsi^t_{k,i},\\
		\z^t_{k,i+1} &= \sum_{\ell\in \cN_k} \overline{a}_{\ell k} \bphi^t_{\ell,i+1}, \label{ed-comb-2-prox} \\
		\w^t_{k,i+1} &= \prox_{\mu R}\{\z^t_{k,i+1}\}.
	}
	{\color{white}{hh}}\hspace{4mm} \textbf{End} \\
	{\color{white}{hh}}\hspace{4mm} set $\w^{t+1}_{k,0} = \w^{t}_{k,\widebar{N}}$ and $\bpsi^{t+1}_{k,0} = \bpsi^{t}_{k,\widebar{N}}$ \\
	\noindent \ \textbf{End}
	\\[-2mm] \HRule\vspace{-4mm}
\end{table}

\section{Simulation Results}\label{sec. simulation}

\subsection{Convergence performance of diffusion-AVRG}
\label{subsec-conv-diffusion-AVRG}
\begin{figure}
	\centering
	\includegraphics[width=2.3in]{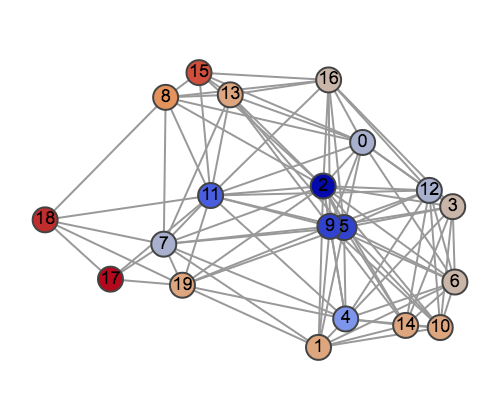}\vspace{-2.5mm}
	\caption{\small{A random connected network with $20$ nodes.}}
	\label{fig-network}
	\vspace{-3mm}
\end{figure}

In this subsection, we illustrate the convergence performance of diffusion-AVRG. We consider problem \eqref{prob-emp-dist} in which $J_k(w)$ takes the form of regularized logistic regression loss function:\vspace{-1mm}
\eq{\label{xcn23bh}
	\hspace{-2mm}J_k(w) \hspace{-0.7mm}\define\hspace{-0.7mm} \frac{1}{N_k}\hspace{-0.8mm}\sum_{n=1}^{N_k}\hspace{-0.8mm}\left(\frac{\rho}{2} \|w\|^2 \hspace{-0.7mm}+\hspace{-0.7mm} \ln\big(1\hspace{-0.7mm}+\hspace{-0.7mm}\exp(-\gamma_{k}(n) h_{k,n}\tran w)\big)\hspace{-0.8mm}\right) 
}
with $q_k = N_k/N$. The vector $h_{k,n}$ is the $n$-th feature vector kept by node $k$ and $\gamma_k(n) \in \{\pm1\}$ is the corresponding label. In all experiments, the factor $\rho$ is set to $1/N$, and the solution $w^\star$ to \eqref{prob-emp-dist} is computed by using the Scikit-Learn Package. All experiments are run over four datasets: Covtype.binary\footnote{\label{footnote.1}\small  \url{http://www.csie.ntu.edu.tw/~cjlin/libsvmtools/datasets/}}, RCV1.binary\footref{footnote.1},
MNIST\footnote{\small \url{http://yann.lecun.com/exdb/mnist/}}, and CIFAR-10\footnote{\small  \url{http://www.cs.toronto.edu/~kriz/cifar.html}}. The last two datasets have been transformed into binary classification problems by considering data with labels 2 and 4, i.e., digital two and four classes for MNIST, and cat and dog classes for CIFAR-10. {\color{black} In Covtype.binary we use $50,000$ samples as training data and each data has dimension $54$. In RCV1 we use $30,000$ samples as training data and each data has dimension $47,236$. In MNIST we use $10,000$ samples as training data and each data has dimension $784$. In CIFAR-10 we use $10,000$ samples as training data and each data has dimension $3072$.} All features have been preprocessed and normalized to the unit vector. We also generate a randomly connected network with $K=20$ nodes, which is shown in Fig. \ref{fig-network}. The associated doubly-stochastic combination matrix $A$ is generated by the Metropolis rule \cite{sayed2014adaptation}.\vspace{-0.5mm}

\begin{figure*}
	\centering
	\includegraphics[scale=0.42]{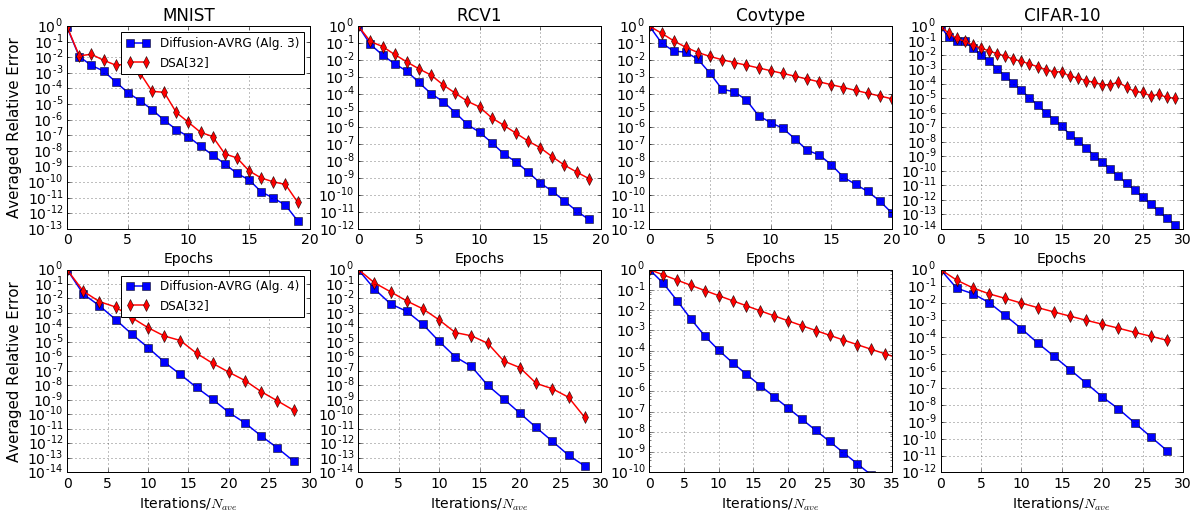}
	\vspace{-2mm}
	\caption{\small{Comparison between diffusion-AVRG and DSA over various datasets. Top: data are evenly distributed over the nodes; Bottom: data are unevenly distributed over the nodes. The average sample size is $N_{\rm ave}= \sum_{k=1}^{K}N_k / K$. The $y$-axis indicates the averaged relative square-error, i.e. $\frac{1}{K}\sum_{k=1}^{K}\bE\|\w_{k,0}^t - \w^\star\|^2/\|\w^\star\|^2$} \vspace{-4mm} }\label{fig-even-data}
\end{figure*}

\begin{figure}
	\centering
	\includegraphics[width=2.5in]{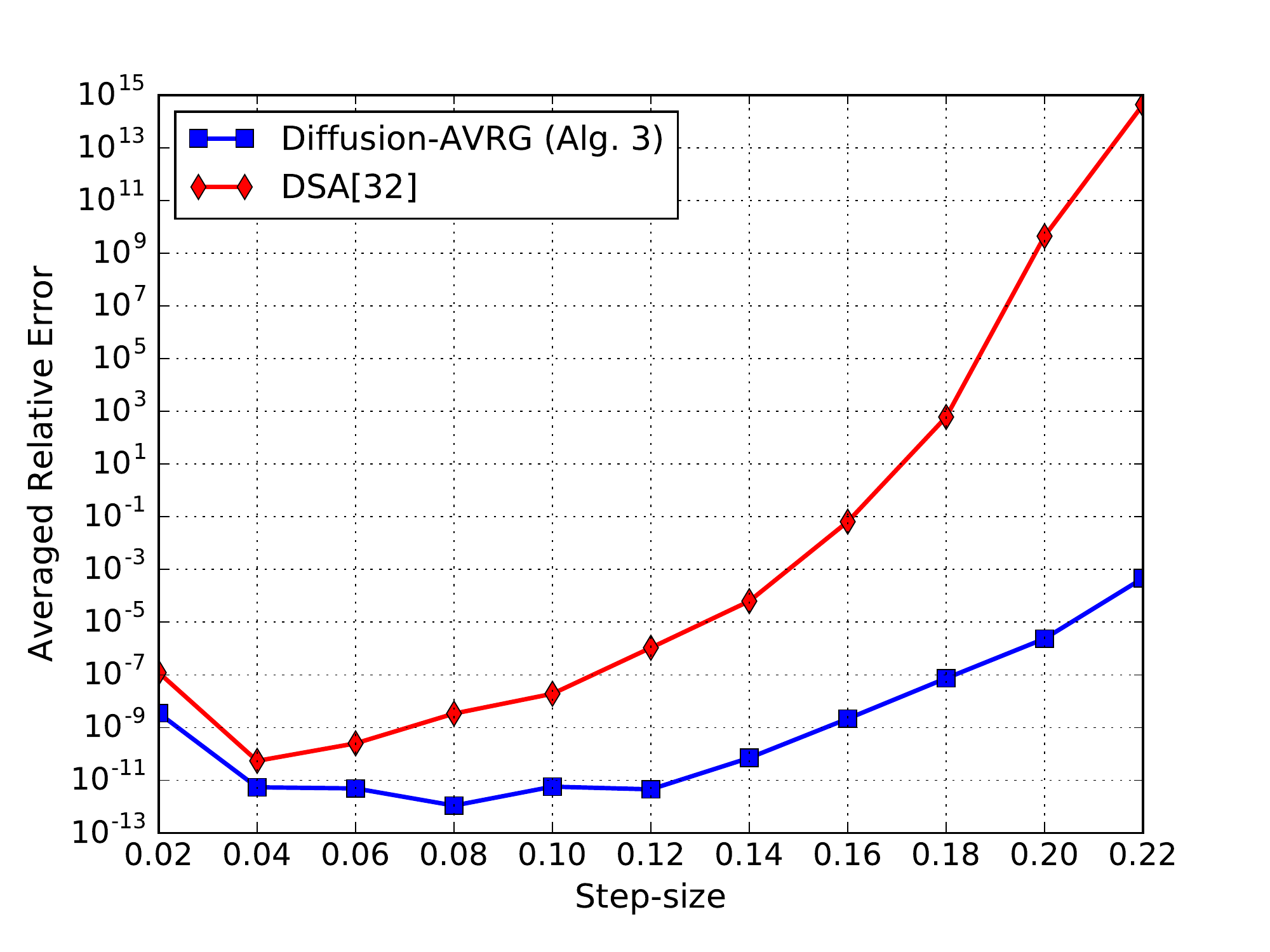}\vspace{-3mm}
	\caption{\small{{\color{black}Diffusion-AVRG is more stable than DSA. The $x$-axis indicates the step-size, and $y$-axis indicates the averaged relative square-error after $20$ epochs.}}}
	\label{fig-even-stability}
	\vspace{-4mm}
\end{figure}

\begin{figure}
	\centering
	\includegraphics[width=2.5in]{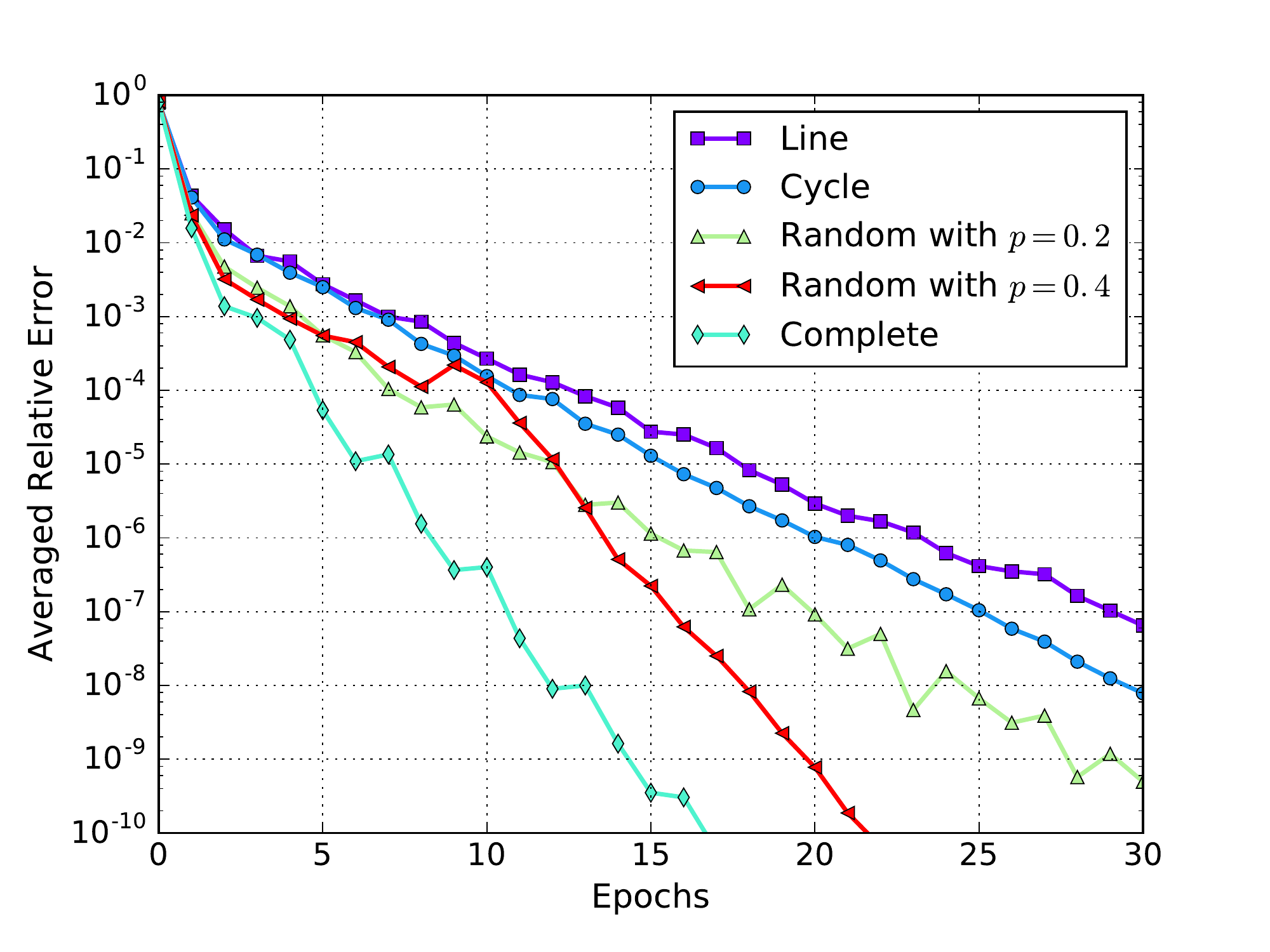}\vspace{-3mm}
	\caption{\small{{\color{black}The effects of topology over diffusion-AVRG.}}}
	\label{fig-topoloty}
	\vspace{-7mm}
\end{figure}

In our first experiment, we test the convergence performance of diffusion-AVRG (Algorithm 3) with even data distribution, i.e., $N_k = N/K$. We compare the proposed algorithm with DSA \cite{mokhtari2016dsa}, which  is based on SAGA\cite{defazio2014saga} and hence has significant memory requirement. In comparison, the proposed diffusion-AVRG algorithm does not need to store the gradient estimates and is quite memory-efficient. The experimental results are shown in the top 4 plots of Fig. \ref{fig-even-data}. To enable fair comparisons, we tune the step-size parameter of each algorithm for fastest convergence in each case. The plots are based on measuring the averaged relative square-error, $\frac{1}{K}\sum_{k=1}^{K}\bE\|\w_{k,0}^t - \w^\star\|^2/\|\w^\star\|^2$. It is observed that both algorithms converge linearly to $w^\star$, while diffusion-AVRG converges faster (especially on Covtype and CIFAR-10).

In our second experiment, data are randomly assigned to each node, and the sample sizes at the nodes may vary drastically. We now compare diffusion-AVRG (Algorithm 3) with DSA. Since there is no epoch for this scenario, we compare the algorithms with respect to the iterations count. In the result shown in bottom 4 plots of Fig. \ref{fig-even-data}, it is also observed that both algorithms converge linearly to $w^\star$, with diffusion-AVRG converging faster than DSA.\vspace{-0.5mm}

{\color{black}
	\subsection{Stability comparison with DSA}
	\label{subsec-stability}
	In this subsection, we compare the stability between DSA and diffusion-AVRG. For simplicity, this experiment is conducted in the context of solving a linear regression problem with synthetic data, and the dimension of the feature vector is set as $M = 10$. Each feature-label pair $(\h_n, \bgamma(n))$ is drawn from a Gaussian distribution $\cN(0, \Lambda)$, where $\Lambda$ is a positive diagonal matrix with the ratio of the largest diagonal value to the smallest diagonal value as $20$. We generate $N=20,000$ data points, which are evenly distributed over the $20$ nodes. The same topology shown in Fig.\ref{fig-network} is used in this experiment. We compare the convergence performance of diffusion-AVRG with DSA over a range of step-sizes from $0.02$ to $0.22$. The result is illustrated in Fig. \ref{fig-even-stability}. The $x$-axis indicates the step-size and $y$-axis indicates the averaged relative square-error. Each point in the curve indicates the convergence accuracy of that algorithm after $20$ epochs with the corresponding step-size. It is observed in Fig. \ref{fig-even-stability} that for all tested step-sizes, diffusion-AVRG is more accurate than DSA after running the same number of epochs. Also, it is observed that DSA starts diverging after step-size $\mu = 0.16$. In contrast, diffusion-AVRG remains convergent for all step-sizes within $[0.02, 0.22]$. This observation illustrates how diffusion-AVRG is endowed with a wider step-size range for stability than DSA. The  improved stability is inherited from the structure of the exact diffusion strategy \cite{yuan2017exact1,yuan2017exact2,sayed2014adaptive}. The improved stability range also helps explain why diffusion-AVRG is faster than DSA in Fig. \ref{fig-even-data}.}

\begin{figure}
	\centering
	\includegraphics[width=2.3in]{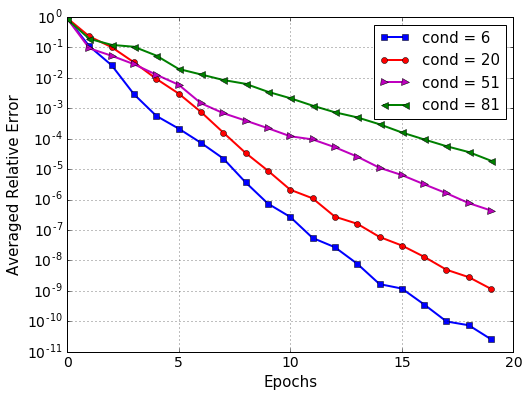}\vspace{-3mm}
	\caption{\small{{\color{black}The effects of condition number over diffusion-AVRG.}}}
	\label{fig-cond-num}
	\vspace{-3mm}
\end{figure}

{\color{black}
	\subsection{Parameters affecting convergence}
	In this subsection we test two parameters that effects the convergence of diffusion-AVRG: network topology and the condition number of the cost function. 
	In Theorem \ref{tho-balanced-a23}, it is observed that when the second largest eigenvalue, $\lambda$, of the combination matrix  is closer to $1$, or the condition number of the cost function $\delta/\nu$ is larger, the step-size should be smaller and hence the convergence rate slower. To illustrate such conclusion, we consider the same linear regression example as in Sec. \ref{subsec-stability}. In the first experiment, we evenly distribute $20,000$ data points over $50$ agents.  We test the convergence of diffusion-AVRG over 5 different topologies: a line graph, a cycle graph, a random graph with connection probability $p=0.2$, a random graph with connection probability $p=0.4$, and a complete graph. The combination matrix over the above graphs are generated according to the Metropolis-Hastings rule, and the value of $\lambda$ corresponding to the above 5 topologies are $0.9987$, $0.9927$, $0.9859$, $0.9381$ and $0$. The experimental result is shown in Fig. \ref{fig-topoloty}. Step-sizes for each topology are adjusted so that each curve reach its fastest convergence. It is observed that the more connected the network is, the faster diffusion-AVRG converges, which is consistent with Theorem \ref{tho-balanced-a23}. 
	
	In the second experiment, we adjust the covariance matrix of the feature vector $\h_n$ so that the condition number $\delta/\nu$ is different. Fig. \ref{fig-cond-num} depicts four convergence curves under different condition numbers. Step-sizes under each  condition number are optimized so that all curves reach their fastest convergence. It is observed that better condition numbers en-able faster convergence, which is consistent with Theorem \ref{tho-balanced-a23}.
}

\begin{figure}
	\centering
	\includegraphics[width=2.5in]{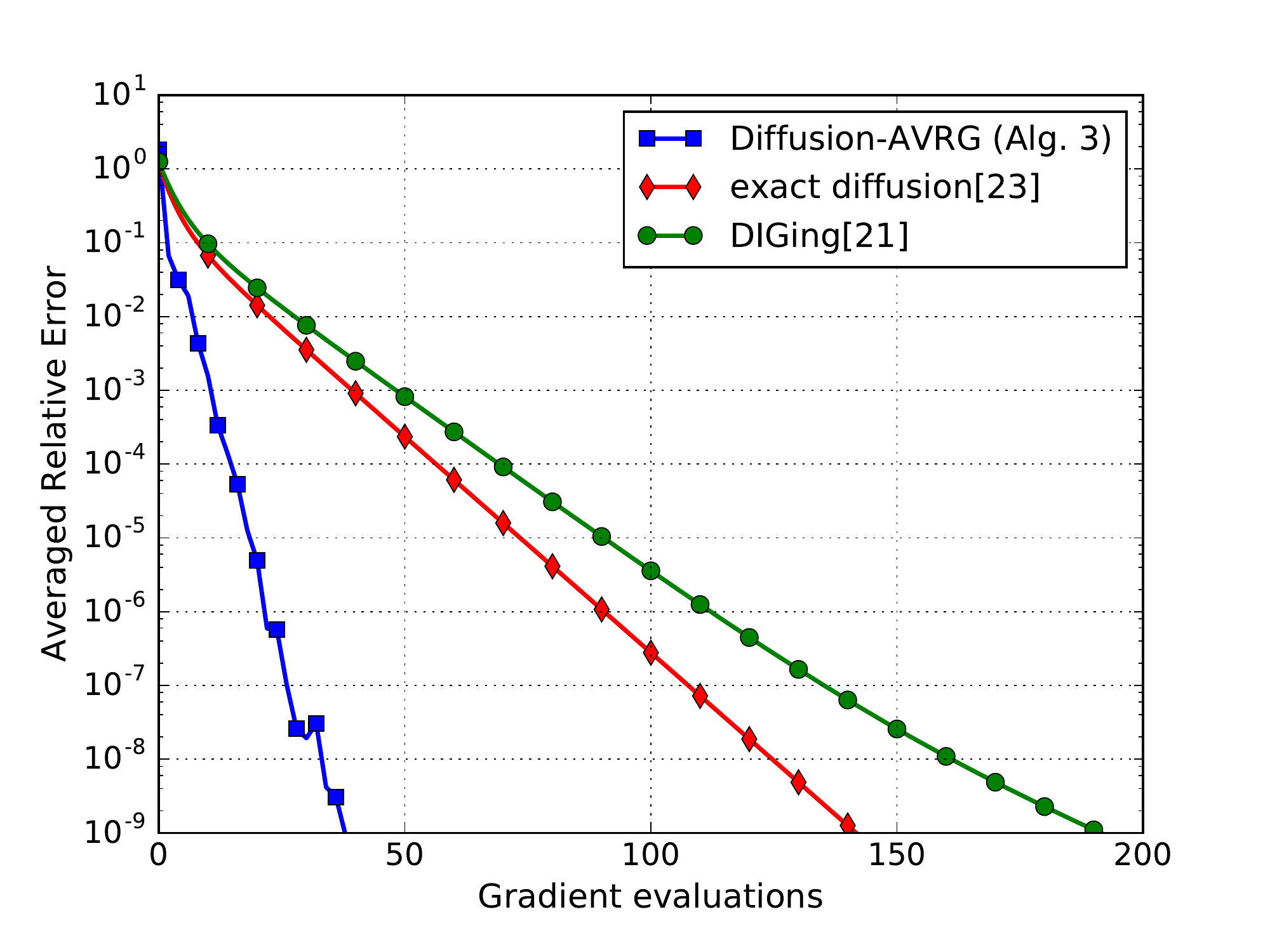}\vspace{-2.5mm}
	\caption{{\color{black}\small{The comparison of computational efficiency between diffusion-AVRG, exact diffusion and DIGing. The unit of $x$-axis is $\widebar{N}=1000$.}}}
	\label{compt_compare}
	\vspace{-5mm}
\end{figure}

{\color{black}
	\subsection{Computational efficiency of diffusion-AVRG}
	It is known that the single agent variance-reduced methods such as SVRG \cite{johnson2013accelerating} and SAGA \cite{defazio2014saga} can save computations compared to the full gradient descent. In this subsection we examine through numerical simulations whether diffusion-AVRG can save computations compared to the corresponding deterministic algorithms such as exact diffusion and DIGing. By ``saving computations'' we mean to reach a desirable convergence accuracy, diffusion-AVRG requires to calculate less gradients than exact diffusion and DIGing. Counting the number of gradient calculations during the convergence process is a common metric to evaluate computational efficiency --- see \cite{johnson2013accelerating, defazio2014saga, mokhtari2016dsa}. Note that diffusion-AVRG needs to calculate two gradients per iteration at agent $k$, and hence $2\widebar{N}$ gradients are required per epoch where $\widebar{N}$ is the size of the local dataset. In contrast, exact diffusion and DIGing will evaluate $\bar{N}$ gradients per iteration. 
	
	We consider the same experimental setting as in Sec. \ref{subsec-stability}. The performance of diffusion-AVRG, exact diffusion\cite{yuan2017exact1} and DIGing \cite{nedich2016achieving} are compared in Fig. \ref{compt_compare}. For each algorithm, we tune its step-size so that fastest convergence is reached. It is observed that to reach the relative accuracy $10^{-9}$, each agent in diffusion-AVRG requires to evaluate $40\widebar{N}$ gradients while exact diffusion and DIGing require $140\widebar{N}$ and $190\widebar{N}$, respectively. This experiment shows that exact-diffusion saves at least $70\%$ of gradient evaluations compared to exact diffusion and DIGing. The cost for such computational efficiency in diffusion-AVRG is more communication rounds. The computation and communication in diffusion-AVRG can be balanced by mini-batch technique as discussed in Sec. \ref{sec. minibatch}.
	
}

\begin{figure}
	\centering
	\includegraphics[width=2.8in]{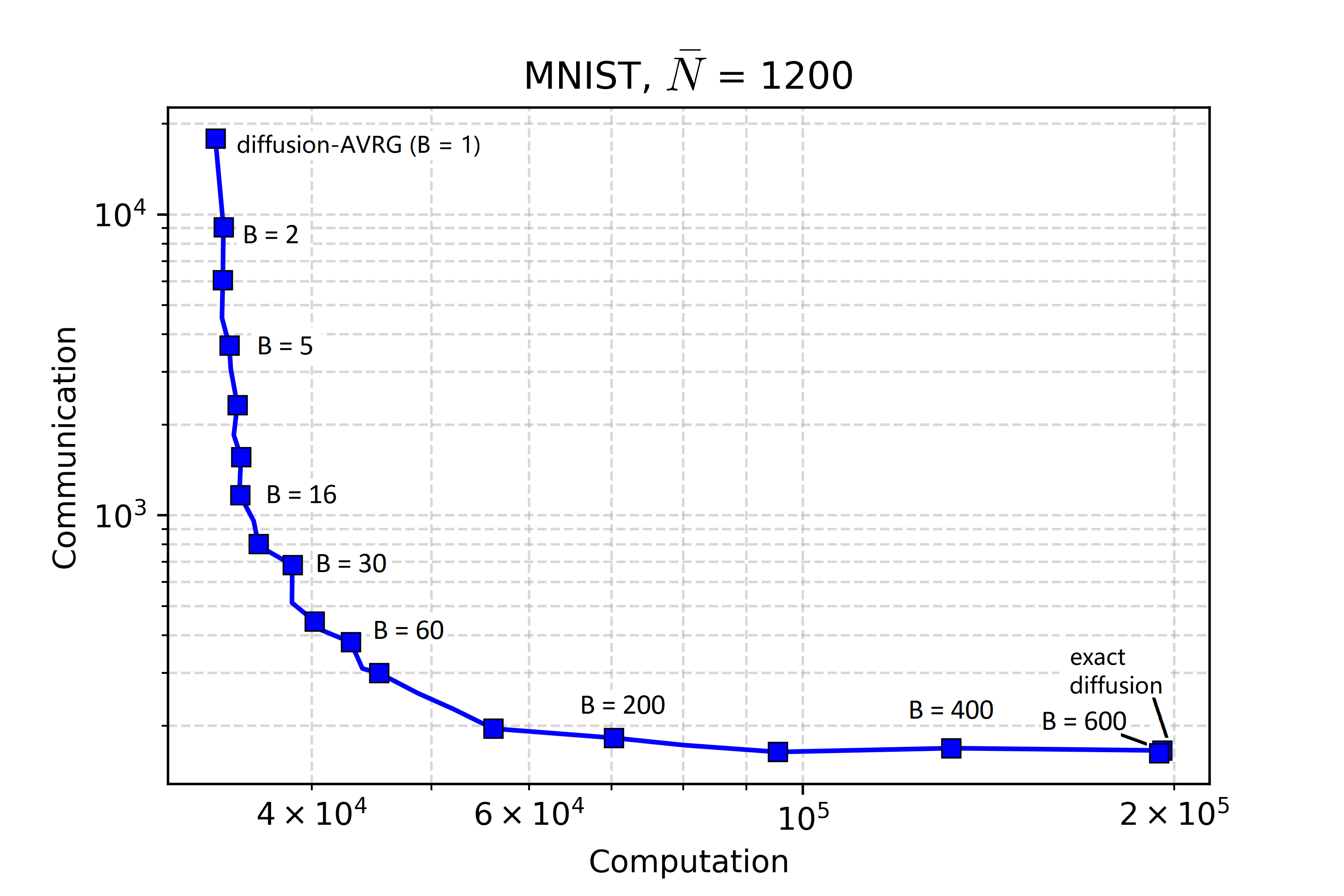}\vspace{-2.5mm}
	\caption{\small{Performance of diffusion-AVRG with different batch sizes on MNIST dataset. Each agent holds $\widebar{N}=1200$ data. In the $x$-axis, the computation is measured by counting the number of one-data gradients $\grad Q(w; x_{n})$ evaluated to reach accuracy $10^{-10}$. In the $y$-axis, the communication is measured by counting the number of communication rounds to reach accuracy $10^{-10}$.}}
	\label{mnist-batch}
	\vspace{-3mm}
\end{figure}

\begin{figure}
	\centering
	\includegraphics[width=2.8in]{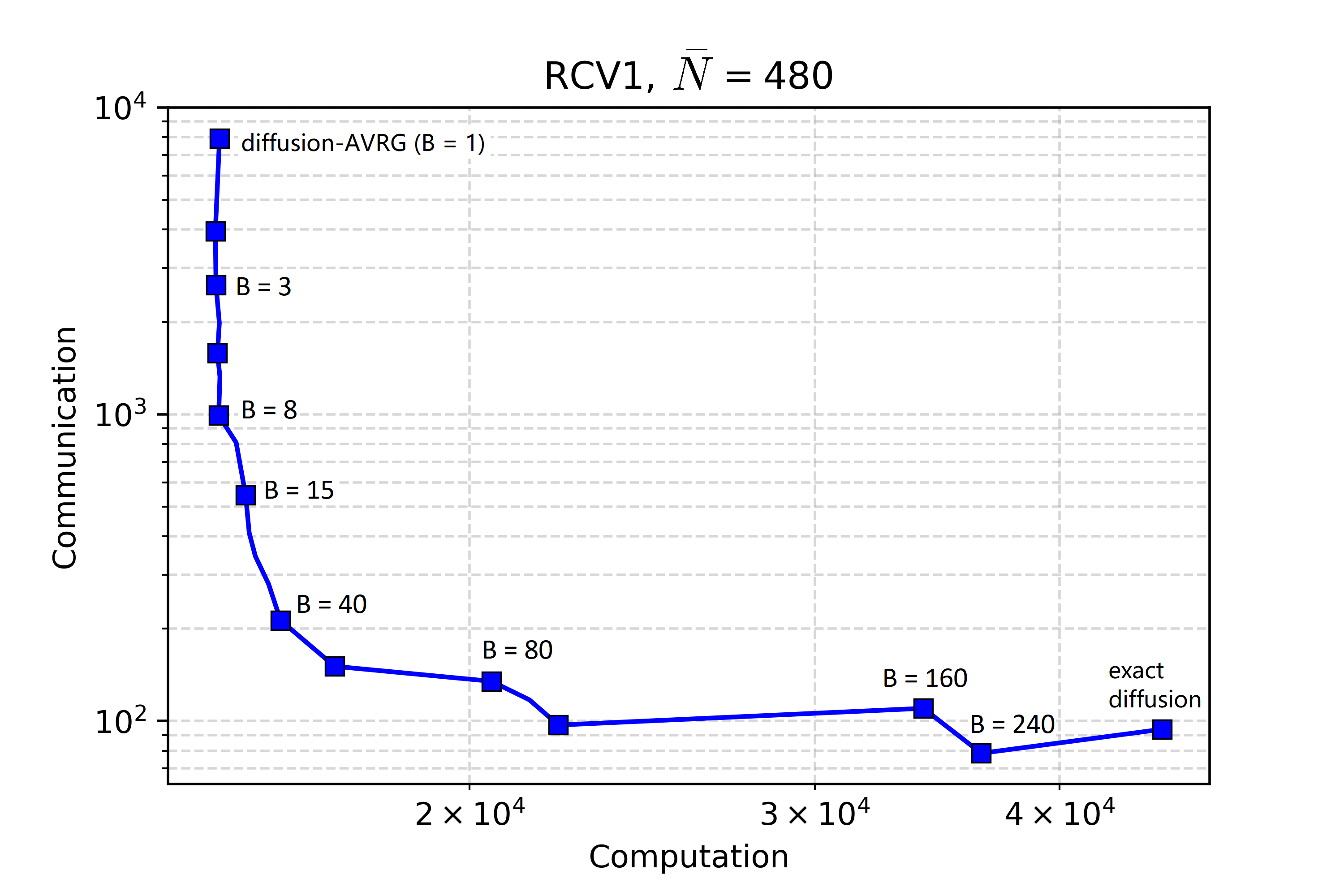}\vspace{-2.5mm}
	\caption{\small{Performance of diffusion-AVRG with different batch sizes on RCV1 dataset. Each agent holds $\widebar{N}=480$ data.}}
	\label{rcv-batch}
	\vspace{-5mm}
\end{figure}

{\color{black}
	\subsection{Balancing communication and computation}
}
\label{subsec-minibatch-experiments}
In this experiment, we test how the mini-batch size $B$ influences the computation and communication efficiency in diffusion-AVRG. The experiment is conducted on the MNIST and RCV1 datasets. For each batch size, we run the algorithm until the relative error reaches $10^{-10}$. The step-size for each batch size is adjusted to be optimal. The communication is examined by counting the number of message passing rounds, and the computation is examined by counting the number of $\grad Q(w; x_{n})$ evaluations. The exact diffusion is also tested for comparison. In Fig. \ref{mnist-batch}, we use ``AVRG" to indicate the standard diffusion-AVRG method. It is observed that standard diffusion-AVRG is more computation efficient than exact diffusion. To reach $10^{-10}$ relative error, exact diffusion needs around $2\times 10^5$ gradient evaluations while diffusion-AVRG just needs around $2 \times 10^4$ gradient evaluations. However, exact diffusion is much more communication efficient than diffusion-AVRG. To see that, exact diffusion requires around $200$ communication rounds to reach $10^{-10}$ error while diffusion-AVRG requires $2\times 10^4$ communication rounds. Similar observation also holds for RCV1 dataset, see Fig.\ref{rcv-batch}. 

It is also observed in Fig. \ref{mnist-batch} that mini-batch can balance the communication and computation for diffusion-AVRG. As batch size grows, the computation expense increases while the communication expense reduces. Diffusion-AVRG with appropriate batch-size is able to reach better performance than exact diffusion. For example, diffusion-AVRG with $B=200$ will save around $60\%$ computations while maintaining almost the same amount of communications. Similar observation also holds for RCV1 dataset, see Fig.\ref{rcv-batch}. 

\begin{figure*}
	\centering
	\includegraphics[width=7in]{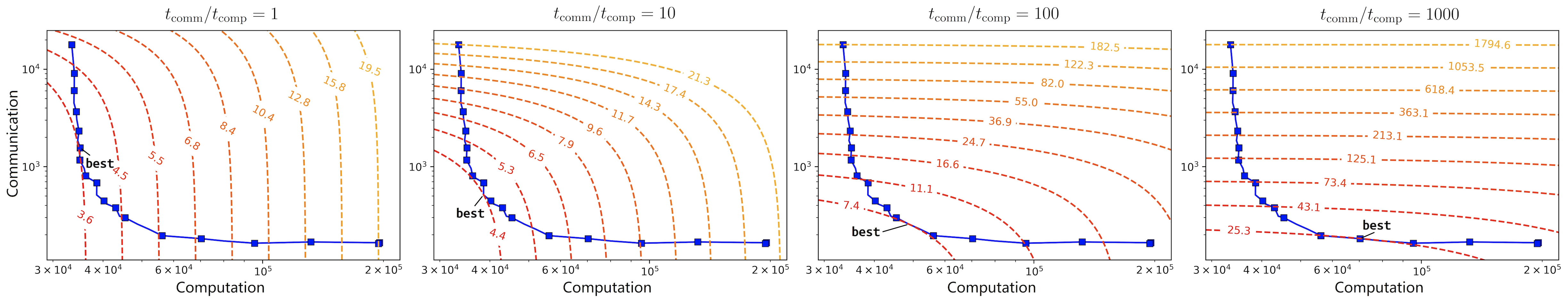}\vspace{-2.5mm}
	\caption{{\color{black}\small{Running time contour line for diffusion-AVRG with mini-batch. The $x$-axis and $y$-axis have the same meaning as in Fig. \ref{mnist-batch}. In all sub-figures, it is assumed that the calculation of one-data gradient takes one unit of time. For each sub-figure from left to right, one round of communication is assumed to take $1$, $10$, $100$ and $1000$ unit(s) of time. The unit for the value of each contour line is $10^4$.}}}
	\label{contour_line_4scenarios}
\end{figure*}

{\color{black}
	Based on the above experiment, we can further test the running time of diffusion-AVRG and compare it with exact diffusion. In this simulation, we assume the calculation of a one-data gradient $\grad Q(w;x_n)$ takes one unit of time, i.e. $t_{\rm comp}=1$.     We then consider four different scenarios in which one round of communication takes $1$, $10$, $100$ and $1000$ unit(s) of time,  respectively. For each scenario we depict the running time contour line. The running time contour line is calculated as follows. Suppose to reach the error $10^{-10}$, one algorithm needs to calculate $n_g$ gradients and communicate $n_c$ rounds, then the total running time is $t_{\rm comp} n_g + t_{\rm comm} n_c$ where $t_{\rm comp}=1$ and $t_{\rm comm}=1,10,100$ or $1000$ in different scenarios. All four scenarios are illustrated in Fig. \ref{contour_line_4scenarios}. The unit for the value of each contour line is $10^4$. In all scenarios, diffusion-AVRG with proper batch size is faster than exact diffusion in terms of running time. Let us take a closer look at the third sub-figure. It is observed that when the best batch size is employed in diffusion-AVRG, the total running time is $7.4 \times 10^4$. As a comparison, the total running time for exact diffusion is between $16.6 \times 10^4$ and $24.7 \times 10^4$. 
}

\begin{figure*}
	\centering
	\includegraphics[scale=0.23]{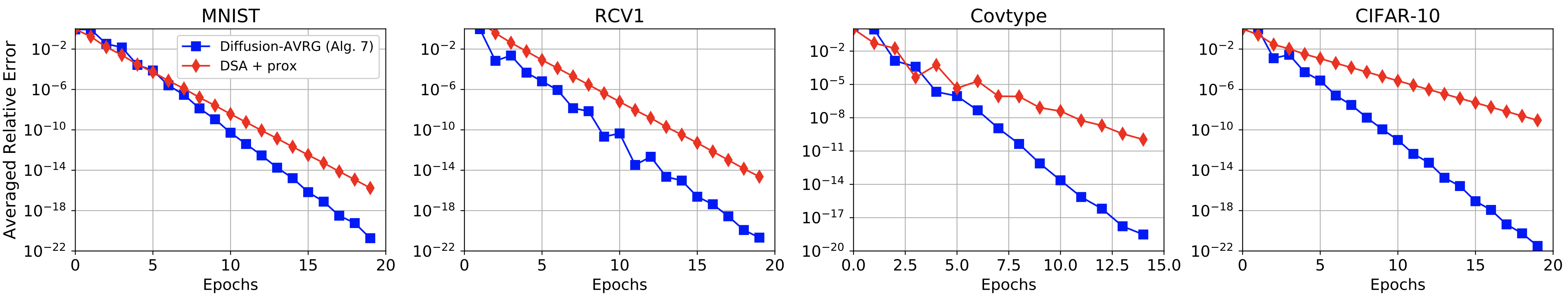}
	\vspace{-2mm}
	\caption{\small{{\color{black}Comparison between prox-diffusion-AVRG and prox-DSA over various datasets}.\vspace{-4mm} }}\label{fig-even-prox}
	\vspace{-3mm}
\end{figure*}

{\color{black}\vspace{2mm}
	\subsection{Prox-diffusion-AVRG}\vspace{-1mm}
	In this subsection we test the performance of prox-diffusion-AVRG listed in Algorithm 7. We consider problem \eqref{prob-prox} with $J_k(w)$ defined in \eqref{xcn23bh}, and $R(w) = \eta \|w\|_1$ where $\eta$ is the sparsity coefficient. For simplicity, we assume the sizes of local datasets are all equal. The experimental setting and datasets are the same as the first experiment in Sec.\ref{subsec-conv-diffusion-AVRG}. For MNIST, RCV1 and Covtype, we set $\eta = \rho = 0.005$. For CIFAR-10, we set $\eta = 0.0005$ and $\rho = 0.01$. We compare the performance of prox-diffusion-AVRG (Alg.7) and prox-DSA\footnote{Note that the original DSA algorithm in \cite{mokhtari2016dsa} cannot handle the composite optimization problem. We therefore combine SAGA and PG-EXTRA\cite{shi2015proximal} to reach prox-DSA that is able to handle non-smooth proximable regularizations.} over these datasets in Fig. \ref{fig-even-prox}. For each dataset, we tune the step-sizes so that both algorithms reach their fastest convergence. It is observed that for all datasets prox-diffusion-AVRG converges linearly, and it is faster than prox-DSA.
	
}

{\color{black}
	\section{Conclusion}
	This paper proposes diffusion-AVRG, which is a fully-distributed variance-reduced stochastic method. It saves computations compared to existing deterministic algorithms such as EXTRA\cite{shi2015extra}, exact diffusion\cite{yuan2017exact1} and DIGing\cite{nedich2016achieving}, and significantly reduces the memory requirement compared to DSA\cite{mokhtari2016dsa}. Moreover, diffusion-AVRG is more suitable for the practical scenarios in which data are distributed unevenly among networked agents. We also propose using mini-batch to balance computations and communications. Possible future work includes establishing convergence guarantees for prox-diffusion-AVRG and and extending diffusion-AVRG to nonconvex optimization and varying networks.}
	
\appendices
\small
\newcommand{\suo}{\hspace{-0.5mm}}
\newcommand{\ssuo}{\hspace{-1mm}}
\newcommand{\kuo}{\hspace{0.5mm}}

\section{Proof of Theorem \ref{tho-balanced-a23}}
\label{app-proof-main-theorem}
In this section we establish the linear convergence property of diffusion-AVRG (Algorithm 2). We start by transforming the exact diffusion recursions into an equivalent linear error dynamics driven by perturbations due to  gradient noise (see Lemma 2). By upper bounding the gradient noise (see Lemma 3), we derive a couple of useful inequalities for the size of the inner iterates (Lemma 4), epoch iterates (Lemma 5), and inner differences (Lemma 6). We finally introduce an energy function and show that it decays exponentially fast (Lemma 7). From this result we will conclude the convergence of $\bE\|\w_{k,0}^t - w^\star\|^2$ (as stated in \eqref{zljshdjdhdudsdsacsd-0} in Theorem 1).  Throughout this section we will consider the practical case where $\widebar{N} \ge 2$. When $\widebar{N} =1$, diffusion-AVRG reduces to the exact diffusion algorithm whose convergence is already established in \cite{yuan2017exact2}.

\subsection{Extended Network Recursion}
Recursions \eqref{ed-adapt-2}--\eqref{ed-comb-2} of Algorithm 2 only involve local variables $\w_{k,i}^t$, $\bphi_{k,i}^t$ and $\bpsi_{k,i}^t$. To analyze the convergence of all $\{\w_{k,i}^t\}_{k=1}^K$, we need to combine all  iterates from across the network into extended vectors. To do so, we introduce
\eq{
	\swb^t_i &= \col\{\w_{1,i}^t, \cdots, \w_{K,i}^t\} \\
	\bphi^t_i &= \col\{\bphi_{1,i}^t, \cdots, \bphi_{K,i}^t\} \\
	\bpsi^t_i &= \col\{\bpsi_{1,i}^t, \cdots, \bpsi_{K,i}^t\} \\
	\grad \cJ(\swb_i^t) &= \col\{\grad J_1(\w_{1,i}^t),\cdots, \grad J_K(\w_{K,i}^t)\}\\
	\widehat{\grad \cJ}(\swb_i^t) &= \col\{\widehat{\grad J}_1(\w_{1,i}^t),\cdots, \widehat{ \grad J}_K(\w_{K,i}^t)\} \\
	\tcA &= \overline{A} \otimes I_M
}
where $\otimes$ is the Kronecker product. With the above notation, for $ 0\le i \le \widebar{N}-1$ and $t\ge 0$,  recursions \eqref{ed-adapt-2}--\eqref{ed-comb-2} of Algorithm 2 can be rewritten as
\eq{\label{akjhb}
	\begin{cases}
		\bpsi_{i+1}^t &\hspace{-3mm}= \swb_i^t - \mu \widehat{\grad \cJ}(\swb_i^t), \\
		\bphi_{i+1}^t &\hspace{-3mm}= \bpsi_{i+1}^t + \swb_i^t - \bpsi_{i}^t, \\
		\swb_{i+1}^t &\hspace{-3mm}= \tcA\, \bphi_{i+1}^t,\\
	\end{cases}
}
{\color{black}and we let $\bpsi_{0}^{t+1} = \bpsi_{\widebar{N}}^t$ and $\swb_{0}^{t+1} = \swb_{\widebar{N}}^t$.} 
In particular, since $\bpsi_0^0$ is initialized to be equal to $\swb_0^0$, for $t=0$ and $i=0$, it holds that
\eq{\label{akjhb-0}
	\begin{cases}
		\bpsi_{1}^0 &\hspace{-3mm}= \swb_0^0 - \mu \widehat{\grad \cJ}(\swb_0^0), \\
		\bphi_{1}^0 &\hspace{-3mm}= \bpsi_{1}^0, \\
		\swb_{1}^0 &\hspace{-3mm}= \tcA\, \bphi_{1}^0,\\
	\end{cases}
}
Substituting the first and second equations of \eqref{akjhb} into the third one, we have that for $1\le i\le \widebar{N}$ and $t\ge 0$:
\eq{\label{zn2bsbd}
	\swb_{i+1}^t = \tcA\left( 2\swb_{i}^t \hspace{-0.8mm}-\hspace{-0.8mm} \swb_{i-1}^t  \hspace{-0.8mm}-\hspace{-0.8mm} \mu [\widehat{\grad \cJ}(\swb_{i}^t) \hspace{-0.8mm}-\hspace{-0.8mm} \widehat{\grad \cJ}(\swb_{i-1}^t)] \right),
}
and we let $\swb_{0}^{t+1} = \swb_{\widebar{N}}^t$ and $\swb_{1}^{t+1} = \swb_{\widebar{N}+1}^t$ for each epoch $t$. Moreover, we can also rewrite \eqref{akjhb-0} as
\eq{
	\swb_1^0 = \tcA \left( \swb_0^0 - \mu \widehat{\grad \cJ}(\swb_0^0) \right).
}
It is observed that recursion \eqref{zn2bsbd} involves two consecutive variables $\swb_i^t$ and $\swb_{i-1}^t$, which complicates the analysis. To deal with this issue, we introduce an auxiliary variable $\syb_i^t$ to make the structure in \eqref{zn2bsbd} more tractable. For that purpose, we first introduce the eigen-decomposition:
\eq{\label{23hsdbs8}
	\frac{1}{2K}(I_K - A) = U \Sigma U\tran,
}
where $\Sigma$ is a nonnegative diagonal matrix (note that $I_K \hspace{-0.8mm}-\hspace{-0.8mm} A$ is  positive semi-definite because $A$ is doubly stochastic), and $U$ is an orthonormal matrix. We also define
\eq{\label{V-definition}
	V \define U \Sigma^{1/2} U\tran, \quad\quad \cV \define V\otimes I_M.
}
Note that $V$ and $\cV$ are symmetric matrices. It can be verified (see Appendix \ref{app-pd-recursion}) that recursion \eqref{zn2bsbd} is equivalent to 
\begin{equation}
\left\{
\begin{aligned}
\swb^{t}_{i+1} =&\ \overline{\cA} \Big(\swb^{t}_{i} - \mu \widehat{\nabla \cJ} (\swb^{t}_{i})\Big) - K \cV \syb^t_{i} \label{s-ed-2-0}\\
\syb^t_{i+1} =&\ \syb^t_{i} +\cV\swb^{t}_{i+1}
\end{aligned}
\right.
\end{equation}
where $0 \le i \le \widebar{N}-1$ and $t\ge 0$, $\syb_0^0$ is initialized at $0$, and $\swb_{0}^{t+1} = \swb_{\widebar{N}}^t$, $\syb_{0}^{t+1} = \syb_{\widebar{N}}^t$ after epoch $t$. 
Note that recursion \eqref{s-ed-2-0} is very close to recursion for exact diffusion (see equation (93) in \cite{yuan2017exact1}), except that $\widehat{\grad \cJ}(\swb_i^t)$ is a stochastic gradient generated by AVRG. We denote the gradient noise by
\eq{
	\s(\swb_{i}^t) &= \widehat{\nabla \cJ} (\swb^{t}_{i}) - \nabla \cJ (\swb^{t}_{i}).\label{noise}
}
Substituting into \eqref{s-ed-2-0}, we get
\begin{equation}
\left\{
\begin{aligned}
\hspace{-1mm}\swb^{t}_{i+1} =&\ \overline{\cA} \Big(\swb^{t}_{i} \hspace{-0.8mm}-\hspace{-0.8mm} \mu {\nabla \cJ} (\swb^{t}_{i})\Big) - K\cV \syb^t_{i} - \mu \tcA\, \s(\swb_{i}^t) \label{s-ed-2-1}\\
\hspace{-1mm}\syb^t_{i+1} =&\ \syb^t_{i} +\cV\swb^{t}_{i+1}
\end{aligned}
\right.
\end{equation}
In summary, the exact diffusion recursions \eqref{ed-adapt-2}--\eqref{ed-comb-2} of Algorithm 2 are equivalent to form \eqref{s-ed-2-1}.


%
%
%
%
%
%

\subsection{Optimality Condition}\label{sec-opt-cond}
It is proved in Lemma 4 of \cite{yuan2017exact2} that there exists a {\em unique} pair of variables $(\sw^\star,\sy_o^\star)$, with $\sy_o^\star$ lying in the range space of $\cV$, such that 
\eq{\label{opt-cond}
	\mu \tcA \grad \cJ(\sw^\star) + K \cV \sy_o^\star = 0 \quad \mbox{and} \quad \cV \sw^\star = 0,
}
where we partition ${\sw}^{\star}$ into block entries of size $M\times 1$ each as follows:  $\sw^\star = \col\{w_1^\star,w_2^\star, \cdots, w_K^\star\} \in \RR^{KM}$. For such $(\sw^\star,\sy_o^\star)$, it further holds that the block entries of $\sw^\star$ are identical and coincide with the unique solution to problem (4), i.e.
\eq{
	w_1^\star = w_2^\star = \cdots = w_K^\star = w^\star.
}
In other words, equation \eqref{opt-cond} is the optimality condition characterizing the solution to problem \eqref{prob-emp-dist}.

\subsection{Error Dynamics}

Let $\tws_i^t = \sw^\star - \swb_i^t$ and $\tys_i^t = \sy^\star_o - \syb_i^t$ denote error vectors relative to the solution pair $(\sw^\star,\sy_o^\star)$. It is proved in Appendix \ref{app-recursion-w-y} that recursion \eqref{s-ed-2-1}, under Assumption \ref{ass}, can be transformed into the following recursion driven by a gradient noise term:
\eq{
	\ba{c}
	\tws_{i+1}^t\\
	\tys^t_{i+1}
	\ea = ( \cB  - \mu \cTb^t_{i})   \ba{c}
	\tws_{i}^t\\
	\tys^t_{i}
	\ea
	+ \mu \cB_l \s(\swb_{i}^t), \label{error-recursion-0}
}
where $0 \hspace{-0.3mm}\le\hspace{-0.3mm} i \hspace{-0.3mm}\le\hspace{-0.3mm} \widebar{N}-1$, $t \ge 0$, and $\tws_{0}^{t+1} = \tws_{\widebar{N}}^t$, $\tys_{0}^{t+1} = \tys_{\widebar{N}}^t$ after epoch $t$. Moreover, $\cB, \cB_l$ and $\cTb_i^t$ are defined as
\eq{
	\cB \hspace{-1mm}\define\hspace{-1mm} \ba{cc}
	\hspace{-2mm}\tcA \hspace{-2mm} &\hspace{-2mm}  - K\cV \hspace{-2mm} \ \\ 
	\hspace{-2mm}\cV\tcA \hspace{-2mm} &\hspace{-2mm} \tcA \hspace{-2mm} \
	\ea, \ 
	\cB_l \hspace{-1mm} \define \hspace{-1mm}
	\ba{c}
	\hspace{-2mm}\tcA \hspace{-1.5mm} \ \\
	\hspace{-2mm}\cV\tcA \hspace{-1.5mm} \
	\ea, \ 
	\cTb^t_{i} \hspace{-1mm} \define \hspace{-1mm}
	\ba{cc}
	\hspace{-2mm}\tcA \cHb^t_{i} &  0 \hspace{-1.5mm} \ \\
	\hspace{-2mm}\cV\tcA \cHb^t_{i} & 0 \hspace{-1.5mm} \
	\ea,
	\label{T-defi}
}
where
\eq{
	\cHb^t_i &= \diag\{ \H_{1,i}^t, \cdots, \H_{K,i}^t \}\in \RR^{KM\times KM},\\
	\H_{k,i}^t &\hspace{-1mm}\define \hspace{-1mm} \int_0^1 \grad^2 J_k\big(w^\star \hspace{-0.5mm}-\hspace{-0.5mm} r\widetilde{\w}_{k,i}^t\big)dr \in \RR^{M\times M}. \label{H_k_i-1}
}
To facilitate the convergence analysis of recursion \eqref{error-recursion-0}, we diagonalize $\cB$ and transform \eqref{error-recursion-0} into an equivalent error dynamics. 
From equations (64)--(67) in \cite{yuan2017exact2}, we know that $\cB$ admits an eigen-decomposition of the form
\eq{\label{cB-decompoision}
	\cB
	&\define \cX\cD \cX^{-1},
}
where $\cX, \cD$ and $\cX^{-1}$ are $KM$ by $KM$ matrices defined as
\eq{
	\cD &\define 
	\ba{ccc}
	I_{M} & 0 & 0\\
	0 & I_M & 0\\
	0 & 0 & \cD_1
	\ea \in \RR^{2KM\times 2KM}, \label{sush}\\
	\cX &\define 
	\ba{ccc}
	\hspace{-1mm}\cR_1 & \cR_2 & \cX_{R}\hspace{-1mm}
	\ea \in \RR^{2KM\times 2KM}, \label{whsnsm}\\
	\cX^{-1} &\define 
	\ba{c}
	\cL_1\tran\\
	\cL_2\tran\\
	\cX_L
	\ea \in \RR^{2KM\times 2KM}.  \label{X-inverse}
}
In \eqref{sush}, matrix $\cD_1 = D_1\otimes I_M$ and $D_1\in \RR^{2(K-1)\times 2(K-1)}$ is a diagonal matrix with $\|D_1\| = \lambda_2(A) \define \lambda <1$. In \eqref{whsnsm} and \eqref{X-inverse}, matrices $\cR_1$, $\cR_2$, $\cL_1$ and $\cL_2$ take the form
\eq{
	&\cR_1= 
	\ba{c}
	\mathds{1}_K\\
	0_K
	\ea \otimes I_M,
	\quad
	\hspace{2.2mm}\cR_2 = 
	\ba{c}
	0_K \\
	\mathds{1}_K
	\ea \otimes I_M \\
	&\cL_1= 
	\ba{c}
	\frac{1}{K}\mathds{1}_K\\
	0_K
	\ea \otimes I_M, \quad
	\cL_2=
	\ba{c}
	0_K\\
	\frac{1}{K}\mathds{1}_K
	\ea \otimes I_M \label{R and L-kron}
}
Moreover, $\cX_R \hspace{-0.5mm}\in\hspace{-0.5mm} \RR^{2KM \times 2(K-1)M}$ and $\cX_L \hspace{-0.5mm}\in\hspace{-0.5mm} \RR^{2(K-1)M \times 2KM}$ are some constant matrices. Since $\cB$ is independent of $\widebar{N}$, $\delta$ and $\nu$, all matrices appearing in \eqref{cB-decompoision}--\eqref{X-inverse} are independent of these variables as well. 
By multiplying $\cX^{-1}$ to both sides of recursion \eqref{error-recursion-0}, we have
\eq{\label{zxzan239}
	& \hspace{-1mm}\cX^{-1}
	\ba{c}
	\hspace{-1mm}\tws^t_{i+1}\hspace{-1mm}\\
	\hspace{-1mm}\tys^t_{i+1}\hspace{-1mm}
	\ea
	\hspace{-1mm}\nnb
	= & \hspace{-1mm}\;
	[\cX^{-1}(\cB  -  \mu\cTb_{i}^t) \cX] \cX^{-1}
	\ba{c}
	\hspace{-1mm}\tws_{i}^t\hspace{-1mm}\\
	\hspace{-1mm}\tys_{i}^t\hspace{-1mm}
	\ea
	+ \mu \cX^{-1} \cB_l
	\s(\swb_{i}^t) \nnb
	\hspace{-1mm}\overset{\eqref{cB-decompoision}}{=}\hspace{-1mm}&\ \Big( \cD \hspace{-0.5mm}-\hspace{-0.5mm} \mu\cX^{-1} \cTb_{i}^t \cX \Big) \left( \cX^{-1}
	\ba{c}
	\hspace{-1mm}\tws_{i}^t\hspace{-1mm}\\
	\hspace{-1mm}\tys_{i}^t\hspace{-1mm}
	\ea  \right) \hspace{-0.5mm}+\hspace{-0.5mm} \mu \cX^{-1}
	\cB_l \s(\swb_{i}^t).
}
Now we define
\eq{\label{vas}
	\ba{c}
	\hspace{-1mm}\bar{\sxb}^t_i\hspace{-1mm}\\
	\hspace{-1mm}\widehat{\sxb}^t_i\hspace{-1mm}\\
	\hspace{-1mm}\check{\sxb}^t_i \hspace{-1mm}
	\ea
	\define \cX^{-1}
	\ba{c}
	\tws_i^t \vspace{0.5mm}\\
	\tys_i^t
	\ea
	\overset{\eqref{X-inverse}}{=}
	\ba{c}
	\cL_1\tran\\
	\cL_2\tran\\
	\cX_L
	\ea
	\ba{c}
	\tws_i^t \vspace{0.5mm}\\
	\tys_i^t
	\ea,
}
as transformed errors. Moreover, we partition $\cX_R$ as 
\eq{
	\cX_R = 
	\ba{c}
	\cX_{R,u}\\
	\cX_{R,d}
	\ea, \quad \mbox{where} \quad \cX_{R,u} \in \RR^{KM\times 2(K-1)M}.
}
With the help of recursion \eqref{zxzan239}, we can establish the following lemma.

\begin{lemma} [\sc Useful Transformation] \label{lm-transform}
	When $\syb_0^0$ is initialized at $0$,  recursion \eqref{error-recursion-0} can be transformed into
	\eq{\label{23nsad-2}
		\ba{c}
		\hspace{-2mm}\bar{\sxb}^t_{i+1} \vspace{0.5mm} \hspace{-2mm} \ \\
		\hspace{-2mm}\check{\sxb}^t_{i+1} \hspace{-2mm} \ 
		\ea \hspace{-1.5mm}=\hspace{-1.5mm}
		\ba{cc}
		\hspace{-2mm}I_{M} \hspace{-1.2mm}-\hspace{-1.2mm} \frac{\mu}{K}\cI\tran \cHb_{i}^t\cI \hspace{-2mm} & \hspace{-2mm} -\frac{\mu}{K}\cI\tran\cHb_{i}^t\cX_{R,u} \vspace{0.5mm} \hspace{-1.5mm}\ \\
		\hspace{-2mm}- \mu\cX_L \cTb_{i}^t \cR_1 \hspace{-2mm}&\hspace{-2mm} \cD_1 \hspace{-1.2mm}-\hspace{-1.2mm} \mu \cX_L \cTb_{i}^t \cX_R \hspace{-1.5mm} \
		\ea
		\hspace{-1.5mm}
		\ba{c}
		\hspace{-2mm} \bar{\sxb}^t_{i}\vspace{0.5mm} \hspace{-1.5mm}\ \\
		\hspace{-2mm} \check{\sxb}^t_{i} \hspace{-1.5mm} \
		\ea
		\hspace{-1.5mm}+\hspace{-1.2mm} \mu \hspace{-1mm}
		\ba{c}
		\hspace{-2mm}\frac{1}{K}\cI\tran \vspace{0.5mm} \hspace{-1.7mm}\ \\
		\hspace{-2mm}\cX_L \cB_l \hspace{-1.7mm}\
		\ea \hspace{-1.5mm} \s(\hspace{-0.6mm}\swb_{i}^t\hspace{-0.4mm})
	}
	where $\cI = \mathds{1}_K \otimes I_M$. Moreover, the relation between $\tws_i^t, \tys_i^t$ and $\bar{\sxb}^t_{i}, \check{\sxb}^t_{i}$ in \eqref{zxzan239} reduces to
	\eq{\label{w-x-relation}
		\ba{c}
		\tws_i^t \vspace{0.5mm}\\
		\tys_i^t
		\ea
		= \cX
		\ba{c}
		\bar{\sxb}_i^t \vspace{0.5mm}\\
		0_M \vspace{0.5mm}\\
		\check{\sxb}_i^t
		\ea.
	}
	Notice that $\cX_L$, $\cX_R$, $\cX_{R,u}$ and $\cX$ are all constant matrices and independent of $\widebar{N}, \delta$ and $\nu$.
	%
\end{lemma}

\noindent \textbf{Proof}. See Appendix \ref{app-transform}. The proof is similar to the derivations in equations (68)--(82) from \cite{yuan2017exact2} except that we have an additional noise term in \eqref{error-recursion-0}. \qd

Starting from \eqref{23nsad-2}, we can derive the following recursions for the mean-square errors of the quantities $\bar{\sxb}_i^t$ and $\check{\sxb}_i^t$.
\begin{lemma}[\sc Mean-square-error Recursion] \label{lm-mse-recursion}
	Under Assum-ption \eqref{ass}, $\syb_0^0 = 0$ and for step-size  $\mu < 1/\delta$, it holds that
	\eq{\label{xngyi}
		\hspace{-3mm}
		\ba{c}
		\hspace{-2mm}\bE\|\bar{\sxb}^t_{i+1}\|^2\hspace{-2mm} \vspace{0.5mm}\\
		\hspace{-2mm}\ \ \bE\|\check{\sxb}^t_{i+1}\|^2 \hspace{-2mm}
		\ea
		& \preceq
		\ba{cc}
		\hspace{-2mm}1- a_1 \mu \nu & \frac{2a_2 \mu \delta^2}{\nu}\hspace{-2mm}\\
		\hspace{-2mm} {a_4 \mu^2 \delta^2} & \lambda + a_3 \mu^2 \delta^2 \hspace{-2mm}
		\ea
		\ba{c}
		\hspace{-2mm}\bE\|\bar{\sxb}^t_{i}\|^2\hspace{-2mm} \vspace{0.5mm}\\
		\hspace{-2mm} \ \ \bE\|\check{\sxb}^t_{i}\|^2 \hspace{-2mm}
		\ea \nnb
		& \quad \quad +
		\ba{c}
		\frac{2\mu}{\nu} \bE\|\s(\swb^t_{i})\|^2 \vspace{0.5mm}\\
		a_5\mu^2 \bE\|\s(\swb^t_{i})\|^2
		\ea,
	}
	where the scalars $a_l, 1\le l \le 5$ are defined in \eqref{a-notation}; they
	are positive constants that are independent of $\widebar{N}$, $\delta$ and $\nu$.
\end{lemma}
\textbf{Proof.} See Appendix \ref{app-lm-mse}. \qd

It is observed that recursion \eqref{xngyi} still mixes gradient noise $\bE\|\s(\swb_i^t)\|^2$ (which is correlated with $\swb_i^t$) with iterates $\bar{\sxb}_i^t$ and $\check{\sxb}_i^t$. To establish the convergence of $\bE\|\bar{\sxb}^t_{i}\|^2$ and $\bE\|\check{\sxb}^t_{i}\|^2$, we need to upper bound  $\bE\|\s(\swb_i^t)\|^2$ with terms related to $\bar{\sxb}_i^t$ and $\check{\sxb}_i^t$. In the following lemma we provide such an upper bound.
\begin{lemma}[\sc Gradient Noise]\label{lm-gradient-noise}
	Under Assumption \ref{ass}, the second moment of the gradient noise term satisfies:
	\eq{\label{zx23b9}
		& \bE\|\s(\swb_i^t)\|^2\nnb
		&\le 6 b \delta^2 \bE\|\bar{\sxb}^t_{i} - \bar{\sxb}^t_{0}\|^2 + 12 b \delta^2 \bE\|\check{\sxb}^t_{i}\|^2 + 18 b \delta^2 \bE\|\check{\sxb}^t_{0}\|^2 \nnb
		&\quad \ + \frac{3b\delta^2}{\widebar{N}}\sum_{j=0}^{\widebar{N}-1}\bE\|\bar{\sxb}^{t-1}_{j} \hspace{-0.8mm}-\hspace{-0.8mm} \bar{\sxb}^{t-1}_{\widebar{N}}\|^2 \hspace{-1mm}+\hspace{-1mm} \frac{6b\delta^2}{\widebar{N}}\sum_{j=0}^{\widebar{N}-1}\bE\|\check{\sxb}^{t-1}_{j}\|^2,
	}
	where $b=\|\cX\|^2$ is a positive constant that is independent of $\widebar{N}$, $\nu$ and $\delta$.
\end{lemma}
\textbf{Proof}. See Appendix \ref{app-gradient-noise}. \qd

In the following subsections, we will exploit the error dynamic \eqref{xngyi} and the upper bound \eqref{zx23b9} to establish the convergence of $\bE\|\bar{\sxb}^t_{i}\|^2$ and $\bE\|\check{\sxb}^t_{i}\|^2$, from which we will conclude later the convergence of $\bE\|\tws_i^t\|^2$. 
%
%

\subsection{Useful Inequalities}
To simplify the notation, we define
\eq{
	\vA^t &\define \frac{1}{\widebar{N}}\sum_{j=0}^{\widebar{N}-1}\bE\|\bar{\sxb}_{j}^t - \bar{\sxb}_{0}^t\|^2, \\
	\vB^t &\define \frac{1}{\widebar{N}}\sum_{j=0}^{\widebar{N}-1}\bE\|\bar{\sxb}_{j}^t - \bar{\sxb}_{\widebar{N}}^t\|^2, \\
	\vC^t &\define \frac{1}{\widebar{N}}\sum_{j=0}^{\widebar{N}-1}\bE\|\check{\sxb}_{j}^t\|^2.
}
All these quantities appear in the upper bound on gradient noise in \eqref{zx23b9}, and their recursions will be required to establish the final convergence theorem.

\begin{lemma}[\sc $\ \bE\|\check{\sxb}_i^t\|^2$ Recursion] \label{lm-check-x}
	Suppose Assumption 1 holds. If the step-size $\mu$ satisfies
	\eq{\label{mu-C1}
		\mu \le C_1\sqrt{\frac{1-\lambda}{\delta^2\widebar{N}}},
	}
	where $C_1 > 0$, which is defined in \eqref{C1}, is a constant independent of $\widebar{N}$, $\nu$ and $\delta$, it then holds that
	\eq{\label{sum-check-x}
		\hspace{-5mm}\vC^t &\le c_1 \mu^2 \delta^2 \widebar{N} \bE\|\bar{\sxb}^t_{0}\|^2 \hspace{-0.8mm}+\hspace{-0.8mm} \lambda_3 \bE\|\check{\sxb}_0^t\|^2   \hspace{-0.8mm}+\hspace{-0.8mm} c_2 \mu^2 \delta^2\widebar{N} \vA^t \nnb 
		&\hspace{1cm} + c_3\mu^2\delta^2\widebar{N}\vB^{t-1} \hspace{-0.8mm}+\hspace{-0.8mm} c_4\mu^2\delta^2\widebar{N}\vC^{t-1}, \\		
		\hspace{-5mm}\bE\|\check{\sxb}_{0}^{t+1}\|^2 &\le c_1 \mu^2 \delta^2 \widebar{N} \bE\|\bar{\sxb}^t_{0}\|^2 \hspace{-0.8mm}+\hspace{-0.8mm} \lambda_2 \bE\|\check{\sxb}_0^t\|^2    \hspace{-0.8mm}+\hspace{-0.8mm} c_2 \mu^2 \delta^2\widebar{N} \vA^t  \nnb
		& \hspace{1cm}\hspace{-0.8mm}+\hspace{-0.8mm} c_3\mu^2\delta^2\widebar{N}\vB^{t-1}  \hspace{-0.8mm}+\hspace{-0.8mm} c_4\mu^2\delta^2\widebar{N}\vC^{t-1},\label{check-x}
	}
	where the constants $\lambda_2<1,\ \lambda_3<1$, and $\{c_l\}_{l=1}^4$, which are defined in Appendix \ref{app-check-x}, are all positive scalars that are independent of $\widebar{N}$, $\nu$ and $\delta$.
\end{lemma}
\textbf{Proof}. See Appendix \ref{app-check-x}. \qd

\begin{lemma}[\sc $\ \bE\|\bar{\sxb}_0^t\|^2$ Recursion] \label{lm-bar-x} 	Suppose Assumption 1 holds. If the step-size $\mu$ satisfies
	\eq{
		\mu \le C_2\left(\frac{\nu\sqrt{1-\lambda}}{\delta^2 \widebar{N}}\right),
	}
	where $C_2 > 0$, which is defined in \eqref{C2}, is a constant independent of $\widebar{N}$, $\nu$ and $\delta$, it then holds that
	\eq{\label{l2nsdsdf}
		& \bE\|\bar{\sxb}^{t+1}_0\|^2 \nnb
		&\le \left(1- \frac{\widebar{N}}{3}a_1 \mu \nu\right)\bE\|\bar{\sxb}^t_{0}\|^2  +  \frac{d_1 \mu \delta^2 \widebar{N} }{ \nu}  \bE\|\check{\sxb}^t_{0}\|^2  \nnb
		&\quad + \frac{d_2\delta^2\mu \widebar{N}}{\nu}\vA^t + \frac{d_3\delta^2\mu \widebar{N}}{\nu}\vB^{t-1}  + \frac{d_4\delta^2\mu \widebar{N}}{\nu} \vC^{t-1}
	}
	where $\{d_l\}_{l=1}^4$, which are defined in \eqref{d-notation}, are positive constants that are independent of $\widebar{N}$, $\nu$ and $\delta$.
\end{lemma}
\textbf{Proof}. See Appendix \ref{app-bar-x}.\qd

\begin{lemma}[\sc Inner Difference Recursion] \label{lm-inner-recursion}Suppose Assumption 1 holds. If the step-size $\mu$ satisfies
	\eq{
		\mu \le C_3\sqrt{\frac{1-\lambda}{\delta^2 \widebar{N}}},
	}
	where $C_3 > 0$, which is defined in \eqref{mu-3}, is a constant independent of $\widebar{N}$, $\nu$ and $\delta$, it then holds that
	\eq{\label{zxn238sd}
		&\vA^t \hspace{-1mm}\le\hspace{-1mm} 12\mu^2\delta^2 \widebar{N}^2 \bE\|\bar{\sxb}_0^t\|^2 \hspace{-1mm}+\hspace{-1mm} e_6 \mu^2 \delta^2 \widebar{N}^2 \bE\|\check{\sxb}_0^t\|^2 \hspace{-1mm}+\hspace{-1mm} 2e_3\mu^2\delta^2\widebar{N}^2 \vA^t \nnb
		&\quad\quad\quad + 2e_4\mu^2\delta^2 \widebar{N}^2 \vB^{t-1} + 2e_5\mu^2\delta^2 \widebar{N}^2 \vC^{t-1},\\
		&\vB^{t} \hspace{-1mm}\le\hspace{-1mm} 12\mu^2\delta^2 \widebar{N}^2 \bE\|\bar{\sxb}_0^t\|^2 \hspace{-1mm}+\hspace{-1mm} e_6 \mu^2 \delta^2 \widebar{N}^2 \bE\|\check{\sxb}_0^t\|^2 \hspace{-1mm}+\hspace{-1mm} 2e_3\mu^2\delta^2\widebar{N}^2  \vA^t \nnb
		&\quad\quad\quad + 2e_4\mu^2\delta^2 \widebar{N}^2\vB^{t-1} + 2e_5\mu^2\delta^2 \widebar{N}^2 \vC^{t-1} \label{zxlwseknsd}
	}
	where $\{e_i\}_{i=3}^6$, which are defined in \eqref{e-notation}, are positive constants that are independent of $\widebar{N}$, $\nu$ and $\delta$.
\end{lemma}
\textbf{Proof}. See Appendix \ref{app-inner-recursion}. \qd

\subsection{Linear Convergence}
With the above inequalities, we are ready to establish the linear convergence of the transformed diffusion-AVRG recursion \eqref{23nsad-2}.
\begin{lemma}[\sc Linear Convergence]\label{tho-balanced} Under Assumption \ref{ass}, if the step-size $\mu$ satisfies
	\eq{\label{2sdbans8}
		\mu \le C\left(\frac{\nu (1-\lambda)}{\delta^2 \widebar{N}}\right),
	}
	where $C>0$, which is defined in \eqref{C}, is a constant independent of $\widebar{N}$, $\nu$ and $\delta$, and $\lambda = \lambda_2(A)$ is second largest eigenvalue of the combination matrix $A$, it then holds that
	\eq{\label{zljshdjdhdudsdsacsd}
		& \left(\bE\|\bar{\sxb}^{t+1}_0\|^2 + \bE\|\check{\sxb}_{0}^{t+1}\|^2\right) + \frac{\gamma}{2} \left( \vA^{t+1} + \vB^t + \vC^t \right) \nnb
		\le& \rho \left\{ \left( \bE\|\bar{\sxb}^t_{0}\|^2 + \bE\|\check{\sxb}^t_{0}\|^2 \right)  + \frac{\gamma}{2} (\vA^t + \vB^{t-1} + \vC^{t-1})\right\}
	}
	where $\gamma = 8f_5\delta^2 \mu \widebar{N}/\nu > 0$ is a constant, and
	\eq{
		\rho = \frac{1- \frac{\widebar{N}}{8}a_1 \mu \nu}{1- 8 f_1 f_5 \mu^3\delta^4 \widebar{N}^3 / \nu} < 1.
	}
	The positive constants $a_1$, $f_1$ and $f_5$ are independent of $\widebar{N}$, $\nu$ and $\delta$. Their definitions are in \eqref{a-notation} and \eqref{f-notation}.
\end{lemma}
\textbf{Proof}. See Appendix \ref{app-tho-balanced}. \qd

Using Lemma \ref{tho-balanced}, we can now establish the earlier Theorem \ref{tho-balanced-a23}.

\noindent \textbf{Proof of Theorem \ref{tho-balanced-a23}.} From recursion \eqref{zljshdjdhdudsdsacsd}, we conclude that
\eq{
	&\left(\bE\|\bar{\sxb}^{t+1}_0\|^2 + \bE\|\check{\sxb}_{0}^{t+1}\|^2\right) + \frac{\gamma}{2} \left( \vA^{t+1} + \vB^t + \vC^t \right)  \nnb
	&\le \rho^{t} \left\{ \left( \bE\|\bar{\sxb}^1_{0}\|^2 + \bE\|\check{\sxb}^1_{0}\|^2 \right)  + \frac{\gamma}{2} (\vA^1 + \vB^{0} + \vC^{0})\right\}.
}
Since $\gamma > 0$, it also holds that
\eq{\label{zxbswdekhjwekjh}
	&\bE\|\bar{\sxb}^{t+1}_0\|^2 + \bE\|\check{\sxb}_{0}^{t+1}\|^2 \nnb
	\le &\ \rho^{t} \left\{ \left( \bE\|\bar{\sxb}^1_{0}\|^2 + \bE\|\check{\sxb}^1_{0}\|^2 \right)  + \frac{\gamma}{2} (\vA^1 + \vB^{0} + \vC^{0})\right\}.
}
On the other hand, from \eqref{w-x-relation} we have
\eq{
	\|\tws_0^{t+1}\|^2 + \|\tys_0^{t+1}\|^2 \le \|\cX\|^2 \left(\| \bar{\sxb}_0^{t+1}\|^2 + \|\check{\sxb}_0^{t+1} \|^2 \right).
}
By taking expectation of both sides, we have
\eq{\label{23sdnsd8}
	\bE\|\tws_0^{t+1}\|^2 + \bE\|\tys_0^{t+1}\|^2 \le \|\cX\|^2 \left( \bE\| \bar{\sxb}_0^{t+1}\|^2 + \bE\|\check{\sxb}_0^{t+1} \|^2 \right).
}
Combining \eqref{zxbswdekhjwekjh} and \eqref{23sdnsd8}, we have
\eq{
	&\ \bE\|\tws_0^{t+1}\|^2 + \bE\|\tys_0^{t+1}\|^2 \nnb
	\le &\ \rho^t \underbrace{ \left(\hspace{-0.8mm}\|\cX\|^2 \hspace{-1mm} \left\{\hspace{-1mm} \left( \bE\|\bar{\sxb}^1_{0}\|^2 \hspace{-1mm}+\hspace{-1mm} \bE\|\check{\sxb}^1_{0}\|^2 \right) \hspace{-1mm}+\hspace{-1mm} \frac{\gamma}{2} (\vA^1 \hspace{-1mm}+\hspace{-1mm} \vB^{0} \hspace{-1mm}+\hspace{-1mm} \vC^{0})\right\}\right)}_{\define D}.
}
Since $\bE\|\tws_0^{t+1}\|^2 = \sum_{k=1}^{K}\bE\|w^\star - \w_{k,0}^{t+1}\|^2 \le  \bE\|\tws_0^{t+1}\|^2 + \bE\|\tys_0^{t+1}\|^2$, we conclude \eqref{zljshdjdhdudsdsacsd-0}.
\qd

\section{Proof of recursion \eqref{s-ed-2-0}}\label{app-pd-recursion}
\noindent Since $V=U\Sigma^{1/2}U\tran$, it holds that
\eq{
	V^2\suo  = \suo U\Sigma U\tran\suo \overset{\eqref{23hsdbs8}}{=}\suo (I_K-A)/2K,
}
which implies that
\eq{\label{23hsnnnn}
	\cV^2\suo = \suo  V^2 \suo \otimes \suo I_M \suo = \suo (I_{KM}\suo -\suo \cA)/2K.
}
Moreover, since $A\mathds{1}_K = \mathds{1}_K$ we get
\eq{
	 V^2 \mathds{1}_K = (I_{KM}- A)\mathds{1}_K/2K = 0.
}
By noting that $\|V \mathds{1}_K\|^2 = \mathds{1}_K\tran V^2 \mathds{1}_K = 0$, we conclude that
\eq{\label{23bsnd8}
	V \mathds{1}_K = 0, \quad \mbox{and} \quad \cV \cI = 0,
}
where $\cI \define \mathds{1}_K \otimes I_M$. {\color{black}Result \eqref{23bsnd8} will be used in Appendix \ref{app-transform}.}

Now, for $t=0$ and $i=0$, substituting $\syb_0^0=0$ into \eqref{s-ed-2-0} we have
\begin{equation}
\left\{
\begin{aligned}
\swb^{0}_{1} =&\ \overline{\cA} \Big(\swb^{0}_{0} - \mu \widehat{\nabla \cJ} (\swb^{0}_{0})\Big) \label{s-ed-2-1-2}\\
\syb^0_{1} =&\ \cV\swb^{0}_{1}
\end{aligned}
\right.
\end{equation}
The first expression in \eqref{s-ed-2-1-2} is exactly the first expression in \eqref{zn2bsbd}. For $t\ge 0$ and $1\le i \le \widebar{N}$, from the first recursion in \eqref{s-ed-2-0} we have
\eq{\label{23l8sdb}
	\swb_{i+1}^t \suo\suo - \suo\suo \swb_{i}^t = &\tcA\left( \swb_i^t \suo - \suo \swb_{i-1}^t \suo - \suo \mu \big(\widehat{\grad \cJ}(\swb_i^t)\suo -\suo \widehat{\grad \cJ}(\swb_{i-1}^t) \big) \right)\suo 
	\nnb
	&-\suo K\cV(\syb_i^t\suo -\suo \syb_{i-1}^t),
}
We let $\swb_{1}^{t+1} = \swb_{\widebar{N}+1}^t$ and $\swb_{0}^{t+1} = \swb_{\widebar{N}}^t$ after epoch $t$. Recalling from the second recursion in \eqref{s-ed-2-0} that $\syb_i^t - \syb_{i-1}^t = \cV \swb_{i}^t$, and substituting into \eqref{23l8sdb} we get
\eq{\label{23l8sdb-2}
	&\hspace{-5mm} \swb_{i+1}^t - \swb_{i}^t \nnb
	=&\ \tcA\left( \swb_i^t\suo -\suo \swb_{i-1}^t\suo-\suo\mu \big(\widehat{\grad \cJ}(\swb_i^t)\suo-\suo\widehat{\grad \cJ}(\swb_{i-1}^t) \big) \right)\suo-\suo K \cV^2 \swb_i^t \nnb
	\overset{\eqref{23hsnnnn}}{=}&\ \tcA\left( \swb_i^t  \suo - \suo  \swb_{i-1}^t  \suo - \suo  \mu \big(\widehat{\grad \cJ}(\swb_i^t)  \suo - \suo  \widehat{\grad \cJ}(\swb_{i-1}^t) \big) \right) \nnb
	&\ - \frac{1}{2}\suo  \left({I_{KM}  \suo - \suo  \cA}\right) \swb_i^t.
}
Using $\tcA = (I_{KM} + \cA)/2$, the above recursion can be rewritten as 
\eq{
	\swb_{i+1}^t = \suo  \tcA\left( 2\swb_i^t \suo - \suo \swb_{i-1}^t \suo - \suo \mu \big(\widehat{\grad \cJ}(\swb_i^t) \suo - \suo \widehat{\grad \cJ}(\swb_{i-1}^t) \big) \right)
}
which is the second recursion in \eqref{zn2bsbd}.

\section{Proof of recursion \eqref{error-recursion-0}}\label{app-recursion-w-y}
The proof of \eqref{error-recursion-0} is similar to (36)--(50) in \cite{yuan2017exact2} except that we have an additional gradient noise term $\s(\swb_i^t)$. We subtract $\sw^\star$ and $\sy_o^\star$ from both sides of \eqref{s-ed-2-1} respectively and use the fact that $\tcA\sw^\star = \frac{1}{2}(I_{MK} + \cA)\sw^\star = \sw^\star$ to get
\begin{equation}
\left\{
\begin{aligned}
\tws^{t}_{i+1} \suo = \suo &\ \overline{\cA} \Big(\tws^{t}_{i} \suo + \suo \mu {\nabla \cJ} (\swb^{t}_{i})\Big) \suo + \suo K\cV \syb^t_{i} \suo + \suo \mu \tcA\, \s(\swb_{i}^t) \label{s-ed-2-2}\\
\tys^t_{i+1} \suo = \suo &\ \tys^t_{i} \suo \suo - \suo \suo \cV\swb^{t}_{i+1}
\end{aligned}
\right.
\end{equation}
Subtracting the optimality condition \eqref{opt-cond} from \eqref{s-ed-2-2} gives
\begin{equation}
\left\{
\begin{aligned}
\tws^{t}_{i+1} =&\ \overline{\cA} \Big(\tws^{t}_{i} + \mu [{\nabla \cJ} (\swb^{t}_{i}) \suo - \suo {\nabla \cJ} (\sw^\star)]\Big)  \\
& \hspace{1cm}+ K \cV (\syb^t_{i} \suo - \suo \sy_o^\star) + \mu \tcA\, \s(\swb_{i}^t) \label{s-ed-2-3}\\
\tys^t_{i+1} =&\ \tys^t_{i} \suo - \suo \cV(\swb^{t}_{i+1} \suo - \suo \sw^\star)
\end{aligned}
\right.
\end{equation}
Recall that $\grad \cJ(\sw)$ is twice-differentiable (see Assumption \ref{ass}). We can then appeal  to  the  mean-value theorem (see equations (40)--(43) in \cite{yuan2017exact2}) to express the gradient difference as
\eq{\label{zxcn239sd}
	{\nabla \cJ} (\swb^{t}_{i}) \suo - \suo {\nabla \cJ} (\sw^\star) = -\cHb_i^t \tws_i^t,
}
where $\cHb_i^t$ is defined in \eqref{H_k_i-1}. With \eqref{zxcn239sd}, recursion \eqref{s-ed-2-3} becomes
\begin{equation}
\left\{
\begin{aligned}
\tws^{t}_{i+1} \suo = \suo &\ \overline{\cA} \Big(I_{MK} \suo - \suo \mu \cHb_{i}^t\Big)\tws^{t}_{i} \suo - \suo K \cV \tys^t_{i} \suo + \suo \mu \tcA\, \s(\swb_{i}^t) \label{s-ed-2-4}\\
\tys^t_{i+1} \suo = \suo &\ \tys^t_{i} \suo + \suo \cV\tws^{t}_{i+1}
\end{aligned}
\right.
\end{equation}
From relations \eqref{23hsdbs8} and \eqref{V-definition}, we conclude that $V^2 = (I_K - A)/2K$, which also implies that $\cV^2 = (I_{MK}- \cA)/2K$. With this fact, we substitute the second recursion in \eqref{s-ed-2-4} into the first recursion to get
\begin{equation}
\left\{
\begin{aligned}
\tcA \tws^{t}_{i+1} \ssuo = \suo &\ \overline{\cA} \Big(I_{MK} \ssuo - \ssuo \mu \cHb_{i}^t\Big)\tws^{t}_{i} \suo - \ssuo K \cV \tys^t_{i+1} \ssuo + \suo \mu \tcA\, \s(\swb_{i}^t) \label{s-ed-2-5}\\
\tys^t_{i+1} \ssuo = \suo &\ \tys^t_{i} \suo + \suo \cV\tws^{t}_{i+1}
\end{aligned}
\right.
\end{equation}
which is also equivalent to
\eq{
	&\hspace{-5mm}\ba{cc}
	\tcA    &    K \cV    \\
	- \cV   &    I_{MK}   
	\ea
	\ba{c}
	\tws^{t}_{i+1}   \\
	\tys^t_{i+1}   
	\ea \nnb
	= &\ba{cc}
	\suo \overline{\cA} \Big(I_{MK} \suo - \suo \mu \cHb_{i}^t\Big)\suo & 0    \\
	0    & \suo I_{MK}   \suo 
	\ea
	\suo \suo 
	\ba{c}
	\tws^{t}_{i}\\
	\tys^t_{i}
	\ea
	\ssuo + \ssuo 
	\ba{c}
	\suo \mu \tcA \suo \\
	0
	\ea
	\suo 
	\s(\swb_i^t)\label{zxn2398}.
}
Also recall \eqref{23hsdbs8} that $A = I_K - 2 K U\Sigma U\tran$. Therefore,
\eq{\label{23b88}
	\overline{A} = \frac{I_K + A}{2} = I_K \suo - \suo KU\Sigma U\tran = U(I_K \suo - \suo K\Sigma) U\tran.
}
This together with the fact that $V=U\Sigma^{1/2}U\tran$ leads to
\eq{\label{23asns9}
	& V \overline{A} \suo = \suo U\Sigma^{1/2}U\tran U(I_K  \suo - \suo K \Sigma) U\tran \\
	\suo =\suo & U \Sigma^{1/2}(I_K  \suo - \suo  K \Sigma) U\tran \suo = \suo U (I_K  \suo - \suo K \Sigma)\Sigma^{1/2} U\tran \suo = \suo \overline{A}V,
}
which also implies that $\cV \tcA = \tcA \cV$. As a result, we can verify that 
\eq{
	\ba{cc}
	\tcA & K \cV \\
	- \cV & I_{MK}
	\ea^{-1} =
	\ba{cc}
	I_{MK} & - K\cV\\
	\cV & \tcA
	\ea.
}
Substituting the above relation into \eqref{zxn2398}, we get
\eq{\label{zxn2398-2}
	\ba{c}
	\tws^{t}_{i+1}\\
	\tys^t_{i+1}
	\ea
	\suo = &\suo \ba{cc}
	\overline{\cA} \Big(I_{MK} \suo - \suo \mu \cHb_{i}^t\Big) & -K\cV \\
	\cV \tcA \Big(I_{MK} \suo - \suo \mu \cHb_{i}^t\Big) & \tcA
	\ea
	\ba{c}
	\tws^{t}_{i}\\
	\tys^t_{i}
	\ea
	\nnb
	& + 
	\mu
	\ba{c}
	\tcA\\
	\cV \tcA
	\ea
	\s(\swb_i^t)
}
which matches equations \eqref{error-recursion-0}--\eqref{T-defi}.

\section{Proof of Lemma \ref{lm-transform}} \label{app-transform}
Now We examine the recursion \eqref{zxzan239}. By following the derivation in equations (71)--(77) from \cite{yuan2017exact2}, we have
\eq{\label{b28}
	\cX^{-1} \cTb_{i}^t \cX \ssuo = \ssuo
	\ba{ccc}
	\frac{1}{K}\cI\tran\cHb^t_{i}\cI &  0 & \frac{1}{K}\cI\tran\cHb^t_{i}\cX_{R,u} \\
	0 & 0 &0  \\
	\cX_L\cTb^t_{i}\cR_1 &  \cX_L\cTb^t_{i}\cR_2  &  \cX_L\cTb^t_{i}\cX_R
	\ea,
}
where $\cI \define \mathds{1}_K \otimes I_M$. 
It can also be verified that
\eq{\label{u28s}
	\cX^{-1}
	\cB_l 
	\ssuo \overset{\eqref{X-inverse}}{=}\ssuo 
	\ba{c}
	\hspace{-1mm}\cL_1\tran \hspace{-1mm}\ \\
	\hspace{-1mm}\cL_2\tran \hspace{-1mm}\ \\
	\hspace{-1mm}\cX_L \hspace{-1mm}\ 
	\ea
	\suo \suo 
	\ba{c}
	\tcA \\
	\cV\tcA
	\ea
	\ssuo \overset{\eqref{R and L-kron}}{=}\ssuo 
	\ba{c}
	\cI\tran \tcA/K \\
	\cI\tran \cV \tcA/K \\
	\cX_L \cB_l
	\ea
	\ssuo \suo = \ssuo \suo 
	\ba{c}
	\cI\tran/K\\
	0\\
	\cX_L \cB_l
	\ea,
}
where the last equality holds because
\eq{
	\cI\tran \tcA &= (\mathds{1}_K\tran \overline{A}) \otimes I_M = \mathds{1}_K\tran \otimes I_M = \cI\tran, \\
	\cI\tran \cV \tcA &= (\mathds{1}_K\tran V \overline{A}) \otimes I_M \overset{\eqref{23bsnd8}}{=} 0.
}
Substituting \eqref{b28} and \eqref{u28s} into recursion \eqref{zxzan239}, and also recalling the definition in \eqref{vas}, we get 
\eq{\label{recursion-transform-2}
	&\ba{c}
	\hspace{-1mm}\bar{\sxb}^t_{i+1}\hspace{-1mm}\\
	\hspace{-1mm}\widehat{\sxb}^t_{i+1}\hspace{-1mm}\\
	\hspace{-1mm}\check{\sxb}^t_{i+1} \hspace{-1mm}
	\ea 
	=
	\ba{ccc}
	\hspace{-3mm}I_{\hspace{-0.3mm} M} \hspace{-1.2mm}-\hspace{-1.2mm} \frac{\mu}{K}\cI\tran\hspace{-0.5mm}\cHb^t_{i}\cI &  \hspace{-0.8mm}0 &\hspace{-1.3mm} -\frac{\mu}{K}\cI\tran\hspace{-0.5mm}\cHb^t_{i}\cX_{R,u} \hspace{-2mm}\ \\
	\hspace{-3mm}0 &  \hspace{-1.3mm}I_{\hspace{-0.3mm} M} &\hspace{-3.3mm} 0 \hspace{-2mm}\ \\
	\hspace{2mm}-\mu\cX_L\cTb^t_{i}\cR_1 &  \hspace{-1.3mm}-\mu\cX_L\cTb^t_{i}\cR_2 \hspace{-1.3mm} &  \cD_1 \hspace{-1mm}-\hspace{-1mm} \mu\cX_L\cTb^t_{i}\cX_R \hspace{-2mm}\ 
	\ea \nnb
	& \hspace{2cm} \cdot
	\ba{c}
	\hspace{-1mm}\bar{\sxb}^t_{i}\hspace{-1mm}\\
	\hspace{-1mm}\widehat{\sxb}^t_{i}\hspace{-1mm}\\
	\hspace{-1mm}\check{\sxb}^t_{i}\hspace{-1mm}
	\ea +  \mu
	\ba{c}
	\frac{1}{K}\cI\tran\\
	0\\
	\cX_L \cB_l
	\ea
	\s(\swb_i^t).	
}
Notice that the second line of the above recursion is
\eq{\label{9sm}
	\widehat{\sxb}^t_{i+1} = \widehat{\sxb}^t_{i}.
}
As a result, $\widehat{\sxb}^t_{i+1}$ will stay at $0$ if the initial value $\widehat{\sxb}^0_{0} = 0$. From \eqref{vas} we can derive that
\eq{\label{9sdj}
	\widehat{\sxb}^0_{0} \suo \overset{\eqref{vas}}{=} \suo \ssuo \cL_2\tran
	\ba{c}
	\tws_0^0 \vspace{0.5mm}\\
	\tys_0^0
	\ea 
	\ssuo \overset{\eqref{R and L-kron}}{=} \ssuo 
	\frac{1}{K}\cI\tran(\sy_o \suo - \suo \syb_0^0) 
	\suo \overset{(a)}{=} \suo 
	\frac{1}{K}\cI\tran \sy_o 
	\suo \overset{(b)}{=} \suo 
	0,
}
where equality (a) holds because $\syb_0^0 = 0$. Equality (b) holds because $\sy_o$ lies in the range space of $\cV$ (see Section \ref{sec-opt-cond}) and $\cI\tran \cV = 0$ (see \eqref{23bsnd8}). Therefore, with \eqref{9sm} and \eqref{9sdj}, we conclude that
\eq{\label{239k}
	\widehat{\sxb}^t_{i} = 0,\quad  \ 0 \le i\le \widebar{N}-1,\ t\ge 0.
}
With \eqref{239k}, the transformed error recursion \eqref{recursion-transform-2} reduces to
\eq{\label{23nsad}
	\ba{c}
	\hspace{-2mm}\bar{\sxb}^t_{i+1}\hspace{-2mm}\\
	\hspace{-2mm}\check{\sxb}^t_{i+1}\hspace{-2mm}
	\ea 
	=
	&\ba{cc}
	\hspace{-2mm}I_{\hspace{-0.3mm} M} \suo - \suo \frac{\mu}{K}\cI\tran\hspace{-0.5mm}\cHb^t_{i}\cI & -\frac{\mu}{K}\cI\tran\cHb^t_{i}\cX_{R,u}\hspace{-2mm}\\
	\hspace{-2mm}- \mu\cX_L \cTb^t_{i} \cR_1 & \cD_1 \suo - \suo \mu \cX_L \cTb^t_{i} \cX_R \hspace{-2mm}
	\ea \hspace{-1.5mm}
	\ba{c}
	\hspace{-2mm}\bar{\sxb}^t_{i}\hspace{-2mm}\\
	\hspace{-2mm}\check{\sxb}^t_{i}\hspace{-2mm}
	\ea
	\nnb
	&+ \mu
	\ba{c}
	\frac{1}{K}\cI\tran \hspace{0.5mm}\\
	\cX_L \cB_l
	\ea \s(\swb_{i}^t),
}
while \eqref{vas} reduces to
\eq{\label{vas1}
	\ba{c}
	\tws_i^t \vspace{0.5mm}\\
	\tys_i^t
	\ea
	\ssuo = \ssuo \cX
	\ba{c}
	\hspace{-1mm}\bar{\sxb}^t_{i}\hspace{-1mm}\vspace{0.7mm}\\
	\hspace{-1mm}0_M\hspace{-1mm}\\
	\hspace{-1mm}\check{\sxb}^t_{i} \hspace{-1mm}
	\ea.
}

\section{Proof of Lemma \ref{lm-mse-recursion}}\label{app-lm-mse}
Since $Q(w;x_n)$ is twice-differentiable, it follows from \eqref{lsc} that $\grad_w^2 Q(w;x_n) \le \delta I_M$ for $1\le n \le N$, which in turn implies that
\eq{
	\grad^2 J_k(w) \suo = \suo \frac{1}{N_k}\sum_{n=1}^{N_k}\grad Q(w;x_{k,n}) \le \delta I_M, 
	\forall\ k\in \{1,\cdots, K\} \label{bw8}
}
Moreover, since all $Q(w;x_n)$ are convex and at least one $Q(w;x_{n_o})$ is strongly convex {\color{black}(see equation \eqref{strongly-convex}}, there must exist at least one node $k_o$ such that
\eq{
	\grad^2 J_{k_o}(w) = \frac{1}{N_{k_o}}\sum_{n=1}^{N_{k_o}} \grad_w^2 Q(w;x_{k_o,n}) \ge \nu I_M, \label{zxn2sdjhaf9}
}
which implies that the global risk function, $J(w)$, is $\nu$-strongly convex as well. Substituting \eqref{bw8} and \eqref{zxn2sdjhaf9} into $\H_{k,i}^t$ defined in \eqref{H_k_i-1}, for $t\ge0$ and $0\le i \le \widebar{N}-1$ it holds that
\eq{
	\H_{k,i}^t &\overset{\eqref{H_k_i-1}}{=} \int_0^1 \grad^2 J_k\big(w^\star \hspace{-0.5mm}-\hspace{-0.5mm} r\widetilde{\w}_{k,i}^t\big)dr 
	\overset{\eqref{bw8}}{\le} \delta I_M, \forall k \in \{1,\cdots, K\} \label{Hki-1}\\
	\H_{k_o,i}^t &\overset{\eqref{H_k_i-1}}{=} \ssuo \ssuo \int_0^1 \grad^2 J_{k_o}\big(w^\star \hspace{-0.5mm}-\hspace{-0.5mm} r\widetilde{\w}_{k_o,i}^t\big)dr \ssuo  \overset{\eqref{zxn2sdjhaf9}}{\ge} \ssuo \nu I_M, \label{Hki-2} \\
	\cHb_i^t &\overset{\eqref{H_k_i-1}}{=} \diag\{ \H_{1,i}^t, \cdots, \H_{K,i}^t \} \overset{\eqref{Hki-1}}{\le} \delta I_M.
}
%
%
%
Now we turn to derive the mean-square-error recursion. From the first line of error recursion \eqref{23nsad-2}, we have
\eq{\label{first-line}
	\bar{\sxb}^t_{i+1} 
	= &\left(I_{\hspace{-0.3mm} M} \suo - \suo \frac{\mu}{K}\cI\tran \cHb^t_{i}\cI \right)\bar{\sxb}_{i}^t \nnb
	&\hspace{1cm}- \frac{\mu}{K} \left( \cI\tran\cHb^t_{i}\cX_{R,u} \right) \check{\sxb}_{i}^t + \frac{\mu}{K}\cI\tran \s(\swb^t_{i}).
}
Recalling that $\cI = \mathds{1}_K \otimes I_M$, it holds that
\eq{\label{xzn27}
	\frac{1}{K}\cI\tran \cHb^t_{i}\cI = \frac{1}{K}\sum_{k=1}^{K}\H_{k,i}^t.
}
Substituting relations \eqref{Hki-1} and \eqref{Hki-2} into \eqref{xzn27}, it holds that
\eq{\label{23nsd9}
	\frac{\nu}{K} I_M \le \frac{1}{K}\cI\tran \cHb^t_{i}\cI \le \delta I_M,
}
which also implies that 
\eq{\label{zx238}
	\left\|I_{\hspace{-0.3mm} M} \suo - \suo \frac{\mu}{K}\cI\tran \cHb^t_{i}\cI \right\|^2 
	&\le \max\left\{ \left(1 \suo - \suo \frac{\mu \nu}{K} \right)^2, (1- \mu \delta)^2 \right\} \nnb
	&\le \left(1 \suo - \suo \frac{\mu \nu}{K} \right)^2,
}
where the last inequality holds when the step-size $\mu$ is small enough so that
\eq{
	\mu < 1/\delta.
}
Now we square both sides of equation \eqref{first-line} and reach
\eq{
	&\|\bar{\sxb}^t_{i+1}\|^2 \nnb
	&\suo=\suo\left\| \left(I_{\hspace{-0.3mm} M} \ssuo -\ssuo \frac{\mu}{K}\cI\tran \cHb^t_{i}\cI \right)\ssuo \bar{\sxb}_{i}^t \suo-\suo \frac{\mu}{K}\ssuo\left( \cI\tran\cHb^t_{i}\cX_{R,u} \right) \check{\sxb}_{i}^t\suo+\suo\frac{\mu}{K}\cI\tran \s(\swb^t_{i})\right\|^2 \nnb
	&\ssuo\overset{(a)}{=}\ssuo\left\| (1-t) \frac{1}{1-t} \left(I_{\hspace{-0.3mm} M} \hspace{-0.6mm}-\hspace{-0.6mm} \frac{\mu}{K}\cI\tran \cHb^t_{i}\cI \right)\bar{\sxb}_{i}^t \right. \nnb
	&\qquad \left. + t\frac{1}{t}\Big[ \hspace{-0.6mm}-\hspace{-0.6mm} \frac{\mu}{K} \left( \cI\tran\cHb^t_{i}\cX_{R,u} \right) \check{\sxb}_{i}^t \hspace{-0.6mm}+\hspace{-0.6mm} \frac{\mu}{K}\cI\tran \s(\swb^t_{i}) \Big]\right\|^2 \nnb
	&\overset{(b)}{\le} \frac{1}{1-t} \left\|I_{\hspace{-0.3mm} M} - \frac{\mu}{K}\cI\tran \cHb^t_{i}\cI \right\|^2 \|\bar{\sxb}_{i}^t\|^2  \nnb
	&\qquad + \frac{1}{t} \left\|\frac{\mu}{K} \left( \cI\tran\cHb^t_{i}\cX_{R,u} \right) \check{\sxb}_{i}^t + \frac{\mu}{K}\cI\tran \s(\swb^t_{i}) \right\|^2  \nnb
	&\overset{(c)}{\le} \frac{1}{1-t} \left\|I_{\hspace{-0.3mm} M} \hspace{-0.8mm}-\hspace{-0.8mm} \frac{\mu}{K}\cI\tran \cHb^t_{i}\cI \right\|^2 \hspace{-0.8mm} \|\bar{\sxb}_{i}^t\|^2  \nnb
	&\qquad + \frac{2\mu^2}{t K^2}\left\| \cI\tran\cHb^t_{i}\cX_{R,u} \right\|^2 \hspace{-0.8mm} \|\check{\sxb}_{i}^t\|^2 \hspace{-0.8mm}+\hspace{-0.8mm} \frac{2\mu^2}{tK^2} \|\cI\tran\|^2 \left\| \s(\swb^t_{i})\right\|^2 \nnb
	&\overset{(d)}{\le} \frac{1}{1-t} \left(1 - \frac{\mu \nu}{K} \right)^2 \|\bar{\sxb}_{i}^t\|^2 \nnb 
	&\qquad + \frac{2\mu^2\delta^2\|\cX_{R,u}\|^2}{Kt} \|\check{\sxb}_{i}^t\|^2 + \frac{2\mu^2}{Kt} \left\| \s(\swb^t_{i})\right\|^2 \nnb
	&\overset{(e)}{=} \left(1 - \frac{\mu \nu}{K} \right) \|\bar{\sxb}_{i}^t\|^2 + \frac{2\mu\delta^2\|\cX_{R,u}\|^2}{\nu} \|\check{\sxb}_{i}^t\|^2 + \frac{2\mu}{\nu} \left\| \s(\swb^t_{i})\right\|^2 \label{23am99}
}
where equality (a) holds for any constant $t\in(0,1)$, inequality (b) holds because of the Jensen's inequality, inequality (c) holds because $\|a+b\|^2 \le 2\|a\|^2 + 2\|b\|^2$ for any two vectors $a$ and $b$, and inequality (d) holds because of relation \eqref{zx238} and
\eq{\label{zxcn239}
	\|\cI\tran\|^2 &= K, \\
	\left\| \cI\tran\cHb^t_{i}\cX_{R,u} \right\|^2 \ssuo &\le \ssuo \|\cI\tran\|^2 \|\cHb^t_{i}\|^2 \|\cX_{R,u}\|^2 \ssuo \le \ssuo K \delta^2 \|\cX_{R,u}\|^2.
}
Equality (e) holds when $t = \mu\nu/K$.

Next we turn to the second line of recursion \eqref{23nsad-2}:
\eq{
	\check{\sxb}_{i+1}^t \ssuo = \suo \cD_1 \check{\sxb}_i^t \suo
	- \suo \mu \Big( \cX_L \cTb_{i}^t \cR_1 \bar{\sxb}_{i}^t \suo + \suo \cX_L \cTb_{i}^t \cX_R \check{\sxb}_{i}^t \suo - \suo \cX_L\cB_l \s(\swb_{i}^t) \Big)
}
By squaring and applying Jensen's inequality, we have
\eq{\label{second-line}
	\|\check{\sxb}_{i+1}^t\|^2 
	&\le \frac{1}{t}\|\cD_1\|^2 \|\check{\sxb}_i^t\|^2
	\hspace{-0.8mm}+\hspace{-0.8mm} \frac{3\mu^2}{1-t} \Big( \|\cX_L \cTb_{i}^t \cR_1\|^2 \|\bar{\sxb}_{i}^t\|^2  \hspace{-0.8mm} \nnb
	&\qquad + \|\cX_L \cTb_{i}^t \cX_R\|^2\|\check{\sxb}_{i}^t\|^2 \hspace{-0.8mm}+\hspace{-0.8mm} \|\cX_L\cB_l\|^2 \|\s(\swb_{i}^t)\|^2 \Big)
}
for any constant $t\in(0,1)$. From the definition of $\cTb_i^t$ in \eqref{T-defi} and recalling from \eqref{23b88} that $\tcA \cV = \cV \tcA$, we have
\eq{\label{zn23sdk}
	\cTb_i^t =
	\ba{cc}
	\tcA & 0\\
	0 & \tcA
	\ea
	\ba{cc}
	I_{KM} & 0\\
	\cV &0
	\ea
	\ba{cc}
	\cHb_i^t & 0\\
	0 & \cHb_i^t
	\ea.
}
It can also be verified that
\eq{\label{zxnqw9}
	&\hspace{-5mm}\left\|
	\ba{cc}
	I_{KM} & 0 \\
	\cV & 0
	\ea
	\right\|^2 \nnb
	&= \lambda_{\max} \left( \ba{cc}
	I_{KM} & 0 \\
	\cV & 0
	\ea\tran \ba{cc}
	I_{KM} & 0 \\
	\cV & 0
	\ea\right) \nnb
	&= \lambda_{\max}\left(
	\ba{cc}
	I_{KM} + \cV^2 & 0\\
	0 & 0
	\ea
	\right) \nnb
	&= \lambda_{\max} \left( I_{KM} + \frac{I_{KM} \suo - \suo \tcA}{2K}\right) \le 2
}
where the last inequality holds because $0 < \lambda(\tcA) \le 1$. With \eqref{zn23sdk}, \eqref{zxnqw9} and the facts that $\lambda_{\max}(\tcA) = 1$, $\lambda_{\max}(\cHb_i^t) \le \delta$, we conclude that
\eq{
	\|\cTb_i^t\|^2 \ssuo \le \ssuo  
	\left\|
	\ba{cc}
	\tcA & 0\\
	0 & \tcA
	\ea
	\right\|^2 \ssuo
	\left\|
	\ba{cc}
	I_{KM} & 0\\
	\cV &0
	\ea
	\right\|^2 \ssuo 
	\left\|
	\ba{cc}
	\cHb_i^t & 0\\
	0 & \cHb_i^t
	\ea
	\right\|^2 \ssuo \le \ssuo 2\delta^2.
	\label{b2s0}
}
{\color{black}
	Similarly, using $\tcA \cV = \cV \tcA$ we can rewrite $\cB_l$ defined in \eqref{T-defi} as
	\eq{
		\cB_l = 
		\ba{cc}
		\tcA & 0\\
		0 & \tcA
		\ea
		\ba{c}
		I_{KM}\\
		\cV
		\ea,
	}
	and it can be verified that 
	\eq{
		\left\|
		\ba{c}
		I_{KM}\\
		\cV
		\ea
		\right\|^2 &= \lambda_{\max}
		\left(
		\ba{c}
		I_{KM}\\
		\cV
		\ea\tran 
		\ba{c}
		I_{KM}\\
		\cV
		\ea
		\right) \nnb
		&= \lambda_{\max}\left(I_{KM} + \cV^2\right) \nnb
		&= \lambda_{\max} \left( I_{KM} + \frac{I_{KM} \suo - \suo \tcA}{2K}\right) \le 2. 
	}
	As a result,
	\eq{\label{b2s1}
		\|\cB_l\|^2 \le
		\left\|
		\ba{cc}
		\tcA & 0\\
		0 & \tcA
		\ea
		\right\|^2
		\left\|
		\ba{c}
		I_{KM}\\
		\cV
		\ea
		\right\|^2 \le 2.
	}}
	Furthermore,
	\eq{\label{b2s2}
		\|\cR_1\|^2 &= \left\| \ba{c} \mathds{1}_K \\ 0 \ea \otimes I_M \right\|^2 \nnb
		&= \lambda_{\max}\left( \ba{c} \mathds{1}_K \\ 0 \ea\tran \ba{c} \mathds{1}_K \\ 0 \ea \otimes I_M \right) = K.
	}
	With \eqref{b2s0}--\eqref{b2s2}, we have
	\eq{\label{zcxmb}
		\|\cX_L \cTb_{i}^t \cR_1\|^2 &\le \|\cX_L\|^2 \|\cTb_{i}^t\|^2 \|\cR_1\|^2 \le 2K\delta^2\|\cX_L\|^2, \\
		\|\cX_L \cTb_{i}^t \cX_R\|^2 &\le 2\delta^2 \|\cX_L\|^2 \|\cX_R\|^2, \\
		\|\cX_L \cB_l\|^2 &\le 2\|\cX_L\|^2.
	}
	Substituting \eqref{zcxmb} into \eqref{second-line} and recalling that $\|\cD_1\|=\lambda < 1$, we have
	\eq{\label{second-line-1}
		&\|\check{\sxb}_{i+1}^t\|^2 \nnb
		&\le \frac{1}{t}\lambda^2 \|\check{\sxb}_i^t\|^2
		\hspace{-0.8mm}+\hspace{-0.8mm} \frac{3\mu^2}{1-t} \Big( 2K\delta^2\|\cX_L\|^2 \|\bar{\sxb}_{i}^t\|^2  \nnb
		&\qquad + 2\delta^2 \|\cX_L\|^2 \|\cX_R\|^2\|\check{\sxb}_{i}^t\|^2 \hspace{-0.8mm}+\hspace{-0.8mm} 2\|\cX_L\|^2 \|\s(\swb_{i}^t)\|^2 \Big) \nnb
		&= \left( \lambda + \frac{6\mu^2\delta^2 \|\cX_L\|^2 \|\cX_R\|^2}{1-\lambda} \right) \|\check{\sxb}_i^t\|^2 \nnb
		&\qquad + \suo \frac{6K\mu^2 \delta^2\|\cX_L\|^2}{1-\lambda}\|\bar{\sxb}_{i}^t\|^2 \suo + \suo \frac{6\|\cX_L\|^2\mu^2}{1-\lambda} \|\s(\swb_{i}^t)\|^2,
	}
	where the last equality holds by setting $t = \lambda$.
	If we let
	\eq{\label{a-notation}
		&\ a_1 = 1/K,\ a_2 = \|\cX_{R,u}\|^2,\ a_3 = \frac{6 \|\cX_L\|^2 \|\cX_R\|^2}{1-\lambda}, \nnb 
		&\ a_4 = \frac{6K\|\cX_L\|^2}{1-\lambda},\ a_5 = \frac{6\|\cX_L\|^2}{1-\lambda}
	}
	and take expectations of inequalities \eqref{second-line} and \eqref{second-line-1}, we arrive at recursion \eqref{xngyi}, where $a_l,\ 1\le l\le 5$ are positive constants that are independent of $\widebar{N}$, $\delta$ and $\nu$.
	
	\section{Proof of Lemma \ref{lm-gradient-noise}}\label{app-gradient-noise}
	We first {\color{black} introduce the gradient noise at node $k$:
		\eq{\label{n28sh}
			\s_k(\w_{k,i}^t) \define \widehat{\grad J}_k(\w_{k,i}^t) \suo - \suo \grad J_k(\w_{k,i}^t).
		}
		With \eqref{n28sh} and \eqref{noise}, we have
		\eq{\label{n2hsg9}
			\s(\swb_i^t) = \col\{\s_1(\w_{1,i}^t), \s_2(\w_{2,i}^t),\cdots,\s_N(\w_{N,i}^t)\}.
		}
		Now we bound the term $\|\s_k(\w_{k,i}^t)\|^2$. Note that
	}
	\eq{\label{zxn238}
		&\s_k(\w_{k,i}^t) \nnb 
		&= \widehat{\grad J}_k(\w_{k,i}^t) \suo - \suo \grad J_k(\w_{k,i}^t) \nnb &\ \hspace{-1mm}\overset{\eqref{sgd-3}}{=} \grad Q(\w^t_{k,i};x_{k,\n_{k,i}^t} ) \suo - \suo \grad Q(\w^t_{k,0};x_{k,\n_{k,i}^t} ) + \g_k^t \suo - \suo \grad J_k(\w_{k,i}^t) \nnb
		& \overset{\eqref{approximate-gradeint}}{=} \grad Q(\w^t_{k,i};x_{k,\n_{k,i}^t} ) \suo - \suo \grad Q(\w^t_{k,0};x_{k,\n_{k,i}^t} ) \nnb
		&\quad\quad  + \frac{1}{\widebar{N}}\sum_{j=0}^{\widebar{N}-1} \grad Q\left(\w_{k,j}^{t-1};x_{k,\n_{k,j}^{t-1}}\right) \suo - \suo \frac{1}{\widebar{N}}\sum_{n=1}^{\widebar{N}}\grad Q\left(\w_{k,i}^{t};x_{k,n}\right)
	}
	Since $\n_{k,j}^{t-1}=\bsigma^{t-1}(j+1)$ is sampled by random reshuffling without replacement, it holds that
	\eq{\label{139asdl}
		\sum_{j=0}^{\widebar{N}-1} \grad Q\left(\w_{k,\widebar{N}}^{t-1};x_{k,\n_{k,j}^{t-1}}\right) 
		&= \sum_{n=1}^{\widebar{N}} \grad Q\left(\w_{k,\widebar{N}}^{t-1};x_{k,n}\right) \nnb
		&\overset{(a)}{=}  \sum_{n=1}^{\widebar{N}} \grad Q\left(\w_{k,0}^{t};x_{k,n}\right)
	}
	where equality (a) holds because $\w_{k,0}^{t} = \w_{k,\widebar{N}}^{t-1}$. With relation \eqref{139asdl}, we can rewrite \eqref{zxn238} as
	\eq{\label{zxn238-1}
		&\s_k(\w_{k,i}^t) \nnb
		&= \grad Q(\w^t_{k,i};x_{k,\n_{k,i}^t} ) \suo - \suo \grad Q(\w^t_{k,0};x_{k,\n_{k,i}^t} ) \nnb
		&\quad + \frac{1}{\widebar{N}}\sum_{j=0}^{\widebar{N}-1} \grad Q\left(\w_{k,j}^{t-1};x_{k,\n_{k,j}^{t-1}}\right) \suo - \suo \frac{1}{\widebar{N}}\sum_{j=0}^{\widebar{N}-1} \grad Q\left(\w_{k,\widebar{N}}^{t-1};x_{k,\n_{k,j}^{t-1}}\right)\nnb
		&\quad  + \frac{1}{\widebar{N}}\sum_{n=1}^{\widebar{N}} \grad Q\left(\w_{k,0}^{t};x_{k,n}\right)  \suo - \suo \frac{1}{\widebar{N}}\sum_{n=1}^{\widebar{N}}\grad Q\left(\w_{k,i}^{t};x_{k,n}\right)
	}
	By squaring and applying Jensen's inequality, we have
	\eq{\label{zxn238-2}
		&\|\s_k(\w_{k,i}^t)\|^2 \nnb
		&\le 3\left\|\grad Q(\w^t_{k,i};x_{k,\n_{k,i}^t} ) \suo - \suo \grad Q(\w^t_{k,0};x_{k,\n_{k,i}^t} )\right\|^2 \nnb
		& \quad + \frac{3}{\widebar{N}}\sum_{j=0}^{\widebar{N}-1} \left\| \grad Q\left(\w_{k,j}^{t-1};x_{k,\n_{k,j}^{t-1}}\right) \suo - \suo  \grad Q\left(\w_{k,\widebar{N}}^{t-1};x_{k,\n_{k,j}^{t-1}}\right) \right\|^2\nnb
		& \quad + \frac{3}{\widebar{N}}\sum_{n=1}^{\widebar{N}} \left\| \grad Q\left(\w_{k,0}^{t};x_{k,n}\right)  \suo - \suo \grad Q\left(\w_{k,i}^{t};x_{k,n}\right) \right\|^2 \nnb
		&\le 6\delta^2\|\w_{k,i}^t \suo - \suo \w_{k,0}^t\|^2 + \frac{3\delta^2}{\widebar{N}}\sum_{j=0}^{\widebar{N}-1}\left\| \w_{k,j}^{t-1} \suo - \suo \w_{k,\widebar{N}}^{t-1} \right\|^2
	}
	where the last inequality holds because of the Lipschitz inequality \eqref{lsc} in Assumption 1. Consequently,
	\eq{\label{zb9sdjk}
		&\|\s(\swb_i^t)\|^2 \nnb 
		&\overset{\eqref{n2hsg9}}{=} \sum_{k=1}^{K} \|\s_k(\w_{k,i}^t)\|^2 \nnb
		&\le 6\delta^2\sum_{k=1}^{K}\|\w_{k,i}^t \suo - \suo \w_{k,0}^t\|^2 + \frac{3\delta^2}{\widebar{N}}\sum_{j=0}^{\widebar{N}-1}\sum_{k=1}^{K}\left\| \w_{k,j}^{t-1} \suo - \suo \w_{k,\widebar{N}}^{t-1} \right\|^2 \nnb
		&= 6\delta^2\|\swb_i^t \suo - \suo \swb_0^t\|^2 + \frac{3\delta^2}{\widebar{N}}\sum_{j=0}^{\widebar{N}-1}\left\| \swb_{j}^{t-1} \suo - \suo \swb_{\widebar{N}}^{t-1} \right\|^2 \nnb
		&= 6\delta^2\|\tws_i^t \suo - \suo \tws_0^t\|^2 + \frac{3\delta^2}{\widebar{N}}\sum_{j=0}^{\widebar{N}-1}\left\| \tws_{j}^{t-1} \suo - \suo \tws_{\widebar{N}}^{t-1} \right\|^2 \nnb
		&\le 6\delta^2 \hspace{-0.8mm}\left(\hspace{-0.8mm} \|\tws_i^t \hspace{-0.8mm}-\hspace{-0.8mm} \tws_0^t\|^2 \hspace{-0.8mm}+\hspace{-0.8mm} \|\tys_i^t \hspace{-0.8mm}-\hspace{-0.8mm} \tys_0^t\|^2 \right)\nnb
		&\quad +\hspace{-0.8mm} \frac{3\delta^2}{\widebar{N}}\sum_{j=0}^{\widebar{N}-1}\left( \left\| \tws_{j}^{t-1} \hspace{-0.8mm}-\hspace{-0.8mm} \tws_{\widebar{N}}^{t-1} \right\|^2 \hspace{-1.2mm}+\hspace{-0.8mm} \left\| \tys_{j}^{t-1} \hspace{-0.8mm}-\hspace{-0.8mm} \tys_{\widebar{N}}^{t-1} \right\|^2\right).
	}
	Now note that
	\eq{
		&\|\tws_i^t \hspace{-0.8mm}-\hspace{-0.8mm} \tws_0^t\|^2 \hspace{-0.8mm}+\hspace{-0.8mm} \|\tys_i^t \hspace{-0.8mm}-\hspace{-0.8mm} \tys_0^t\|^2 \nnb
		&= \left\|
		\ba{c}
		\tws_i^t\\
		\tys_i^t
		\ea -
		\ba{c}
		\tws_0^t\\
		\tys_0^t
		\ea
		\right\|^2
		\overset{\eqref{w-x-relation}}{\le} \|\cX\|^2
		\left\|
		\ba{c}
		\hspace{-1mm}\bar{\sxb}^t_{i}\hspace{-1mm}\vspace{0.7mm}\\
		\hspace{-1mm}0_M\hspace{-1mm}\\
		\hspace{-1mm}\check{\sxb}^t_{i} \hspace{-1mm}
		\ea
		-
		\ba{c}
		\hspace{-1mm}\bar{\sxb}^t_{0}\hspace{-1mm}\vspace{0.7mm}\\
		\hspace{-1mm}0_M\hspace{-1mm}\\
		\hspace{-1mm}\check{\sxb}^t_{0} \hspace{-1mm}
		\ea
		\right\|^2 \nnb
		&= \|\cX\|^2  \left( \|\bar{\sxb}^t_{i} \suo - \suo \bar{\sxb}^t_{0}\|^2 + \|\check{\sxb}^t_{i} \suo - \suo \check{\sxb}^t_{0} \|^2 \right) \nnb
		&\le \|\cX\|^2 \|\bar{\sxb}^t_{i} \suo - \suo \bar{\sxb}^t_{0}\|^2 + 2\|\cX\|^2\|\check{\sxb}^t_{i}\|^2 + 2\|\cX\|^2\|\check{\sxb}^t_{0}\|^2
		\label{zxcn99}
	}
	Similarly, it holds that
	\eq{\label{b239}
		&\quad \left\| \tws_{j}^{t-1} \hspace{-0.8mm}-\hspace{-0.8mm} \tws_{\widebar{N}}^{t-1} \right\|^2 \hspace{-1.2mm}+\hspace{-0.8mm} \left\| \tys_{j}^{t-1} \hspace{-0.8mm}-\hspace{-0.8mm} \tys_{\widebar{N}}^{t-1} \right\|^2 \nnb
		&\le \|\cX\|^2 \|\bar{\sxb}^{t-1}_{j} \hspace{-0.8mm}-\hspace{-0.8mm} \bar{\sxb}^{t-1}_{\widebar{N}}\|^2 \hspace{-0.8mm}+\hspace{-0.8mm} 2\|\cX\|^2\|\check{\sxb}^{t-1}_{j}\|^2 \hspace{-0.8mm}+\hspace{-0.8mm} 2\|\cX\|^2\|\check{\sxb}^t_{0}\|^2.
	}
	Substituting \eqref{zxcn99} and \eqref{b239} into \eqref{zb9sdjk} and letting $b = \|\cX\|^2$, we have
	\eq{
		\|\s(\swb_i^t)\|^2 &\le 6 b \delta^2\|\bar{\sxb}^t_{i} \suo - \suo \bar{\sxb}^t_{0}\|^2 + 12 b \delta^2 \|\check{\sxb}^t_{i}\|^2 + 18 b \delta^2 \|\check{\sxb}^t_{0}\|^2 \nnb
		&  + \frac{3b\delta^2}{\widebar{N}}\sum_{j=0}^{\widebar{N}-1}\|\bar{\sxb}^{t-1}_{j} \hspace{-0.8mm}-\hspace{-0.8mm} \bar{\sxb}^{t-1}_{\widebar{N}}\|^2 + \frac{6b\delta^2}{\widebar{N}}\sum_{j=0}^{\widebar{N}-1}\|\check{\sxb}^{t-1}_{j}\|^2
	}
	By taking expectations, we achieve inequality \eqref{zx23b9}.
	
	\section{Proof of Lemma \ref{lm-check-x}}\label{app-check-x}
	{\color{black}It is established in Lemma \ref{lm-mse-recursion} that when step-size $\mu$ satisfies 
		\eq{\label{trivial-stepsize}
			\mu < \frac{1}{\delta},
		}
		the dynamic system \eqref{xngyi} holds.} Using Jensen's inequality, the second line of \eqref{xngyi} becomes
	\eq{\label{b139dj}
		&\bE\|\check{\sxb}_{i+1}^t\|^2 \nnb
		&\le \left( \lambda \hspace{-0.6mm}+\hspace{-0.6mm} a_3 \mu^2 \delta^2 \right)\bE\|\check{\sxb}_i^t\|^2 \hspace{-0.6mm}+\hspace{-0.6mm} 2a_4\mu^2\delta^2\bE\|\bar{\sxb}_i^t \hspace{-0.6mm}-\hspace{-0.6mm} \bar{\sxb}_0^t\|^2 \nnb &\quad +\hspace{-0.6mm} 2a_4\mu^2\delta^2\bE\| \bar{\sxb}_0^t\|^2 \hspace{-0.6mm}+\hspace{-0.6mm} a_5 \mu^2\bE\|\s(\swb_i^t)\|^2 \nnb
		&\overset{\eqref{zx23b9}}{\le} \Big(\lambda + (a_3 + 12 a_5b) \mu^2\delta^2  \Big)\bE\|\check{\sxb}_i^t\|^2 \nnb 
		&\quad + (2a_4 + 6 a_5 b) \mu^2 \delta^2 \bE\|\bar{\sxb}_i^t \suo - \suo \bar{\sxb}_0^t\|^2 + 2a_4\mu^2\delta^2\bE\| \bar{\sxb}_0^t\|^2 \nnb 
		&\quad + 18 a_5 b \mu^2 \delta^2\bE\|\check{\sxb}^t_{0}\|^2 + \frac{3a_5b\mu^2\delta^2}{\widebar{N}}\sum_{j=0}^{\widebar{N}-1}\bE\|\bar{\sxb}^{t-1}_{j} \hspace{-0.8mm}-\hspace{-0.8mm} \bar{\sxb}^{t-1}_{\widebar{N}}\|^2 \nnb 
		&\quad + \frac{6a_5b\mu^2\delta^2}{\widebar{N}}\sum_{j=0}^{\widebar{N}-1}\bE\|\check{\sxb}^{t-1}_{j}\|^2.
	}
	%
	%
	Now we let $\lambda_1 = (1+\lambda)/2 < 1$. It can be verified that when the step-size $\mu$ is small enough so that
	\eq{\label{zbwn999}
		\mu \le \sqrt{\frac{1-\lambda}{2(a_3+12a_5b)\delta^2}},
	}
	it holds that
	\eq{\label{xan29}
		\lambda + (a_3 + 12 a_5b) \mu^2\delta^2 \le \lambda_1 < 1.
	}
	Substituting \eqref{xan29} into \eqref{b139dj}, we have
	\eq{\label{b139dj-1}
		&\bE\|\check{\sxb}_{i+1}^t\|^2 \nnb
		&\le \lambda_1 \bE\|\check{\sxb}_i^t\|^2 \hspace{-0.6mm}+\hspace{-0.6mm} (2a_4 \hspace{-0.6mm}+\hspace{-0.6mm} 6 a_5 b) \mu^2 \delta^2 \bE\|\bar{\sxb}_i^t \suo - \suo \bar{\sxb}_0^t\|^2 \nnb &\quad +\hspace{-0.6mm} 2a_4\mu^2\delta^2\bE\| \bar{\sxb}_0^t\|^2 \hspace{-0.6mm}+\hspace{-0.6mm} 18 a_5 b \mu^2 \delta^2\bE\|\check{\sxb}^t_{0}\|^2\nnb
		&\quad + \frac{3a_5b\mu^2\delta^2}{\widebar{N}} \ssuo \sum_{j=0}^{\widebar{N}-1} \ssuo \bE\|\bar{\sxb}^{t-1}_{j} \hspace{-0.8mm}-\hspace{-0.8mm} \bar{\sxb}^{t-1}_{\widebar{N}}\|^2 \ssuo + \ssuo \frac{6a_5b\mu^2\delta^2}{\widebar{N}}\ssuo\sum_{j=0}^{\widebar{N}-1}\ssuo\bE\|\check{\sxb}^{t-1}_{j}\|^2.
	}
	Iterating \eqref{b139dj-1}, for $0\le i \le \widebar{N}-1$, we get
	\eq{\label{b139dj-2}
		&\bE\|\check{\sxb}_{i+1}^t\|^2 \nnb
		&\le \lambda_1^{i+1} \bE\|\check{\sxb}_0^t\|^2 + (2a_4 + 6 a_5 b) \mu^2 \delta^2 \sum_{j=0}^{i}\lambda_1^{i-j} \bE\|\bar{\sxb}_j^t \suo - \suo \bar{\sxb}_0^t\|^2 \nnb
		&\quad + \Big( 2a_4\mu^2\delta^2\bE\| \bar{\sxb}_0^t\|^2+ 18 a_5 b \mu^2 \delta^2\bE\|\check{\sxb}^t_{0}\|^2  \Big)\sum_{j=0}^{i}\lambda_1^{i-j} \nnb 
		&\quad + \left(\frac{3a_5b\mu^2\delta^2}{\widebar{N}}\sum_{j=0}^{\widebar{N}-1}\bE\|\bar{\sxb}^{t-1}_{j} \hspace{-0.8mm}-\hspace{-0.8mm} \bar{\sxb}^{t-1}_{\widebar{N}}\|^2 \right. \nnb
		&\qquad \quad + \left. \frac{6a_5b\mu^2\delta^2}{\widebar{N}} \sum_{j=0}^{\widebar{N}-1}\bE\|\check{\sxb}^{t-1}_{j}\|^2 \right)\sum_{j=0}^{i}\lambda_1^{i-j} \nnb
		&\overset{(a)}{\le} \lambda_1^{i+1} \bE\|\check{\sxb}_0^t\|^2 + (2a_4 + 6 a_5 b) \mu^2 \delta^2 \sum_{j=0}^{i} \bE\|\bar{\sxb}_j^t \suo - \suo \bar{\sxb}_0^t\|^2 \nnb
		&\quad +  2a_4\mu^2\delta^2 (i+1) \bE\| \bar{\sxb}_0^t\|^2 +  18 a_5 b \mu^2 \delta^2 (i+1) \bE\|\check{\sxb}^t_{0}\|^2 \nnb
		&\quad + \frac{3a_5b\mu^2\delta^2(i+1)}{\widebar{N}}\sum_{j=0}^{\widebar{N}-1}\bE\|\bar{\sxb}^{t-1}_{j} \hspace{-0.8mm}-\hspace{-0.8mm} \bar{\sxb}^{t-1}_{\widebar{N}}\|^2 \nnb
		&\quad + \frac{6a_5b\mu^2\delta^2(i+1)}{\widebar{N}} \sum_{j=0}^{\widebar{N}-1}\bE\|\check{\sxb}^{t-1}_{j}\|^2  \nnb
		&= \Big(\lambda_1^{i+1} +18a_5b\mu^2\delta^2(i+1) \Big) \bE\|\check{\sxb}_0^t\|^2 \nnb
		&\quad + (2a_4 + 6 a_5 b) \mu^2 \delta^2 \sum_{j=0}^{i} \bE\|\bar{\sxb}_j^t \suo - \suo \bar{\sxb}_0^t\|^2   \nnb
		&\quad  +  2 a_4 \mu^2 \delta^2 (i+1) \bE\|\bar{\sxb}^t_{0}\|^2 \nnb
		&\quad + \frac{3a_5b\mu^2\delta^2(i+1)}{\widebar{N}} \sum_{j=0}^{\widebar{N}-1}\bE\|\bar{\sxb}^{t-1}_{j} \hspace{-0.8mm}-\hspace{-0.8mm} \bar{\sxb}^{t-1}_{\widebar{N}}\|^2 \nnb
		&\quad + \frac{6a_5b\mu^2\delta^2(i+1)}{\widebar{N}}\sum_{j=0}^{\widebar{N}-1}\bE\|\check{\sxb}^{t-1}_{j}\|^2,
	}
	where (a) holds because $\lambda_1 < 1$ and {\color{black}hence $\sum_{j=0}^{i}\lambda_1^{i-j} \le i+1$.} Next we let $\lambda_2 = (1+\lambda_1)/2 < 1$. If the step-size $\mu$ is chosen small enough such that 
	\eq{\label{mu-1}
		\lambda_1^{i+1} +2a_4\mu^2\delta^2 (i+1) \le \lambda_2,\quad \forall\ i = 0, \cdots, \widebar{N}-1
	}
	then it follows that
	\eq{\label{b139dj-3}
		&\bE\|\check{\sxb}_{i+1}^t\|^2 \nnb
		&\le \lambda_2 \bE\|\check{\sxb}_0^t\|^2 + (2a_4 + 6 a_5 b) \mu^2 \delta^2 \sum_{j=0}^{i} \bE\|\bar{\sxb}_j^t \suo - \suo \bar{\sxb}_0^t\|^2   \nnb
		&\quad  +  2 a_4 \mu^2 \delta^2 (i+1) \bE\|\bar{\sxb}^t_{0}\|^2 \nnb
		&\quad + \frac{3a_5b\mu^2\delta^2(i+1)}{\widebar{N}} \sum_{j=0}^{\widebar{N}-1}\bE\|\bar{\sxb}^{t-1}_{j} \hspace{-0.8mm}-\hspace{-0.8mm} \bar{\sxb}^{t-1}_{\widebar{N}}\|^2 \nnb
		&\quad + \frac{6a_5b\mu^2\delta^2(i+1)}{\widebar{N}}\sum_{j=0}^{\widebar{N}-1}\bE\|\check{\sxb}^{t-1}_{j}\|^2 \nnb
		&\le \lambda_2 \bE\|\check{\sxb}_0^t\|^2 \suo + \suo (2a_4\suo + \suo 6 a_5 b) \mu^2 \delta^2 \sum_{j=0}^{\widebar{N}-1} \bE\|\bar{\sxb}_j^t \suo - \suo \bar{\sxb}_0^t\|^2 \nnb
		& \quad + \suo  2 a_4 \mu^2 \delta^2 \widebar{N} \bE\|\bar{\sxb}^t_{0}\|^2 \nnb
		& \quad + 3a_5b\mu^2\delta^2\widebar{N}\left(\frac{1}{\widebar{N}}\sum_{j=0}^{\widebar{N}-1}\bE\|\bar{\sxb}^{t-1}_{j} \hspace{-0.8mm}-\hspace{-0.8mm} \bar{\sxb}^{t-1}_{\widebar{N}}\|^2 \right) \nnb
		&\quad + 6a_5b\mu^2\delta^2\widebar{N}\left(\frac{1}{\widebar{N}}\sum_{j=0}^{\widebar{N}-1}\bE\|\check{\sxb}^{t-1}_{j}\|^2\right)\hspace{-0.6mm},\quad \forall\ i = 0, \cdots, \widebar{N}-1 
	}
	Notice that 
	\eq{
	\lambda_1^{i+1} +2a_4\mu^2\delta^2 (i+1) \le \lambda_1 + 2a_4\mu^2\delta^2 \widebar{N}, \ \forall i = 0, \cdots, \widebar{N}-1.
	}
	Therefore, to guarantee \eqref{mu-1}, it is enough to set
	\eq{\label{zbwn999-1}
		\lambda_1 +2a_4\mu^2\delta^2 \widebar{N} \le \lambda_2 \Longleftrightarrow	\mu \le \sqrt{\frac{\lambda_2 \suo - \suo \lambda_1}{2a_4\delta^2 \widebar{N}}}.
	}
	From \eqref{b139dj-3} we can derive
	\eq{
		&\hspace{-5mm} \sum_{i=1}^{\widebar{N}-1}\bE\|\check{\sxb}_{i}^t\|^2 \nnb &\le\lambda_2(\widebar{N}-1) \bE\|\check{\sxb}_0^t\|^2 \nnb
		&\quad + \suo (2a_4\suo + \suo 6 a_5 b) \mu^2 \delta^2 (\widebar{N}-1) \sum_{j=0}^{\widebar{N}-1} \bE\|\bar{\sxb}_j^t \suo - \suo \bar{\sxb}_0^t\|^2  \nnb
		&\quad + \suo  2 a_4 \mu^2 \delta^2 \widebar{N}  (\widebar{N} \hspace{-0.6mm}-\hspace{-0.6mm}1) \bE\|\bar{\sxb}^t_{0}\|^2  \nnb
		&\quad + 3a_5b\mu^2\delta^2\widebar{N}(\widebar{N}-1)\left(\frac{1}{\widebar{N}}\sum_{j=0}^{\widebar{N}\hspace{-0.6mm}-\hspace{-0.6mm}1}\bE\|\bar{\sxb}^{t-1}_{j} \hspace{-0.8mm}-\hspace{-0.8mm} \bar{\sxb}^{t-1}_{\widebar{N}}\|^2 \right)\nnb
		&\quad + \suo 6a_5b\mu^2\delta^2\widebar{N}(\widebar{N}-1)\left(\frac{1}{\widebar{N}}\sum_{j=0}^{\widebar{N}-1}\bE\|\check{\sxb}^{t-1}_{j}\|^2\right).
	}
	As a result,
	\eq{
		&\hspace{-5mm}\frac{1}{\widebar{N}}\sum_{i=0}^{\widebar{N}-1}\bE\|\check{\sxb}_{i}^t\|^2 \nnb
		&= \frac{1}{\widebar{N}}\left(\sum_{i=1}^{\widebar{N}-1}\bE\|\check{\sxb}_{i}^t\|^2\suo + \suo \bE\|\check{\sxb}_{0}^t\|^2 \right)\nnb
		&\le \frac{\lambda_2(\widebar{N}-1)\suo + \suo 1}{\widebar{N}}\bE\|\check{\sxb}_0^t\|^2 \nnb
		&\quad + \suo (2a_4\suo + \suo 6 a_5 b) \mu^2 \delta^2\widebar{N} \left(\frac{1}{\widebar{N}} \sum_{j=0}^{\widebar{N}-1} \bE\|\bar{\sxb}_j^t \suo - \suo \bar{\sxb}_0^t\|^2 \right) \nnb
		&\quad + 2 a_4 \mu^2 \delta^2 \widebar{N} \bE\|\bar{\sxb}^t_{0}\|^2 \nnb
		&\quad + \suo 3a_5b\mu^2\delta^2\widebar{N} \left(\frac{1}{\widebar{N}}\sum_{j=0}^{\widebar{N}-1}\bE\|\bar{\sxb}^{t-1}_{j} \hspace{-0.8mm}-\hspace{-0.8mm} \bar{\sxb}^{t-1}_{\widebar{N}}\|^2 \right)\nnb
		&\quad + 6a_5b\mu^2\delta^2\widebar{N} \left(\frac{1}{\widebar{N}}\sum_{j=0}^{\widebar{N}-1}\bE\|\check{\sxb}^{t-1}_{j}\|^2\right).
	}
	To simplify the notation, we let
	\eq{\label{c1-c4}
		&\lambda_3 = \frac{\lambda_2(\widebar{N}-1)\suo + \suo 1}{\widebar{N}}, \nnb
		&c_1 = 2a_4,\ c_2 = 2a_4\suo + \suo 6 a_5 b ,\ c_3 = 3a_5b,\ c_4 = 6a_5 b.
	}
	{\color{black}Using $\lambda_2<1$, we have
		\eq{
			\lambda_3 = \frac{\lambda_2(\widebar{N}-1)\suo + \suo 1}{\widebar{N}} < \frac{\widebar{N}-1\suo + \suo 1}{\widebar{N}} = 1.
		}	
	}
	\hspace{-1mm}In summary, when $\mu$ satisfies \eqref{trivial-stepsize}, \eqref{zbwn999} and \eqref{zbwn999-1}, i.e. 
	\eq{\label{zn27}
		\mu \le \min \left\{ {\color{black}\frac{1}{\delta}}, 
		\sqrt{\frac{1-\lambda}{2(a_3+12a_5b)\delta^2}}, \sqrt{\frac{\lambda_2 \suo - \suo \lambda_1}{2a_4\delta^2 \widebar{N}}}
		\right\},
	}
	we conclude recursion \eqref{sum-check-x}. To get a simple form for the step-size, with $\lambda_2 - \lambda_1 = (1-\lambda)/4$ we can further restrict $\mu$ as
	\eq{\label{C1}
		\mu &\le \min\left\{{\color{black}1}, 
		\sqrt{\frac{1}{2(a_3+12a_5b)}}, \sqrt{\frac{1}{8a_4}} \right\} \sqrt{\frac{1-\lambda}{\delta^2 \widebar{N}}} \nnb &\define C_1 \sqrt{\frac{1-\lambda}{\delta^2 \widebar{N}}}.
	}
	It is obvious that all step-sizes within the range defined in \eqref{C1} will also satisfy \eqref{zn27}. Moreover, recursion \eqref{check-x} holds by setting $i=\widebar{N}-1$ in \eqref{b139dj-3}.
	
	\section{Proof of Lemma \ref{lm-bar-x}}\label{app-bar-x}
	\noindent Substituting \eqref{zx23b9} into the first line of \eqref{xngyi}, we have
	\eq{\label{23am99-2}
		&\bE\|\bar{\sxb}^t_{i+1}\|^2 \nnb
		&\le (1- a_1 \mu \nu) \bE\|\bar{\sxb}^t_{i}\|^2\suo + \suo \frac{2a_2 \mu \delta^2 }{ \nu} \bE\|\check{\sxb}^t_{i}\|^2\suo + \suo \frac{2\mu}{\nu} \bE\|\s(\swb^t_{i})\|^2 \nnb
		&\overset{\eqref{zx23b9}}{\le} (1- a_1\mu \nu) \bE\|\bar{\sxb}^t_{i}\|^2\suo + \suo \frac{2a_2 \mu \delta^2 }{ \nu} \bE\|\check{\sxb}^t_{i}\|^2 \nnb
		&\quad + \suo \frac{12 b \delta^2\mu}{\nu} \bE\|\bar{\sxb}^t_{i} \suo - \suo \bar{\sxb}^t_{0} \|^2 + \frac{24b\delta^2 \mu}{\nu} \bE\|\check{\sxb}^t_{i}\|^2 \nnb
		&\quad + \suo \frac{36 b \delta^2 \mu}{\nu} \bE\|\check{\sxb}^t_{0}\|^2 + \frac{6b\delta^2 \mu}{\nu \widebar{N}}\sum_{j=0}^{\widebar{N}-1} \bE\|\bar{\sxb}^{t-1}_{j} \suo - \suo \bar{\sxb}^{t-1}_{\widebar{N}} \|^2 \nnb
		&\quad + \suo \frac{12 b \delta^2 \mu}{\widebar{N} \nu}\sum_{j=0}^{\widebar{N}-1}\bE\|\check{\sxb}^{t-1}_{j}\|^2 \nnb
		&= (1- a_1\mu \nu) \bE\|\bar{\sxb}^t_{i}\|^2\suo + \suo \frac{(2a_2\suo + \suo 24b) \mu \delta^2 }{ \nu} \bE\|\check{\sxb}^t_{i}\|^2 \nnb
		&\quad + \suo \frac{12 b \delta^2\mu}{\nu} \bE\|\bar{\sxb}^t_{i} \suo - \suo \bar{\sxb}^t_{0} \|^2 + \frac{36 b \delta^2 \mu}{\nu} \bE\|\check{\sxb}^t_{0}\|^2 \nnb
		&\quad + \suo \frac{6b\delta^2 \mu}{\nu \widebar{N}}\sum_{j=0}^{\widebar{N}-1} \bE\|\bar{\sxb}^{t-1}_{j} \suo - \suo \bar{\sxb}^{t-1}_{\widebar{N}} \|^2\suo + \suo \frac{12 b \delta^2 \mu}{\widebar{N} \nu}\sum_{j=0}^{\widebar{N}-1}\bE\|\check{\sxb}^{t-1}_{j}\|^2
	}
	Iterate \eqref{23am99-2}, then for $0\le i\le \widebar{N}-1$ it holds that
	\eq{\label{zxcn23}
		&\bE\|\bar{\sxb}^t_{i+1}\|^2 \nnb
		&\le (1- a_1 \mu \nu)^{i+1}\bE\|\bar{\sxb}^t_{0}\|^2 \nnb
		&\quad + \suo \frac{(2a_2\suo + \suo 24b) \mu \delta^2 }{ \nu} \sum_{j=0}^{i} (1-a_1\mu\nu)^{i-j} \bE\|\check{\sxb}^t_{j}\|^2 \nnb
		&\quad + \frac{12 b \delta^2\mu}{\nu}\sum_{j=0}^{i}(1- a_1\mu \nu)^{i-j}\bE\|\bar{\sxb}^t_{j} \suo - \suo \bar{\sxb}^t_{0} \|^2\nnb
		&\quad + \suo \left(\frac{36 b \delta^2 \mu}{\nu} \bE\|\check{\sxb}^t_{0}\|^2 + \frac{6b\delta^2 \mu}{\nu \widebar{N}}\sum_{j=0}^{\widebar{N}-1} \bE\|\bar{\sxb}^{t-1}_{j} \suo - \suo \bar{\sxb}^{t-1}_{\widebar{N}} \|^2 \right.\nnb
		&\qquad\quad \left. + \frac{12 b \delta^2 \mu}{\widebar{N} \nu}\sum_{j=0}^{\widebar{N}-1}\bE\|\check{\sxb}^{t-1}_{j}\|^2\right)\sum_{j=0}^{i}(1-a_1\mu\nu)^j \nnb
		&\le (1- a_1 \mu \nu)^{i+1}\bE\|\bar{\sxb}^t_{0}\|^2 + \frac{(2a_2 + 24b) \mu \delta^2 }{ \nu} \sum_{j=0}^{i}  \bE\|\check{\sxb}^t_{j}\|^2 \nnb
		&\quad + \frac{12 b \delta^2\mu}{\nu}\sum_{j=0}^{i}\bE\|\bar{\sxb}^t_{j} \suo - \suo \bar{\sxb}^t_{0} \|^2 \suo + \suo \left( \frac{36 b \delta^2 \mu}{\nu} \bE\|\check{\sxb}^t_{0}\|^2 \right. \nnb
		&\qquad + \frac{6b\delta^2 \mu}{\nu \widebar{N}}\sum_{j=0}^{\widebar{N}-1} \bE\|\bar{\sxb}^{t-1}_{j} \suo - \suo \bar{\sxb}^{t-1}_{\widebar{N}} \|^2 \nnb
		&\qquad \left. + \frac{12 b \delta^2 \mu}{\widebar{N} \nu}\sum_{j=0}^{\widebar{N}-1}\bE\|\check{\sxb}^{t-1}_{j}\|^2 \right) (i + 1),
	}
	{\color{black}where the last inequality hold when we choose $\mu$ small enough such that 
		\eq{\label{23bsdns99}
			0 < 1-a_1\mu\nu < 1 \Longleftrightarrow \mu < \frac{1}{a_1\nu}.
		}
	}
	Let $i=\widebar{N} - 1$ in \eqref{zxcn23}. It holds that
	\eq{\label{l2nsd}
		&\bE\|\bar{\sxb}^{t+1}_0\|^2 \nnb
		&\le (1- a_1 \mu \nu)^{\widebar{N}}\bE\|\bar{\sxb}^t_{0}\|^2 + \frac{(2a_2 + 24b) \mu \delta^2 }{ \nu} \sum_{j=0}^{\widebar{N} - 1}  \bE\|\check{\sxb}^t_{j}\|^2\nnb &\quad +\frac{12 b \delta^2\mu}{\nu}\sum_{j=0}^{\widebar{N} - 1}\bE\|\bar{\sxb}^t_{j} \suo - \suo \bar{\sxb}^t_{0} \|^2 + \left(\frac{36 b \delta^2 \widebar{N} \mu }{\nu} \bE\|\check{\sxb}^t_{0}\|^2 \right. \nnb
		&\quad \left. \suo + \suo \frac{6b\delta^2 \mu}{\nu }\suo\sum_{j=0}^{\widebar{N}-1}\suo \bE\|\bar{\sxb}^{t-1}_{j} \suo - \suo \bar{\sxb}^{t-1}_{\widebar{N}} \|^2 + \frac{12 b \delta^2 \mu}{ \nu}\suo\sum_{j=0}^{\widebar{N}-1}\suo\bE\|\check{\sxb}^{t-1}_{j}\|^2\right) \nnb
		&=\suo(1 \suo - \suo a_1 \mu \nu)^{\widebar{N}}\bE\|\bar{\sxb}^t_{0}\|^2 \suo + \suo \frac{(2a_2 + 24b) \mu \delta^2 \widebar{N} }{\nu}\ssuo\left( \suo\frac{1}{\widebar{N}}\suo\sum_{j=0}^{\widebar{N}-1}\suo\bE\|\check{\sxb}^t_{j}\|^2 \ssuo \right) \nnb
		&\quad + \frac{12b\delta^2\mu \widebar{N}}{\nu}\ssuo\left( \frac{1}{\widebar{N}}\sum_{j=0}^{\widebar{N}-1}\bE\|\bar{\sxb}^t_{j} \suo - \suo \bar{\sxb}^t_{0} \|^2 \right)\ssuo+\suo \frac{36 b \delta^2 \widebar{N} \mu}{\nu} \bE\|\check{\sxb}^t_{0}\|^2 \nnb
		&\quad + \frac{6b\delta^2 \mu \widebar{N}}{\nu}\left(\frac{1}{\widebar{N}}\sum_{j=0}^{\widebar{N}-1} \bE\|\bar{\sxb}^{t-1}_{j} \suo - \suo \bar{\sxb}^{t-1}_{\widebar{N}} \|^2 \right) \nnb 
		&\quad + \frac{12b\delta^2 \mu \widebar{N}}{\nu}\left( \frac{1}{\widebar{N}}\sum_{j=0}^{\widebar{N}-1}\bE\|\check{\sxb}^{t-1}_{j}\|^2\right).
	}
	{\color{black}According to Lemma \ref{lm-check-x}, the inequality \eqref{sum-check-x} holds when step-size $\mu$ satisfies
		\eq{\label{2bsnsnsn}
			\mu \le C_1 \sqrt{\frac{1-\lambda}{\delta^2 \overline{N}}}.
		}	
	}
	Substituting \eqref{sum-check-x} into \eqref{l2nsd}, we get
	\eq{\label{l2nsd-2}
		&\bE\|\bar{\sxb}^{t+1}_0\|^2 \nnb
		&\le \left((1- a_1 \mu \nu)^{\widebar{N}} + \frac{c_1(2a_2 + 24b) \mu^3 \delta^4 \widebar{N}^2 }{ \nu}\right)\bE\|\bar{\sxb}^t_{0}\|^2 \nnb
		&\quad + \left( \frac{36 b \delta^2 \widebar{N} \mu}{\nu} +\frac{\lambda_3(2a_2 + 24b) \mu \delta^2 \widebar{N} }{ \nu} \right) \bE\|\check{\sxb}^t_{0}\|^2 \nnb
		&\quad + \left(\frac{12b\delta^2\mu \widebar{N}}{\nu} +  \frac{c_2(2a_2 + 24b) \mu^3 \delta^4 \widebar{N}^2 }{ \nu} \right) \nnb 
		&\qquad  \cdot \left( \frac{1}{\widebar{N}}\sum_{j=0}^{\widebar{N}-1}\bE\|\bar{\sxb}^t_{j} \suo - \suo \bar{\sxb}^t_{0} \|^2 \right)  \nnb
		&\quad  + \left(\frac{6b\delta^2\mu \widebar{N}}{\nu} +  \frac{c_3(2a_2 + 24b) \mu^3 \delta^4 \widebar{N}^2 }{ \nu} \right)\nnb 
		&\qquad \cdot \left(\frac{1}{\widebar{N}}\sum_{j=0}^{\widebar{N}-1} \bE\|\bar{\sxb}^{t-1}_{j} \suo - \suo \bar{\sxb}^{t-1}_{\widebar{N}} \|^2 \right) \nnb
		&\quad + \left(\frac{12b\delta^2\mu \widebar{N}}{\nu} +  \frac{c_4(2a_2 + 24b) \mu^3 \delta^4 \widebar{N}^2 }{ \nu} \right)\nnb 
		&\qquad \cdot \left( \frac{1}{\widebar{N}}\sum_{j=0}^{\widebar{N}-1}\bE\|\check{\sxb}^{t-1}_{j}\|^2\right).
	}
	For the term $(1-a_1\mu \nu)^{\widebar{N}}$, it is established in Appendix \ref{app-up-bound} that if 
	\eq{\label{2asdhsdnd9}
		\mu \le \frac{1}{a_1 \widebar{N} \nu},
	}
	then the inequality $(1-a_1\mu \nu)^{\widebar{N}} \le 1 -{ a_1 \widebar{N}  \mu \nu}/{2}$ holds.  Furthermore, if the step-size $\mu$ is chosen small enough such that
	\eq{
		1- \frac{a_1 \widebar{N} \mu \nu}{2} + \frac{c_1(2a_2 + 24b) \mu^3 \delta^4 \widebar{N}^2 }{ \nu} &\le 1 \suo - \suo \frac{a_1 \widebar{N}\mu \nu}{3} \nnb
		\frac{12b\delta^2\mu \widebar{N}}{\nu} +  \frac{c_2(2a_2 + 24b) \mu^3 \delta^4 \widebar{N}^2 }{ \nu} &\le \frac{24 b \delta^2 \widebar{N} \mu}{\nu} \nnb
		\frac{6b\delta^2\mu \widebar{N}}{\nu} +  \frac{c_3(2a_2 + 24b) \mu^3 \delta^4 \widebar{N}^2 }{ \nu} &\le \frac{12b\delta^2\mu \widebar{N}}{\nu} \nnb
		\frac{12b\delta^2\mu \widebar{N}}{\nu} +  \frac{c_4(2a_2 + 24b) \mu^3 \delta^4 \widebar{N}^2 }{ \nu} &\le
		\frac{24b\delta^2\mu \widebar{N}}{\nu} \label{z2sdh}
	}
	recursion \eqref{l2nsd-2} will imply
	\eq{\label{l2nsd-3}
		&\bE\|\bar{\sxb}^{t+1}_0\|^2 \nnb
		&\le \left(1- \frac{\widebar{N}}{3}a_1 \mu \nu\right)\bE\|\bar{\sxb}^t_{0}\|^2 \nnb 
		&\quad + \left( \frac{(36b + 2\lambda_3a_2 + 24\lambda_3b) \mu \delta^2 \widebar{N} }{ \nu} \right) \bE\|\check{\sxb}^t_{0}\|^2 \nnb
		&\quad + \frac{24b\delta^2\mu \widebar{N}}{\nu}\left( \frac{1}{\widebar{N}}\sum_{j=0}^{\widebar{N}-1}\bE\|\bar{\sxb}^t_{j} \suo - \suo \bar{\sxb}^t_{0} \|^2 \right) \nnb
		&\quad + \frac{12b\delta^2\mu \widebar{N}}{\nu} \left(\frac{1}{\widebar{N}}\sum_{j=0}^{\widebar{N}-1} \bE\|\bar{\sxb}^{t-1}_{j} \suo - \suo \bar{\sxb}^{t-1}_{\widebar{N}} \|^2 \right) \nnb
		&\quad + \frac{24b\delta^2\mu \widebar{N}}{\nu} \left( \frac{1}{\widebar{N}}\sum_{j=0}^{\widebar{N}-1}\bE\|\check{\sxb}^{t-1}_{j}\|^2\right).
	}
	To simplify the notation, we let
	\eq{\label{d-notation}
		d_1 = 36b + 2\lambda_3a_2 + 24\lambda_3b, d_2 =24b, d_3 = 12b, d_4 = 24b,
	}
	then recursion \eqref{l2nsdsdf} is proved. To guarantee {\color{black}\eqref{23bsdns99}, \eqref{2bsnsnsn},} \eqref{2asdhsdnd9} and \eqref{z2sdh}, it is enough to set
	\eq{\label{zb2ns9}
		\mu &\le \min\left\{{\color{black}\frac{1}{a_1 \nu},\  C_1\sqrt{\frac{1\suo - \suo \lambda}{\delta^2 \widebar{N}}}},\ \frac{1}{a_1 \widebar{N} \nu}, \right. \nnb
		& \left. \sqrt{\frac{a_1}{6c_1(2a_2 + 24b)\widebar{N}}}\left(\frac{\nu}{\delta^2}\right),   \sqrt{\frac{12b}{c_2(2a_2+24b)\delta^2 \widebar{N}}}, \right.\nnb
		& \left. \sqrt{\frac{6b}{c_3(2a_2+24b)\delta^2 \widebar{N}}}, \sqrt{\frac{12b}{c_4(2a_2+24b)\delta^2 \widebar{N}}}\ \right\}
	}
	Note that $\nu^2/\delta^2<1$ and $1-\lambda < 1$. To get a simple form for the step-size, we can further restrict $\mu$ as
	\eq{\label{C2}
		\mu &\le \min\left\{{\color{black}C_1}, \frac{1}{a_1}, \sqrt{\frac{a_1}{2c_1(2a_2 + 24b)}}, \sqrt{\frac{12b}{c_2(2a_2+24b)}}, \right. \nnb
		&      \left. \sqrt{\frac{6b}{c_3(2a_2+24b)}}, \sqrt{\frac{12b}{c_4(2a_2+24b)}}\right\}\left(\frac{\nu \sqrt{1-\lambda}}{\delta^2 \widebar{N}}\right) \nnb 
		&\define C_2 \left(\frac{\nu\sqrt{1-\lambda}}{\delta^2 \widebar{N}}\right),
	}
	where $C_2$ is independent of $\nu$, $\delta$ and $\widebar{N}$.
	
	\section{Upper Bound on $(1- a_1 \mu \nu)^{\widebar{N}}$} \label{app-up-bound}
	We first examine the term $(1-x)^{\widebar{N}}$ where $x\in (0,1)$. Using Taylor's theorem, $(1-x)^{\widebar{N}}$ can be expanded as
	%
	%
	\eq{
		(1-x)^{\widebar{N}} = 1 \suo - \suo \widebar{N} x + \frac{\widebar{N}(\widebar{N}-1)(1-\tau)^{\widebar{N}-2}}{2}x^2,
	}
	where $\tau \in (0, x)$ is some constant, and hence, $\tau < 1$. To ensure $(1-x)^{\widebar{N}} \le 1 - \frac{1}{2}\widebar{N}x$, we require 
	\eq{\label{23sdn2ns8}
		& 1 \suo - \suo \widebar{N} x + \frac{\widebar{N}(\widebar{N}-1)(1-\tau)^{\widebar{N}-2}}{2}x^2 \le 1 \suo - \suo \frac{\widebar{N} x}{2} \nnb  \Longleftrightarrow\ 
		& x \le \frac{1}{(\widebar{N}-1)(1-\tau)^{\widebar{N}-2}}.
	}
	Note that
	\eq{
		\frac{1}{\widebar{N}} < 	\frac{1}{\widebar{N}-1} <  \frac{1}{(\widebar{N}-1)(1-\tau)^{\widebar{N}-2}}.
	}
	If we choose $x \le {1}/{\widebar{N}}$, then it will also satisfy \eqref{23sdn2ns8}. By letting $x = a_1 \mu \nu$, it holds that
	\eq{
		(1- a_1 \mu \nu)^{\widebar{N}} \le 1 -\frac{a_1 \widebar{N} \mu \nu}{2}.
	} 
	when $\mu \le 1/(a_1 \widebar{N} \nu)$.

	\section{Proof of Lemma \ref{lm-inner-recursion}}\label{app-inner-recursion}
	\noindent From the first line in recursion \eqref{23nsad-2}, we have
	\eq{
		\bar{\sxb}_{i+1}^t \ssuo - \ssuo \bar{\sxb}_{i}^t \ssuo = \ssuo -\frac{\mu}{K}\cI\tran \cHb_i^t \cI \bar{\sxb}_i^t \suo - \suo \frac{\mu}{K}\cI\tran \cHb_i^t \cX_{R,u} \check{\sxb}_i^t \ssuo + \ssuo \frac{\mu}{K}\cI\tran \s(\swb_i^t)
	}
	By squaring and applying Jensen's inequality, we have
	\eq{
		&\|\bar{\sxb}_{i+1}^t \suo - \suo \bar{\sxb}_{i}^t\|^2 \nnb
		&\le\ 3\mu^2\left\|\frac{1}{K}\cI\tran \cHb_i^t \cI\right\|^2\|\bar{\sxb}_i^t\|^2 \nnb 
		&\quad + \frac{3\mu^2}{K^2}\|\cI\tran \cHb_i^t \cX_{R,u}\|^2 \|\check{\sxb}_i^t\|^2  \nnb 
		&\quad + \frac{3\mu^2}{K^2}\|\cI\tran\|^2\|\s(\swb_i^t)\|^2 \nnb
		&\overset{(a)}{\le}\ 3\mu^2 \delta^2 \|\bar{\sxb}_i^t\|^2 + \frac{3\mu^2}{K}\delta^2 \|\cX_{R,u}\|^2 \|\check{\sxb}_i^t\|^2 + \frac{3\mu^2}{K}\|\s(\swb_i^t)\|^2
	}
	where inequality (a) holds because of equations \eqref{23nsd9} and \eqref{zxcn239}. By taking expectations, we have
	\eq{\label{asdh}
		&\bE\|\bar{\sxb}_{i+1}^t \suo - \suo \bar{\sxb}_{i}^t\|^2 \nnb
		&\le 3\mu^2 \delta^2 \bE\|\bar{\sxb}_i^t\|^2 + \frac{3\mu^2}{K}\delta^2 \|\cX_{R,u}\|^2 \bE\|\check{\sxb}_i^t\|^2 + \frac{3\mu^2}{K}\bE\|\s(\swb_i^t)\|^2 \nnb
		&\le 6\mu^2\delta^2 \bE\|\bar{\sxb}_0^t\|^2 + 6\mu^2\delta^2 \bE\|\bar{\sxb}_i^t \suo - \suo \bar{\sxb}_0^t\|^2 \nnb 
		&\quad + \frac{3\mu^2}{K}\delta^2 \|\cX_{R,u}\|^2 \bE\|\check{\sxb}_i^t\|^2 + \frac{3\mu^2}{K}\bE\|\s(\swb_i^t)\|^2 \nnb
		&\overset{\eqref{zx23b9}}{\le}\ssuo 6\mu^2\delta^2 \bE\|\bar{\sxb}_0^t\|^2 + \frac{54b\mu^2\delta^2}{K}\bE\|\check{\sxb}_0^t\| \nnb 
		&\quad + \left(\frac{3\|\cX_{R,u}\|^2 + 36b}{K}\right)\mu^2\delta^2\bE\|\check{\sxb}_i^t\|^2 \nnb
		&\quad + \left(6 + \frac{18b}{K}\right)\mu^2\delta^2\bE\|\bar{\sxb}_i^t \suo - \suo \bar{\sxb}_0^t\|^2 \nnb
		&\quad + \frac{9b\delta^2 \mu^2}{K}\left(\frac{1}{\widebar{N}}\sum_{j=0}^{\widebar{N}-1}\bE\|\bar{\sxb}^{t-1}_{j} \hspace{-0.8mm}-\hspace{-0.8mm} \bar{\sxb}^{t-1}_{\widebar{N}}\|^2\right) \nnb
		&\quad + \frac{18b\delta^2 \mu^2}{K}\left(\frac{1}{\widebar{N}}\sum_{j=0}^{\widebar{N}-1}\bE\|\check{\sxb}^{t-1}_{j}\|^2\right),  0\le i\le \widebar{N}-1
	}
	For simplicity, if we let
	\eq{\label{e-notation}
		&e_1 = \frac{54b}{K}, \ \ e_2 = \frac{3\|\cX_{R,u}\|^2 + 36b}{K}, \nnb 
		&e_3 = 6 + \frac{18b}{K},\ \ e_4 = \frac{9b}{K},\ \ e_5 = \frac{18b}{K},
	}
	inequality \eqref{asdh} becomes
	\eq{\label{nsdl9}
		&\bE\|\bar{\sxb}_{i+1}^t \suo - \suo \bar{\sxb}_{i}^t\|^2 \nnb
		&\le 6\mu^2\delta^2 \bE\|\bar{\sxb}_0^t\|^2 + e_1 \mu^2 \delta^2\bE\|\check{\sxb}_0^t\|^2 + e_2\mu^2\delta^2\bE\|\check{\sxb}_i^t\|^2 \nnb 
		&\quad + e_3\mu^2\delta^2\bE\|\bar{\sxb}_i^t \suo - \suo \bar{\sxb}_0^t\|^2 \nnb 
		&\quad + e_4\mu^2\delta^2 \left(\frac{1}{\widebar{N}}\sum_{j=0}^{\widebar{N}-1}\bE\|\bar{\sxb}^{t-1}_{j} \hspace{-0.8mm}-\hspace{-0.8mm} \bar{\sxb}^{t-1}_{\widebar{N}}\|^2\right) \nnb 
		&\quad + e_5\mu^2\delta^2\left(\frac{1}{\widebar{N}}\sum_{j=0}^{\widebar{N}-1}\bE\|\check{\sxb}^{t-1}_{j}\|^2\right).
	}
	For $1\le i\le \widebar{N}-1$, we have
	\eq{\label{zxnnshsusu}
		&\bE\|\bar{\sxb}_{i}^t \suo - \suo \bar{\sxb}_{0}^t\|^2 \nnb
		&\le i \sum_{j=1}^{i} \bE\|\bar{\sxb}_{j}^t \suo - \suo \bar{\sxb}_{j-1}^t\|^2 \nnb
		&\overset{\eqref{nsdl9}}{\le} 6\mu^2\delta^2 i^2 \bE\|\bar{\sxb}_0^t\|^2 + e_1 \mu^2 \delta^2 i^2 \bE\|\check{\sxb}_0^t\|^2 \nnb
		&\quad + \suo e_2\mu^2\delta^2 i \suo \sum_{j=1}^{i}\suo \bE\|\check{\sxb}_{j-1}^t\|^2 \suo + \suo e_3\mu^2\delta^2i\suo \sum_{j=1}^{i}\suo\bE\|\bar{\sxb}_{j-1}^t \suo - \suo \bar{\sxb}_0^t\|^2 \nnb
		&\quad + e_4\mu^2\delta^2 i^2 \left(\frac{1}{\widebar{N}}\sum_{j=0}^{\widebar{N}-1}\bE\|\bar{\sxb}^{t-1}_{j} \hspace{-0.8mm}-\hspace{-0.8mm}  \bar{\sxb}^{t-1}_{\widebar{N}}\|^2\right) \nnb 
		&\quad + e_5\mu^2\delta^2i^2\left(\frac{1}{\widebar{N}}\sum_{j=0}^{\widebar{N}-1}\bE\|\check{\sxb}^{t-1}_{j}\|^2\right) \nnb
		&\le 6\mu^2\delta^2 \widebar{N}^2 \bE\|\bar{\sxb}_0^t\|^2 + e_1 \mu^2 \delta^2 \widebar{N}^2 \bE\|\check{\sxb}_0^t\|^2 \nnb
		&\quad + e_2\mu^2\delta^2 \widebar{N}^2\left( \frac{1}{\widebar{N}}\sum_{j=0}^{\widebar{N}-1}\bE\|\check{\sxb}_{j}^t\|^2\right)\nnb 
		&\quad + e_3\mu^2\delta^2 \widebar{N}^2 \left(\frac{1}{\widebar{N}}\sum_{j=0}^{\widebar{N}-1}\bE\|\bar{\sxb}_{j}^t \suo - \suo \bar{\sxb}_0^t\|^2 \right) \nnb
		& \quad + e_4\mu^2\delta^2 \widebar{N}^2 \left(\frac{1}{\widebar{N}}\sum_{j=0}^{\widebar{N}-1}\bE\|\bar{\sxb}^{t-1}_{j} \hspace{-0.8mm}-\hspace{-0.8mm} \bar{\sxb}^{t-1}_{\widebar{N}}\|^2\right) \nnb 
		& \quad + e_5\mu^2\delta^2\widebar{N}^2\left(\frac{1}{\widebar{N}}\sum_{j=0}^{\widebar{N}-1}\bE\|\check{\sxb}^{t-1}_{j}\|^2\right).
	}
	From the above recursion, we can also derive
	\eq{\label{23nsd0}
		&\frac{1}{\widebar{N}}\sum_{i=0}^{\widebar{N}-1}\bE\|\bar{\sxb}_{i}^t \suo - \suo \bar{\sxb}_{0}^t\|^2  \nnb
		&\le 6\mu^2\delta^2 \widebar{N}^2 \bE\|\bar{\sxb}_0^t\|^2 + e_1 \mu^2 \delta^2 \widebar{N}^2 \bE\|\check{\sxb}_0^t\|^2 \nnb
		&\quad + e_2\mu^2\delta^2 \widebar{N}^2\left( \frac{1}{\widebar{N}}\sum_{j=0}^{\widebar{N}-1}\bE\|\check{\sxb}_{j}^t\|^2\right) \nnb 
		&\quad + e_3\mu^2\delta^2 \widebar{N}^2 \left(\frac{1}{\widebar{N}}\sum_{j=0}^{\widebar{N}-1}\bE\|\bar{\sxb}_{j}^t \suo - \suo \bar{\sxb}_0^t\|^2 \right) \nnb
		&\quad + e_4\mu^2\delta^2 \widebar{N}^2 \left(\frac{1}{\widebar{N}}\sum_{j=0}^{\widebar{N}-1}\bE\|\bar{\sxb}^{t-1}_{j} \hspace{-0.8mm}-\hspace{-0.8mm} \bar{\sxb}^{t-1}_{\widebar{N}}\|^2\right) \nnb 
		&\quad + e_5\mu^2\delta^2\widebar{N}^2\left(\frac{1}{\widebar{N}}\sum_{j=0}^{\widebar{N}-1}\bE\|\check{\sxb}^{t-1}_{j}\|^2\right)
	}
	{\color{black}According to Lemma \ref{lm-check-x}, the inequality \eqref{sum-check-x} holds when step-size $\mu$ satisfies
		\eq{\label{2bsnsnsn-2}
			\mu \le C_1 \sqrt{\frac{1-\lambda}{\delta^2 \overline{N}}}.
		}	
	}
	\hspace{-1mm}Substituting \eqref{sum-check-x} into \eqref{23nsd0}, we have
	\eq{
		&\frac{1}{\widebar{N}}\sum_{i=0}^{\widebar{N}-1}\bE\|\bar{\sxb}_{i}^t \suo - \suo \bar{\sxb}_{0}^t\|^2  \nnb
		&\le \left(6\mu^2\delta^2 \widebar{N}^2 + c_1 e_2 \mu^4 \delta^4 \widebar{N}^3\right) \bE\|\bar{\sxb}_0^t\|^2 \nnb
		&\quad+ (e_1 + \lambda_3 e_2) \mu^2 \delta^2 \widebar{N}^2 \bE\|\check{\sxb}_0^t\|^2 \nnb
		&\quad + \suo \left(e_3\mu^2\delta^2\widebar{N}^2 \suo + \suo c_2e_2\mu^4\delta^4\widebar{N}^3\right) \ssuo \left(\frac{1}{\widebar{N}}\sum_{j=0}^{\widebar{N}-1}\bE\|\bar{\sxb}_{j}^t \suo - \suo \bar{\sxb}_0^t\|^2 \right) \nnb
		&\quad + \suo \left(e_4\mu^2\delta^2 \widebar{N}^2 \suo + \suo c_3e_2\mu^4\delta^4\widebar{N}^3\right) \ssuo \left(\frac{1}{\widebar{N}}\suo\sum_{j=0}^{\widebar{N}-1}\suo\bE\|\bar{\sxb}^{t-1}_{j} \hspace{-0.8mm}-\hspace{-0.8mm} \bar{\sxb}^{t-1}_{\widebar{N}}\|^2\right) \nnb
		&\quad + \suo \left(e_5\mu^2\delta^2 \widebar{N}^2 \suo + \suo c_4e_2\mu^4\delta^4\widebar{N}^3\right) \ssuo  \left(\frac{1}{\widebar{N}}\sum_{j=0}^{\widebar{N}-1}\bE\|\check{\sxb}^{t-1}_{j}\|^2\right).
	}
	If the step-size $\mu$ is chosen small enough such that
	\eq{\label{mu-2}
		&6\mu^2\delta^2 \widebar{N}^2 + c_1 e_2 \mu^4 \delta^4 \widebar{N}^3  \le 12\mu^2\delta^2 \widebar{N}^2,  \nnb
		&e_3\mu^2\delta^2\widebar{N}^2 + c_2e_2\mu^4\delta^4\widebar{N}^3  \le 2 e_3\mu^2\delta^2\widebar{N}^2, \nnb
		&e_4\mu^2\delta^2 \widebar{N}^2 + c_3e_2\mu^4\delta^4\widebar{N}^3  \le 2 e_4\mu^2\delta^2 \widebar{N}^2, \nnb
		&e_5\mu^2\delta^2 \widebar{N}^2 + c_4e_2\mu^4\delta^4\widebar{N}^3 \le 2e_5\mu^2\delta^2 \widebar{N}^2.
	}
	then recursion \eqref{23nsd0} can be simplified to equation \eqref{zxn238sd}, where we define $e_6 \define e_1 + \lambda_2 e_2$.
	To guarantee {\color{black}\eqref{2bsnsnsn-2} and} \eqref{mu-2}, it is enough to set
	\eq{\label{mu-3}
		\mu 
		&\le \min\left\{{\color{black}C_1}, \sqrt{\frac{6}{c_1e_2}}, \sqrt{\frac{e_3}{c_2e_2}}, \sqrt{\frac{e_4}{c_3e_2}}, \sqrt{\frac{e_5}{c_4e_2}}\right\}\sqrt{\frac{1-\lambda}{\delta^2 \widebar{N}}} \nnb
		&\define C_3 \sqrt{\frac{1-\lambda}{\delta^2 \widebar{N}}}.
	}
	Next we establish the recursion for $\sum_{i=0}^{\widebar{N}-1}\bE\|\bar{\sxb}^{t}_{i} \hspace{-0.8mm}-\hspace{-0.8mm} \bar{\sxb}^{t}_{\widebar{N}}\|^2/\overline
	N$. Note that for $0\le i \le \widebar{N}-1$, it holds that 
	\eq{\label{zxc,n238}
		&\bE\|\bar{\sxb}^{t}_{i} \hspace{-0.8mm}-\hspace{-0.8mm} \bar{\sxb}^{t}_{\widebar{N}}\|^2 \nnb
		&\le (\widebar{N} \suo - \suo i)\sum_{j={i}}^{\widebar{N}-1}\bE\|\bar{\sxb}^{t}_{j+1} \hspace{-0.8mm}-\hspace{-0.8mm} \bar{\sxb}^{t}_{j}\|^2 \nnb
		&\overset{\eqref{nsdl9}}{\le} 6\mu^2\delta^2 (\widebar{N}-i)^2 \bE\|\bar{\sxb}_0^t\|^2 + e_1 \mu^2 \delta^2 (N-i)^2 \bE\|\check{\sxb}_0^t\|^2 \nnb
		&\quad + e_2\mu^2\delta^2 (\widebar{N}-i) \sum_{j=i}^{\widebar{N}-1}\bE\|\check{\sxb}_{j}^t\|^2 \nnb
		&\quad + e_3\mu^2\delta^2 (N-i) \sum_{j=i}^{\widebar{N}-1}\bE\|\bar{\sxb}_{j-1}^t \suo - \suo \bar{\sxb}_0^t\|^2 \nnb
		&\quad + e_4\mu^2\delta^2 (\widebar{N} \hspace{-1mm} \suo - \suo \hspace{-1mm} i)^2\hspace{-1mm} \left(\hspace{-1mm}\frac{1}{\widebar{N}}\sum_{j=0}^{\widebar{N}-1}\bE\|\bar{\sxb}^{t-1}_{j} \hspace{-0.8mm}-\hspace{-0.8mm} \bar{\sxb}^{t-1}_{\widebar{N}}\|^2 \hspace{-1mm}\right) \nnb
		&\quad + e_5\mu^2\delta^2(\widebar{N}\hspace{-1mm}-\hspace{-1mm}i)^2\left(\hspace{-1mm}\frac{1}{\widebar{N}}\sum_{j=0}^{\widebar{N}-1}\bE\|\check{\sxb}^{t-1}_{j}\|^2\hspace{-1mm}\right) \nnb
		&\le 6\mu^2\delta^2 \widebar{N}^2 \bE\|\bar{\sxb}_0^t\|^2 + e_1 \mu^2 \delta^2 \widebar{N}^2 \bE\|\check{\sxb}_0^t\|^2 \nnb
		&\quad + e_2\mu^2\delta^2 \widebar{N}^2\left( \frac{1}{\widebar{N}}\sum_{j=0}^{\widebar{N}-1}\bE\|\check{\sxb}_{j}^t\|^2\right) \nnb
		&\quad + e_3\mu^2\delta^2 \widebar{N}^2 \left(\frac{1}{\widebar{N}}\sum_{j=0}^{\widebar{N}-1}\bE\|\bar{\sxb}_{j}^t\suo - \suo \bar{\sxb}_0^t\|^2 \right) \nnb
		&\quad + e_4\mu^2\delta^2 \widebar{N}^2 \left(\frac{1}{\widebar{N}}\sum_{j=0}^{\widebar{N}-1}\bE\|\bar{\sxb}^{t-1}_{j} \hspace{-0.8mm}-\hspace{-0.8mm} \bar{\sxb}^{t-1}_{\widebar{N}}\|^2\right)\nnb
		&\quad + e_5\mu^2\delta^2\widebar{N}^2\left(\frac{1}{\widebar{N}}\sum_{j=0}^{\widebar{N}-1}\bE\|\check{\sxb}^{t-1}_{j}\|^2\right).
	}
	Since the right-hand side of inequality \eqref{zxc,n238} is the same as inequality \eqref{zxnnshsusu}, we can follow \eqref{23nsd0}--\eqref{mu-3} to conclude recursion \eqref{zxlwseknsd}.
	
	\section{Proof of Theorem \ref{tho-balanced}}\label{app-tho-balanced}
	With Lemmas \ref{lm-check-x}, \ref{lm-bar-x} and \ref{lm-inner-recursion}, when the step-size $\mu$ satisfies
	\eq{\label{box-1}
		\boxed{
			\mu \le \min\left\{ C_1 \sqrt{\frac{1-\lambda}{\delta^2 \widebar{N}}},\ \ C_2\left(\frac{\nu \sqrt{1-\lambda}}{\delta^2 \widebar{N}}\right),\ \ C_3\sqrt{\frac{1-\lambda}{\delta^2 \widebar{N}}}  \right\}, }
	}
	it holds that
	\eq{
		\bE\|\bar{\sxb}^{t+1}_0\|^2
		&\le \left(1- \frac{\widebar{N}}{3}a_1 \mu \nu\right)\bE\|\bar{\sxb}^t_{0}\|^2  +  \frac{d_1 \mu \delta^2 \widebar{N} }{ \nu}  \bE\|\check{\sxb}^t_{0}\|^2  \nnb
		&   + \frac{d_2\delta^2\mu \widebar{N}}{\nu}\vA^t + \frac{d_3\delta^2\mu \widebar{N}}{\nu}\vB^{t-1}  + \frac{d_4\delta^2\mu \widebar{N}}{\nu} \vC^{t-1} \label{i1} \\ \nnb
		\bE\|\check{\sxb}_{0}^{t+1}\|^2 &\le c_1 \mu^2 \delta^2 \widebar{N} \bE\|\bar{\sxb}^t_{0}\|^2 + \lambda_2 \bE\|\check{\sxb}_0^t\|^2   \nnb
		&  + c_2 \mu^2 \delta^2\widebar{N} \vA^t  + c_3\mu^2\delta^2\widebar{N}\vB^{t-1}  + c_4\mu^2\delta^2\widebar{N}\vC^{t-1} \label{i2}\vspace{5mm}\\ \nnb
		\vA^{t+1}  &\le 12\mu^2\delta^2 \widebar{N}^2 \bE\|\bar{\sxb}_0^{t+1}\|^2 + e_6 \mu^2 \delta^2 \widebar{N}^2 \bE\|\check{\sxb}_0^{t+1}\|^2  \nnb
		& +\hspace{-0.8mm} 2e_3\mu^2\delta^2\widebar{N}^2\hspace{-0.5mm} \vA^{t\hspace{-0.2mm}+\hspace{-0.2mm}1} \hspace{-0.8mm}+\hspace{-0.8mm} 2e_4\mu^2\delta^2 \widebar{N}^2 \vB^t \hspace{-0.8mm}+\hspace{-0.8mm} 2e_5\mu^2\delta^2 \widebar{N}^2 \vC^t \label{i3}\vspace{5mm}\\ \nnb
		\vB^{t}  &\le 12\mu^2\delta^2 \widebar{N}^2 \bE\|\bar{\sxb}_0^{t}\|^2 + e_6 \mu^2 \delta^2 \widebar{N}^2 \bE\|\check{\sxb}_0^{t}\|^2  \nnb
		& +\hspace{-0.8mm} 2e_3\mu^2\hspace{-0.4mm}\delta^2\hspace{-0.4mm}\widebar{N}^2\hspace{-0.6mm} \vA^{\hspace{-0.6mm}t} \hspace{-0.8mm}+\hspace{-0.8mm} 2e_4\mu^2\hspace{-0.6mm}\delta^2\hspace{-0.6mm} \widebar{N}^2 \hspace{-0.2mm}\vB^{t\hspace{-0.2mm}-\hspace{-0.2mm}1} \hspace{-0.8mm}+\hspace{-0.8mm} 2e_5\mu^2 \hspace{-0.6mm}\delta^2\hspace{-0.6mm} \widebar{N}^2 \vC^{t\hspace{-0.2mm}-\hspace{-0.2mm}1} \label{i4}\vspace{5mm}\\ \nnb
		\vC^t &\le c_1 \mu^2 \delta^2 \widebar{N} \bE\|\bar{\sxb}^t_{0}\|^2 + \lambda_3 \bE\|\check{\sxb}_0^t\|^2 \nnb
		&     + c_2 \mu^2 \delta^2\widebar{N} \vA^t  + c_3\mu^2\hspace{-0.6mm}\delta^2\hspace{-0.6mm}\widebar{N}\vB^{t-1}  + c_4\mu^2\delta^2\widebar{N}\vC^{t-1} \label{i5}
	}
	Let $\gamma$ be an arbitrary positive constant whose value will be decided later. From the above inequalities we have
	\eq{\label{zxnwdj9}
		&\ \bE\|\bar{\sxb}^{t+1}_0\|^2 + \bE\|\check{\sxb}_{0}^{t+1}\|^2 + \gamma \left( \vA^{t+1} + \vB^t + \vC^t \right) \nnb
		\le& \left(1 \hspace{-1mm}-\hspace{-1mm} \frac{\widebar{N}}{3}a_1 \mu \nu \hspace{-0.8mm}+\hspace{-0.8mm} c_1 \mu^2 \delta^2 \widebar{N} \right) \bE\|\bar{\sxb}^t_{0}\|^2 \hspace{-0.8mm}+\hspace{-0.8mm} \left( \hspace{-0.8mm} \lambda_2 \hspace{-0.8mm}+\hspace{-0.8mm} \frac{d_1 \mu \delta^2 \widebar{N} }{ \nu} \hspace{-0.8mm} \right) \hspace{-1mm} \bE\|\check{\sxb}^t_{0}\|^2 \nnb
		& \hspace{-0.8mm}+\hspace{-0.8mm}\left(  \frac{d_2\delta^2\mu \widebar{N}}{\nu} \hspace{-0.8mm}+\hspace{-0.8mm} c_2 \mu^2 \delta^2\widebar{N}\hspace{-0.8mm} \right)\hspace{-0.8mm}\vA^t \hspace{-0.8mm}+\hspace{-0.8mm} \left(\hspace{-0.8mm} \frac{d_3\delta^2\mu \widebar{N}}{\nu} + c_3 \mu^2 \delta^2\widebar{N} \right)\vB^{t-1} \nnb
		&  \hspace{-0.8mm}+\hspace{-0.8mm}  \left( \frac{d_4\delta^2\mu \widebar{N}}{\nu} + c_4 \mu^2 \delta^2\widebar{N} \right)\vC^{t-1} \nnb
		& + \gamma f_1 \mu^2\delta^2 \widebar{N}^2 \left( \bE\|\bar{\sxb}_0^{t+1}\|^2 + \bE\|\check{\sxb}_0^{t+1}\|^2 \right) \nnb
		&   + \gamma f_2 \mu^2\delta^2 \widebar{N}^2 (\vA^{t+1} + \vB^t + \vC^t) + \gamma f_3 \mu^2\delta^2 \widebar{N}^2 \bE\|\bar{\sxb}^t_{0}\|^2 \nnb
		&  + \gamma (\lambda_3 + e_6\mu^2\delta^2\widebar{N}^2)\bE\|\check{\sxb}_0^t\|^2 \nnb
		&	+ \gamma f_4 \mu^2 \delta^2  \widebar{N}^2(\vA^{t} + \vB^{t-1} + \vC^{t-1}),
	}
	where the constants $\{f_i\}_{i=1}^4$ are defined as
	\eq{\label{f-notation}
		&f_1 = \max\{12, e_6\},  f_2=2\max\{e_3,e_4,e_5\}, \\
		&f_3 = 12 + c_1,  \hspace{8mm} f_4 \hspace{-0.8mm}=\hspace{-0.8mm} \max\{2e_3\hspace{-0.8mm}+\hspace{-0.8mm}c_2, 2e_4\hspace{-0.8mm}+\hspace{-0.8mm}c_3,2e_5\hspace{-0.8mm}+\hspace{-0.8mm}c_4\}.
	}
	If the step-size $\mu$ is chosen small enough such that
	\eq{
		&\ 1- \frac{\widebar{N}}{3}a_1 \mu \nu + c_1 \mu^2 \delta^2 \widebar{N} \le 1- \frac{\widebar{N}}{4}a_1 \mu \nu, \\
		&\  \lambda_2 + \frac{d_1 \mu \delta^2 \widebar{N} }{ \nu} \le \frac{1+\lambda_2}{2}\define \lambda_4 < 1, \label{mu-4} \\
		&\ \frac{d_2\delta^2\mu \widebar{N}}{\nu} + c_2 \mu^2 \delta^2\widebar{N} \le \frac{2d_2\delta^2\mu \widebar{N}}{\nu}, \\
		&\  \frac{d_3\delta^2\mu \widebar{N}}{\nu} + c_3 \mu^2 \delta^2\widebar{N} \le \frac{2d_3\delta^2\mu \widebar{N}}{\nu}, \label{mu-4-2} \\
		&\ \frac{d_4\delta^2\mu \widebar{N}}{\nu} + c_4 \mu^2 \delta^2\widebar{N} \le \frac{2d_4\delta^2\mu \widebar{N}}{\nu},\label{mu-4-3}
	}
	recursion \eqref{zxnwdj9} can be simplified to
	\eq{\label{zxnwdj9-2}
		&\hspace{-2mm} (1- \gamma f_1 \mu^2\delta^2 \widebar{N}^2 )\left(\bE\|\bar{\sxb}^{t+1}_0\|^2 + \bE\|\check{\sxb}_{0}^{t+1}\|^2\right) \nnb
		&\quad \quad + \gamma (1-f_2 \mu^2\delta^2 \widebar{N}^2) \left( \vA^{t+1} + \vB^t + \vC^t \right) \nnb
		\le&\ \left(1- \frac{\widebar{N}}{4}a_1 \mu \nu \right) \bE\|\bar{\sxb}^t_{0}\|^2 + \lambda_4 \bE\|\check{\sxb}^t_{0}\|^2 \nnb
		&\ +  \frac{2d_2\delta^2\mu \widebar{N}}{\nu}\vA^t +\frac{2d_3\delta^2\mu \widebar{N}}{\nu}\vB^{t-1} + \frac{2d_4\delta^2\mu \widebar{N}}{\nu}\vC^{t-1}  \nnb
		&\    + \gamma f_3 \mu^2\delta^2 \widebar{N}^2 \bE\|\bar{\sxb}^t_{0}\|^2  + \gamma (\lambda_3 + e_6\mu^2\delta^2\widebar{N}^2)\bE\|\check{\sxb}_0^t\|^2 \nnb
		&\ + \gamma f_4 \mu^2 \delta^2  \widebar{N}^2(\vA^{t} + \vB^{t-1} + \vC^{t-1}) \nnb
		\le&\ \left(1- \frac{\widebar{N}}{4}a_1 \mu \nu + \gamma f_3 \mu^2\delta^2 \widebar{N}^2 \right) \bE\|\bar{\sxb}^t_{0}\|^2 \nnb
		&\ + \left[ \lambda_4 +\gamma (\lambda_3 + e_6\mu^2\delta^2\widebar{N}^2) \right] \bE\|\check{\sxb}^t_{0}\|^2 \nnb
		&\  +  \left( \frac{f_5\delta^2\mu \widebar{N}}{\nu} + \gamma f_4 \mu^2 \delta^2  \widebar{N}^2 \right) (\vA^t + \vB^{t-1} + \vC^{t-1}),
	}
	where $f_5 \define 2\max\{d_2, d_3, d_4\}$. To guarantee \eqref{mu-4}--\eqref{mu-4-3}, it is enough to set
	\eq{\label{zxcnwejhud}
		\mu \le \min\left\{ \frac{a_1\nu}{12c_1\delta^2}, \frac{(1-\lambda_2)\nu}{2d_1\delta^2 \widebar{N}}, \frac{d_2}{c_2 \nu}, \frac{d_3}{c_3 \nu}, \frac{d_4}{c_4 \nu} \right\}.
	}
	Since $\nu/\delta <1$, it holds that
	\eq{
		\frac{d_l}{c_l \nu} \ge \frac{d_l}{c_l \nu} \frac{\nu^2}{\delta^2 \widebar{N}} = \frac{d_l \nu}{c_l \delta^2 \widebar{N}},  2\le l \le 4.
	}
	Also recall that $1-\lambda_2 = (1-\lambda)/4$. Therefore, if $\mu$ satisfies
	\eq{\label{box-2}
		\boxed{
			\mu \hspace{-0.8mm}\le\hspace{-0.8mm} \min\hspace{-0.8mm}\left\{\hspace{-0.8mm} \frac{a_1}{12c_1}, \frac{1}{8d_1}, \frac{d_2}{c_2}, \frac{d_3}{c_3 }, \frac{d_4}{c_4 } \hspace{-0.8mm} \right\} \frac{\nu(1-\lambda)}{\delta^2 \widebar{N}} \hspace{-0.8mm}\define\hspace{-0.8mm} C_4 \frac{\nu(1-\lambda)}{\delta^2 \widebar{N}}}
	}
	it also satisfies \eqref{zxcnwejhud}. Next we continue simplifying recursion \eqref{zxnwdj9-2}. Suppose $\mu$ and $\gamma$ are chosen such that
	\eq{\label{gamma-mu-1}
		1- \frac{\widebar{N}}{4}a_1 \mu \nu + \gamma f_3 \mu^2\delta^2 \widebar{N}^2 &\le 1- \frac{\widebar{N}}{8}a_1 \mu \nu,\\
		\lambda_4 +\gamma (\lambda_3 + e_6\mu^2\delta^2\widebar{N}^2) &\le \frac{1+\lambda_4}{2}\define \lambda_5 < 1, \label{gamma-mu-2} \\
		\frac{f_5\delta^2\mu \widebar{N}}{\nu} + \gamma f_4 \mu^2 \delta^2  \widebar{N}^2 &\le \frac{2f_5\delta^2\mu \widebar{N}}{\nu},\label{gamma-mu-3}
	}
	recursion \eqref{zxnwdj9-2} can be further simplified to
	\eq{\label{zxnwdj9-3}
		&\ (1- \gamma f_1 \mu^2\delta^2 \widebar{N}^2 )\left(\bE\|\bar{\sxb}^{t+1}_0\|^2 + \bE\|\check{\sxb}_{0}^{t+1}\|^2\right) \nnb
		& \hspace{2cm} + \gamma (1-f_2 \mu^2\delta^2 \widebar{N}^2) \left( \vA^{t+1} + \vB^t + \vC^t \right) \nnb
		\le&\ \left(1- \frac{\widebar{N}}{8}a_1 \mu \nu \right) \bE\|\bar{\sxb}^t_{0}\|^2 \nnb
		&\ + \lambda_5 \bE\|\check{\sxb}^t_{0}\|^2  + \frac{2f_5\delta^2\mu \widebar{N}}{\nu} (\vA^t + \vB^{t-1} + \vC^{t-1}).
	}
	Now we check the conditions on $\mu$ and $\gamma$ to satisfy \eqref{gamma-mu-1}--\eqref{gamma-mu-3}. Since $\lambda_3<1$, if we choose $\mu$ and $\gamma$ such that
	\eq{
		&\ \lambda_3 + e_6\mu^2\delta^2\widebar{N}^2 \le 1,  \\
		&\ \lambda_4 + \gamma \le \frac{1+\lambda_4}{2}, \label{gamma-mu-4}
	}
	then inequality \eqref{gamma-mu-2} holds. To guarantee \eqref{gamma-mu-1}, \eqref{gamma-mu-3} and \eqref{gamma-mu-4}, it is enough to set
	\eq{\label{x3hsdkjhsd}
		\gamma \le \frac{1-\lambda_4}{2},  \mu \le \sqrt{\frac{1-\lambda_3}{e_6 \delta^2 \widebar{N}^2}},   \gamma \mu \le \min\left\{\frac{a_1 \nu}{8f_3\delta^2 \widebar{N}}, \frac{f_5}{f_4 \nu \widebar{N}} \right\}.
	}
	Moreover, if we further choose step-size $\mu$ such that
	\eq{
		\boxed{\label{23hnsdns}
			\lambda_5 \le 1- \frac{\widebar{N}}{8}a_1 \mu \nu \Longleftrightarrow \mu \le \frac{8(1-\lambda_5)}{a_1 \nu \widebar{N}},}
	}
	recursion \eqref{zxnwdj9-3} becomes
	\eq{\label{m2n8}
		&\hspace{-1cm} (1- \gamma f_1 \mu^2\delta^2 \widebar{N}^2 )\left(\bE\|\bar{\sxb}^{t+1}_0\|^2 + \bE\|\check{\sxb}_{0}^{t+1}\|^2\right) \nnb
		&\hspace{5mm}+ \gamma (1-f_2 \mu^2\delta^2 \widebar{N}^2) \left( \vA^{t+1} + \vB^t + \vC^t \right) \nnb
		\le&\ \left(1- \frac{\widebar{N}}{8}a_1 \mu \nu \right) \left( \bE\|\bar{\sxb}^t_{0}\|^2 + \bE\|\check{\sxb}^t_{0}\|^2 \right) \nnb
		&\hspace{1cm} + \frac{2f_5\delta^2\mu \widebar{N}}{\nu} (\vA^t + \vB^{t-1} + \vC^{t-1})
	}
	When $\mu$ and $\gamma$ are chosen such that
	\eq{\label{nshshshsh}
		1- \gamma f_1 \mu^2\delta^2 \widebar{N}^2 > 0  \Longleftrightarrow  \gamma \mu^2 < \frac{1}{f_1\delta^2 \widebar{N}^2},
	}
	recursion \eqref{m2n8} is equivalent to
	\eq{\label{zxnwsjed9}
		&\hspace{-8mm} \left(\bE\|\bar{\sxb}^{t+1}_0\|^2 + \bE\|\check{\sxb}_{0}^{t+1}\|^2\right) \nnb
		& + \gamma \left(\frac{1-f_2 \mu^2\delta^2 \widebar{N}^2}{1- \gamma f_1 \mu^2\delta^2 \widebar{N}^2}\right)  \left( \vA^{t+1} + \vB^t + \vC^t \right) \nnb
		\le&\ \frac{1- \frac{\widebar{N}}{8}a_1 \mu \nu}{1- \gamma f_1 \mu^2\delta^2 \widebar{N}^2}\left\{ \left( \bE\|\bar{\sxb}^t_{0}\|^2 + \bE\|\check{\sxb}^t_{0}\|^2 \right) \right. \nnb
		& + \left. \frac{2f_5\delta^2\mu \widebar{N}}{\nu(1- a_1\widebar{N}\mu \nu/8)} (\vA^t + \vB^{t-1} + \vC^{t-1})\right\}
	}
	If we also choose $\mu$ such that
	\eq{\label{z823hdn}
		1-f_2 \mu^2\delta^2 \widebar{N}^2 \ge \frac{1}{2},\quad  \mbox{and}\quad   1- \frac{1}{8}a_1\widebar{N} \mu \nu \ge \frac{1}{2},
	}
	recursion \eqref{zxnwsjed9} can be simplified as
	\eq{\label{zljshdjdhdud}
		&\hspace{-10mm} \left(\bE\|\bar{\sxb}^{t+1}_0\|^2 + \bE\|\check{\sxb}_{0}^{t+1}\|^2\right) + \frac{\gamma}{2} \left( \vA^{t+1} + \vB^t + \vC^t \right) \nnb
		\le&\ \frac{1- \frac{1}{8}a_1\widebar{N} \mu \nu}{1- \gamma f_1 \mu^2\delta^2 \widebar{N}^2}\left\{ \left( \bE\|\bar{\sxb}^t_{0}\|^2 + \bE\|\check{\sxb}^t_{0}\|^2 \right) \right. \nnb
		 &\ \left. + \frac{4f_5\delta^2\mu \widebar{N}}{\nu} (\vA^t + \vB^{t-1} + \vC^{t-1})\right\}.
	}
	To guarantee \eqref{z823hdn}, it is enough to set
	\eq{\label{box-3}
		\boxed{
			\mu \le \min\left\{ \sqrt{\frac{1}{2f_2 \delta^2 \widebar{N}^2}}, \frac{4}{a_1 \nu \widebar{N}} \right\}.}
	}
	If we let
	\eq{\label{gamma}\gamma = 8f_5\delta^2\mu \widebar{N}/\nu > 0,}
	then recursion \eqref{zljshdjdhdud} is equivalent to
	\eq{\label{zljshdjdhdudsd}
		&\hspace{-10mm} \left(\bE\|\bar{\sxb}^{t+1}_0\|^2 + \bE\|\check{\sxb}_{0}^{t+1}\|^2\right) + \frac{\gamma}{2} \left( \vA^{t+1} + \vB^t + \vC^t \right) \nnb
		\le&\ \frac{1- \frac{\widebar{N}}{8}a_1 \mu \nu}{1- 8 f_1 f_5 \mu^3\delta^4 \widebar{N}^3 / \nu}\left\{ \left( \bE\|\bar{\sxb}^t_{0}\|^2 + \bE\|\check{\sxb}^t_{0}\|^2 \right) \right.  \nnb
		& \hspace{5mm} + \left. \frac{\gamma}{2} (\vA^t + \vB^{t-1} + \vC^{t-1})\right\}.
	}
	If $\mu$ is small enough such that
	\eq{\label{box-4}
		1 \hspace{-0.8mm}-\hspace{-0.8mm} \frac{8 f_1 f_5 \mu^3\delta^4 \widebar{N}^3}{\nu}\hspace{-0.8mm} >\hspace{-0.8mm} 1 \hspace{-0.8mm}-\hspace{-0.8mm} \frac{1}{8}a_1\widebar{N} \mu \nu  \Longleftrightarrow  \boxed{\mu < \sqrt{\frac{a_1}{64f_1 f_5}}\frac{\nu}{\delta^2 \widebar{N}}}
	}
	it then holds that
	\eq{\label{zljshdjdhdudsds}
		&\hspace{-0.5cm} \left(\bE\|\bar{\sxb}^{t+1}_0\|^2 + \bE\|\check{\sxb}_{0}^{t+1}\|^2\right) + \frac{\gamma}{2} \left( \vA^{t+1} + \vB^t + \vC^t \right) \nnb
		\le&\ \rho \left\{ \left( \bE\|\bar{\sxb}^t_{0}\|^2 + \bE\|\check{\sxb}^t_{0}\|^2 \right)  + \frac{\gamma}{2}(\vA^t + \vB^{t-1} + \vC^{t-1})\right\},
	}
	where
	\eq{
		\rho = \frac{1- \frac{\widebar{N}}{8}a_1 \mu \nu}{1- 8 f_1 f_5 \mu^3\delta^4 \widebar{N}^3 / \nu} < 1.
	}
	Finally, we decide the feasible range of step-size $\mu$. Substituting $\gamma$ into \eqref{x3hsdkjhsd} and
	\eqref{nshshshsh}, it requires
	\eq{\label{237sdy}
		\mu &\le \min\left\{ \frac{1-\lambda_4}{16 f_5} \frac{\nu}{\delta^2 \widebar{N}}, \sqrt{\frac{1-\lambda_3}{e_6}}\sqrt{\frac{1}{\delta^2 \widebar{N}}}, \sqrt{\frac{a_1}{64 f_3 f_5}}\left(\frac{\nu}{\delta^2 \widebar{N}}\right), \right. \nnb
		&\hspace{1cm} \left. \hspace{1cm} \sqrt{\frac{1}{8 f_4}}\frac{1}{\delta \widebar{N}}, \left(\frac{\nu}{8f_1f_5\delta^4 \widebar{N}^3}\right)^{1/3} \right\}.
	}
	Note that $1-\lambda_4 = (1-\lambda)/8$ and $1-\lambda_3 \ge (1-\lambda)/8$, and hence if we restrict $\mu$ as
	\eq{\label{box-5}
			\mu &\le \min\left\{\frac{1}{128 f_5},\sqrt{\frac{1}{8e_6}}, \sqrt{\frac{a_1}{64 f_3 f_5}}, \sqrt{\frac{1}{8 f_4}}, \right.\nnb
			& \hspace{1.2cm} \left. \left(\frac{1}{8f_1f_5}\right)^{1/3}  \right\} \frac{\nu(1-\lambda)}{\delta^2 \widebar{N}} \define  \frac{C_5 \nu (1-\lambda)}{\delta^2 \widebar{N}}}
	it can be verified that such $\mu$ satisfies \eqref{237sdy}. Combining all step-size requirements in \eqref{box-1}, \eqref{box-2}, \eqref{23hnsdns}, \eqref{box-3}, \eqref{box-4} and \eqref{box-5} recalling $1-\lambda_5=(1-\lambda)/16$, we can always find a constant $C$.
	\eq{\label{C}
		C \define \min\left\{ C_1, C_2, C_3, C_4, C_5, \frac{1}{2a_1}, \sqrt{\frac{1}{2f_2}}, {\frac{4}{a_1}}, \sqrt{\frac{a_1}{64f_1f_5}} \right\}
	}
	such that if step-size $\mu$ satisfies
	\eq{
		\mu < \frac{C\nu(1-\lambda)}{\delta^2 \widebar{N}},
	}
	then all requirements in \eqref{box-1}, \eqref{box-2}, \eqref{23hnsdns}, \eqref{box-3}, \eqref{box-4} and \eqref{box-5} will be satisfied. Note that $C$ is independent of $\nu$, $\delta$ and $\widebar{N}$.

\normalsize
\bibliographystyle{IEEEbib}
\bibliography{reference}

\end{document}